\documentclass{article}

\usepackage{arxiv}

\usepackage[utf8]{inputenc} 
\usepackage[T1]{fontenc}    
\usepackage{hyperref}       
\usepackage{url}            
\usepackage{booktabs}       
\usepackage{amsfonts}       
\usepackage{nicefrac}       
\usepackage{microtype}      
\usepackage{lipsum}
\usepackage{graphicx}
\usepackage{subcaption}
\usepackage{amssymb}
\usepackage{amsmath}
\usepackage[percent]{overpic}
\usepackage{caption}

\usepackage{etoolbox}
\graphicspath{ {./images/} }
\title{Physics-Informed Neural Networks for Joint Source and Parameter Estimation in Advection-Diffusion Equations}

\author{
 Brenda Anague \\
 School of Agriculture and Science\\
 University of KwaZulu-Natal \\
 Pietermaritzburg, SA 3201 \\
 \&\\
  AIMS RIC\\
  Kigali, Rwanda\\
   \And
 Bamdad Hosseini \\
  Department of Applied Mathematics \\
  University of Washington\\
   Seattle, WA, USA\\
  \And
 Issa Karambal \\
  AIMS RIC\\
  Kigali, Rwanda \\
  \And 
 Jean Medard Ngnotchouye \\
 School of Agriculture and Science\\
 University of KwaZulu-Natal \\
 Pietermaritzburg, SA 3201 \\
}

\begin{document}
\maketitle
\begin{abstract}
Source inversion problems under sparse measurements for parabolic partial differential equations (PDEs) are particularly challenging to solve using physics-informed neural networks (PINNs), due to their severe ill-posedness and the multiple unknowns involved including the source function and the PDE parameters. Although the neural tangent kernel (NTK) of PINNs has been widely used in forward problems involving a single neural network, its extension to inverse problems involving multiple neural networks remains less explored. In this work, we propose a weighted adaptive approach based on the NTK of PINNs including multiple networks representing the solution, the unknown source, and the PDE parameters. The key idea behind our methodology is to simultaneously solve the joint recovery of these unknowns thereby using the underlying PDE as a constraint that couples multiple unknowns functional parameters, leading to more efficient use of the limited information in the measurements. Furthermore, we provide a theoretical analysis of the approximation error of PINNs in a source inverse problem governed by the advection-diffusion equation (ADE) with known velocity and diffusion coefficients. In particular, we derive an error bound for the estimated source function in terms of the PINN's training error, the noise level and the discretization error. This analysis provides theoretical insight into the factors influencing the approximation accuracy. We present numerical experiments on 2D and 3D ADE using different types of measurements data reflecting practical engineering systems. Our results demonstrate the successful performance of PINNs in solving our source inversion problem and its robustness to noise in the measurements.
\end{abstract}

\keywords{
Source inversion, 
physics-informed neural networks, 
joint recovery.
}

\section{Introduction}
Physics-informed Neural Networks (PINNs) \cite{raissi2019deep} have been presented in the literature as promising deep learning approaches for solving partial differential equations (PDEs) problems that incorporate the PDE as a constraint, allowing for a unified solution of forward and inverse problems \cite{karniadakis2021physics,raissi2019deep,yu2022gradient}. 
One of the advantages of using PINNs is their flexibility in terms of their mathematical formulation and numerical implementation \cite{raissi2019deep}. 
PINNs have also been successfully applied to solve high-dimensional PDEs  in several application domains \cite{zhang2025annealed,Hu2024}. 

In this article we are 
primarily interested in PDE models related to 
models of atmospheric pollutant dispersion, often based on an Advection Diffusion Equation (ADE) \cite{taylor1915eddy}. Many studies have focused on developing forward models for simulating the dispersion of pollutants using physics and empirical data to build models for various ADE parameters that dictate the advection and diffusion processes \cite{seinfeld2016atmospheric}. 
At the same time, one of the core problems in environmental modeling is the task of estimating sources of emissions which amounts to a source inversion problem for the ADE. However, lack of knowledge or presence of uncertainties in the ADE model parameters often amount to severe ill-posedness beyond those of standard source inversion problems. Adding to this the fact that data for source inversion is often quite noisy and scarce leads to an extremely challenging inverse problem.

Motivated by PINNs' capacity to seamlessly integrate data with PDE constraints \cite{raissi2019deep}, in this work, we develop an all-at-once approach  using PINNs to simultaneously invert for sources and estimate unknown parameters of an ADE. 
Broadly speaking, our proposed method takes the following abstract form
\begin{equation}\label{eq:joint-opt}
	\text{minimize}_{u, \gamma, f} \; 
	\text{misfit} (u, \gamma, f, \mathbf{z}^\ast)  
	\quad \text{subject to} \quad \mathcal{F}(u, \gamma) = f,
\end{equation}
where $\mathcal{F}$ denotes the differential operator defining ADE, $u$ is the solution describing the concentration of the pollutant, $\gamma$ denotes unknown parameters such as velocities and diffusion coefficients, $f$ is the unknown source, and $\mathbf{z}^\ast$ is the vector of measurements. We then discretize $u, \gamma, f$ using neural networks and relax the PDE constraint with a Lagrange multiplier to obtain our PINN formulation, the details of this methodology are summarized in Section~\ref{sec:method}. This approach is successful because the PDE constraint restricts the manifold of possible solutions of the inverse problem which together with other prior constraints such as positivity of the source and the solution, and some of the other parameters in the model, enables us to solve the above optimization problem accurately.

\subsection{Main contributions}
To this end, our main contributions in this article can be summarized as follows:

\begin{itemize}
	\item We introduce the joint optimization problem \eqref{eq:joint-opt},
	discretize and solve it using the PINN approach for solving 
	source inversion problems with positivity constraints on unknown 
	parameter fields $\gamma$ in Section~\ref{sec:method}.
	
	\item We derive a Neural Tangent Kernel (NTK) for our source inversion problem  involving multiple neural networks and provide a theoretical justified 
	training procedure for our joint optimization approach in 
	Section~\ref{sec:NTK}.

	\item We apply the training procedure mentioned above to solve 
	various challenging 2D and 3D source inversion problems with sparse 
	observations and constant or variable velocity and diffusion fields 
	that are also assumed to be unknown and only informed by simple 
	prior modeling assumptions in Section~\ref{sec:NumResults}.
	
	\item We use Carleman estimates to derive theoretical error 
	bounds for our source inversion problem in terms of the training error, observation noise levels, and discretization resolution 
	in the idealized setting of known PDE parameters in Section~\ref{generroranalysis}.
\end{itemize}

\subsection{Related work}
There have been many studies on source inversion problems in atmospheric dispersion modeling making use of several PDE solvers as well as both deterministic and Bayesian formulations \cite{addepalli2011source,azevedo2014adaptive,garcia2021simultaneous,hosseini2016bayesian,hosseini2017estimating,hwang2019bayesian, ShankarRao2007,enting2002inverse,Senocak2008,Wade2013, albani2019tikhonov,albani2020accurate,albani2021uncertainty}. In the Bayesian framework, emission sources can be modeled as unknown random variables given by appropriate prior distributions to be updated using the data via Bayes' rule \cite{hosseini2016bayesian, huang2015source, hwang2019bayesian}. However, these works assume that the advection and diffusion parameters of the PDE are known, thereby obtaining a linear source inversion problem.
The work \cite{garcia2021simultaneous} introduced a method for simultaneous source inversion and parameter estimation for a small number of scalar parameters and point sources using Gaussian process model emulators.

PINNs have been widely used for the solution of inverse problems in geosciences \cite{schuster2024review,rasht2022physics, depina2022application,song2021wavefield, huang2024microseismic} as well as numerous engineering applications \cite{kim2024review, jagtap2022physics, sahin2024solving}. The idea of using PINNs for simultaneous 
learning of model parameters and the solution of forward or inverse problems 
has also been explored recently in \cite{chen2021physics}, in source inversion problems for tracking chemical sources \cite{naderi2024towards} as well as the
analogous Gaussian process formulations \cite{jalalian2025data, jalalian2025hamiltonian}, though these works consider the equation learning problem \cite{rudy2017data}; the problem of learning the governing PDE itself. Our work is inspired by these contributions and views inverse problems as a specific instance of equation learning where the unknown source as well as other model parameters take the role of unknown terms in the PDE at hand.

PINN losses is usually unbalanced, since they combine different loss terms which can have different magnitude or convergence rates during the training process, which stands out as one of the challenges associated to the PINN methodology. To ensure a well-balanced loss function, the loss weights multiplying each term have to be chosen carefully in a way that each of the loss terms have the same magnitude during the backpropagation of the loss by the gradient descent algorithm. Several approaches are presented in \cite{wang2021understanding,liu2021dual,wight2020solving,braga2021self,wang2022and}, as adaptive weighting methods to dynamically learn these weights during training. Among them, the NTK introduced in \cite{jacot2018neural} has been used as an adaptive weighting method to describe the training dynamics and convergence of PINNs \cite{wang2022and,wang2021eigenvector,faroughi2025neural}. For these purposes, our work is presented as an extension of the NTK theory of PINNs initially developed for forward problems to a source inversion problem involving multiple neural networks.

\subsection{Outline}
The paper is organized as follows: Section~\ref{sec:method} summarizes our 
problem formulation; Section~\ref{sec:NTK} contains our derivation of loss weights 
using the NTK theory; Section~\ref{sec:NumResults} contains our experiments; 
and Section~\ref{generroranalysis} outlines our error analysis. Concluding 
remarks are given in Section~\ref{sec:conclusion} followed by 
proofs and additional numerical results in the appendix.

\section{Problem setup}
\label{sec:method}
In this section, we recall some preliminary results and methodologies that form the basis of our proposed method followed by our problem setup in detail.

\subsection{PINNs}
Numerical methods  such as Finite Element Methods (FEM) and Finite Difference Methods (FDM), are widely used for solving inverse problems and they have well-established theoretical foundations \cite{braess2001finite,strikwerda2004finite}. 
Recently, \cite{raissi2019deep,raissi2017physics,rudy2017data} introduced PINNS  as a promising tool to solve PDE problems by embedding the physics directly into the loss function of a neural net model, which is then minimized during the training process. This physics loss constrains the neural network not only to learn from the data, as is often the case using artificial neural networks, but also to learn while respecting the physics underneath. The attractive 
feature of PINNs is that they can seamlessly accommodate both forward and inverse PDE problems\cite{raissi2019deep}.  

Let us consider a PDE problem written as 
\begin{eqnarray}
	\left\{
	\begin{array}{r c l}
		\mathcal{F}[u(\mathbf{x},t); \gamma] &=& f(\mathbf{x},t), \quad \mathbf{x}\in\Omega, \quad t\in [0,T], \quad \Omega\subseteq
		\mathbb{R}^k, \quad T\in\mathbb{R}\\
		u(\mathbf{x},0)&=& u_{0}(\mathbf{x}), \quad \mathbf{x}\in\Omega\\
		\mathcal{B}[u(\mathbf{x},t)]&=& g(\mathbf{x},t),  \quad \mathbf{x}\in\partial\Omega, \quad t\in [0,T]
	\end{array}
	\right.
	\label{genPDe}
\end{eqnarray}
with $\Omega$ an open bounded domain of $\mathbb{R}^{k}(k\geq 1)$, $\mathcal{F}$ the differential operator in $\Omega\times [0,T]$, $\mathcal{B}$ the differential operator on $\partial\Omega$, $u_{0}, g$ appropriate functions representing initial and boundary data, 
and $\gamma$ the set of parameters in the differential equation. We will briefly 
outline how PINNs can be used to solve this PDE as well as the associated source 
inversion problem.

\subsubsection{Solving the forward problem} In this context, PINNs aim to approximate the solution $u$ of (\ref{genPDe}), by parameterizing it as a neural net $u(\mathbf{x}, t; \pmb\theta)$ parameterized by a vector of weights $\pmb \theta$, and then minimizing the training loss function:
\begin{eqnarray*}
	\mathcal{L}(\pmb\theta)&=& \frac{\lambda_{r}}{N_{r}}\sum_{j=1}^{N_r}(\mathcal{F}[u(\mathbf{x}_{r}^{j},t_{r}^{j};\pmb\theta); \gamma]- f(\mathbf{x}_{r}^{j},t_{r}^{j}))^{2} 
	+ \frac{\lambda_{i}}{N_{i}}\sum_{j=1}^{N_{i}}(u(\mathbf{x}_{i}^{j},t_{i}^{j};\pmb\theta)- u_{0}(\mathbf{x}_{i}^{j}))^{2}  \\
	&+&  \frac{\lambda_{b}}{N_{b}} \sum_{j=1}^{N_b}(\mathcal{B}[u(\mathbf{x}_{b}^{j},t_{b}^{j};\pmb{\theta_{u}})]- g(\mathbf{x}_{b}^{j},t_{b}^{j}))^{2}, 
	\label{lossfwd}
\end{eqnarray*}
with $\lambda_{r}\in \mathbb{R}^{+}, \lambda_{i}\in \mathbb{R}^{+}, \lambda_{b} \in \mathbb{R}^{+}$ representing the weights multiplying the PDE residual, the initial data misfit,  and the boundary data misfit, respectively. The 
$\{\mathbf{x}_{r}^{j}, t_{r}^{j}\}_{j=1}^{N_r}, \{\mathbf{x}_{i}^{j}, t_{i}^{j}\}_{j=1}^{N_i}, 
\{\mathbf{x}_{b}^{j}, t_{b}^{j}\}_{j=1}^{N_b}$ denote the  sets of collocation points 
in the interior, initial condition, and boundary of the domain of the PDE, used to evaluate the pertinent residuals, constituting a penalty method 
for solving the PDE. This loss function is then minimized using gradient-based algorithms, typically the stochastic gradient descent algorithm, in order to find the optimal neural network parameters $\pmb{\theta}^\ast$.\\

\subsubsection{Solving the source inversion problem} The variational nature of PINNs allows us to 
solve inverse problems in a very similar manner to the forward problem. More precisely, 
multiple neural nets can be employed to represent the solution $u$, as well as the  source term $f$, 
and other unknown parameters $\gamma$ of  (\ref{genPDe}) from noisy data  
$\mathbf{z}^\ast \in \mathbb{R}^{N_z}$ \cite{hwang2019bayesian, lu2021deepxde,raissi2017physics}. Therefore, 
the total loss being minimized takes the form:
\begin{eqnarray*}
	\mathcal{L}(\pmb{\theta}_{u},\pmb{\theta}_{f}, \pmb{\theta}_{\gamma})&=& \frac{\lambda_{r}}{N_{r}}\sum_{j=1}^{N_r}(\mathcal{F}[u(\mathbf{x}^{j}_{r},t^{j}_{r}; \pmb{\theta}_{u}); \pmb{\theta}_{\gamma}]- f(\textbf{x}^{j}_{r},t^{j}_{r};\pmb{\theta}_{f}))^{2}\\
	 &+& \frac{\lambda_{z}}{N_{z}}\sum_{j=1}^{N_{z}}
	( \tau_j \left( u(\cdot;\pmb{\theta}_{u}) \right) - z^\ast_j)^{2} 
	+ \frac{\lambda_{i}}{N_{i}}\sum_{j=1}^{N_{i}}(u(\mathbf{x}_{i}^{j},t_{i}^{j};\pmb\theta_{u})- u_{0}(\mathbf{x}_{i}^{j}))^{2} \\
	&+&  \frac{\lambda_{b}}{N_{b}} \sum_{j=1}^{N_b}(\mathcal{B}[u(\mathbf{x}_{b}^{j},t_{b}^{j};\pmb{\theta}_{u})]- g(\mathbf{x}_{b}^{j},t_{b}^{j}))^{2}
\end{eqnarray*}
where 
$\tau_j$ denotes the measurement operator  associated to the $j$-th measurement entry in $\mathbf{z}^\star$ and
we added the new weight $\lambda_{z}\in \mathbb{R}^{+}$, for the data misfit term. The data loss on the boundary data can also be added in the loss if the boundary data are available. The training process consists in finding the optimal neural network parameters 
$\{\pmb{\theta}_{u}^\ast,\pmb{\theta}_{f}^\ast,\pmb{\theta}_{\gamma}^\ast\}$.

\subsection{The case of ADEs in atmospheric dispersion}
Let $\Omega$ be an open bounded domain in $\mathbb{R}^{k}$ ($k\le 3$) and fix $T>0$ and consider the  ADE:
\begin{eqnarray}
	\left\{
	\begin{array}{r c c}
		u_{t}({\mathbf{x}}, t) + \nabla \cdot\big( V(\mathbf{x}, t) u(\mathbf{x},t)-D(\mathbf{x}, t)\nabla u(\mathbf{x},t)\big)  = 
		f(\mathbf{x}, t), &\quad (\mathbf{x},t)\in\Omega\times [0, T]   \\
		u(\mathbf{x},0)= g(\mathbf{x}) &\quad \mathbf{x}\in\Omega \\
		\mathcal{B}(u)(\mathbf{x},t)=h(\mathbf{x},t), &\quad t\in [0,T], \mathbf{x}\in\partial\Omega,
	\end{array}
	\right.
	\label{ade}
\end{eqnarray}
where $ g:\Omega \rightarrow \mathbb{R}$ and $h: \Omega \times[0,T]  \rightarrow \mathbb{R}$ are known continuous functions describing the boundary and initial data, $V:\Omega \times [0,T] \rightarrow 
\mathbb{R}^k$ is the advection/velocity field, $D:\Omega \times [0,T] \rightarrow \mathbb{R}_{\ge 0}$ is the diffusion coefficient, $f: \Omega \times [0, T] \rightarrow \mathbb{R}$ is the source term, and $\mathcal{B}: \partial\Omega \times [0,T] \rightarrow\mathbb{R}$ is an appropriate boundary operator. We consider the following setting:
\begin{itemize}
	\item We have few noisy observations of the solution $u$ of \eqref{ade} with  additive Gaussian noise:
	\begin{eqnarray}
		z^\ast_i = \tau_i(u) + \varepsilon_{i}, \quad \varepsilon_{i}\sim\mathcal{N}(0,\sigma_i^{2}), \quad \sigma_i\in\mathbb{R}, \quad i = 1, \dots, N_z,
		\label{observ_operator}
	\end{eqnarray}
	where the $\tau_i$ are bounded and linear operators that model our measurements and $\varepsilon_{i}$ represents the noise. 
	\item The source function $f$, the solution $u$, and the parameters $V$ and $D$ are unknowns that we need to estimate.
\end{itemize}
We now parameterize the solution $u$, the velocity $V$, the diffusion $D$ and source term $f$ as neural nets, with slight abuse of notation we write $u(\mathbf{x}, t; \pmb{\theta}_{u})$, $V(\mathbf{x}, t; \pmb{\theta}_{V}), D(\mathbf{x}, t; \pmb{\theta}_{D})$ 
and $f(\mathbf{x}; \pmb{\theta}_{f})$ to denote these neural nets with $\pmb{\theta}$'s denoting the weights and biases of the pertinent networks. For problems where the velocity field $V$ and the diffusion $D$ are constant we simply consider them as parameters for 
the solution, i.e., within the vector $\pmb{\theta}_{u}$.  We then proceed to simultaneously train our neural networks by formulating the following PINN loss:
\begin{eqnarray}
	\mathcal{L}(\boldsymbol{\Theta}) &=& \underbrace{\lambda_{r}\mathcal{L}_{r}(\boldsymbol{\Theta})}_{\textbf{PDE residual}} + \underbrace{\lambda_{b}\mathcal{L}_{b}(\pmb{\theta}_{u})}_{\textbf{Boundary loss}} + \underbrace{\lambda_{z}\mathcal{L}_{z}(\pmb{\theta}_{u})}_{\textbf{Data loss}},
	\label{loss}
\end{eqnarray}
with  $\mathbf{\Theta}=\{\pmb{\theta}_{u}, \pmb{\theta}_{V},\pmb{\theta}_{D}, \pmb{\theta}_{f}\}$, where $\mathcal{L}_{r},\mathcal{L}_{b}$ and $\mathcal{L}_{z}$ denote the residual, boundary, and data losses, respectively:
\begin{eqnarray}
	\mathcal{L}_{r}(\boldsymbol{\Theta})&=& \frac{1}{N_{r}}\sum_{i=1}^{N_{r}}(  \mathcal{F}[u(\textbf{x}^{i}_{r},t^{i}_{r};\pmb{\theta}_{u});\pmb{\theta}_{V};\pmb{\theta}_{D}] -f(\textbf{x}^{i}_{r},t^{i}_{r};\pmb{\theta}_{f}))^{2}\nonumber \\
	\mathcal{L}_{b}(\pmb{\theta}_{u})&=& \frac{1}{N_{b}}\sum_{i=1}^{N_{b}}(\mathcal{B}[u(\textbf{x}^{i}_{b}, t_{b}^{i};\pmb{\theta}_{u})]- h(\textbf{x}^{i}_{b},t^{i}_{b}))^{2} \nonumber\\
	\mathcal{L}_{z}(\pmb{\theta}_{u})&=& 
	\frac{1}{N_z}\sum_{i=1}^{N_{z}} 
	(\tau_i[u(\cdot, \cdot;\pmb{\theta}_{u})]- z^{\ast}_{i} )^{2}\nonumber
	\label{Alllossdef}
\end{eqnarray}
where we introduced the shorthand notation $\mathcal{F}[u(\cdot;\pmb{\theta}_{u});\pmb{\theta}_{V};\pmb{\theta}_{D}]:= u_{t}(\cdot; \pmb{\theta}_{u}) + \nabla \cdot\big(V(\cdot;\pmb{\theta}_{V}) u(\cdot;\pmb{\theta}_{u})- D(\cdot;\pmb{\theta}_{D})\nabla u(\cdot;\pmb{\theta}_{u})\big).$
Given that the PINN loss is usually unbalanced (loss components do not have the same magnitude), therefore, during the training process, the weights $\lambda_{b},\lambda_{r},\lambda_{z}$
need to be  adaptively learned. Next we outline 
a methodology towards this following the NTK theory of \cite{wang2022and}.


\section{The PINNs NTK for source inversion}
\label{sec:NTK}
In this section, we extend the NTK theory of PINNs for forward problems to a source inversion problem with parameter estimation based on the ADE \eqref{ade}. The derived NTK is a combined neural tangent kernel matrix of two separate neural networks, one for the source function $f$ and the second for the solution $u$. In the case of unknown non-constant velocity and diffusion coefficients, the theory remains the same and will simply be extended to a combination of four NTKs components corresponding to their respective neural networks. 
Thus, to keep the notation simple we only present our results for 
the two network case with constant PDE coefficients.

Consider a gradient flow for our PINN loss:
\begin{eqnarray}
	\frac{d\pmb{\Theta}(s)}{ds}= -\nabla_{\pmb{\Theta}}\mathcal{L}(\pmb{\Theta}), \quad \text{with } \pmb{\Theta}(s)=\{\pmb{\theta}_{u}(s), \pmb{\theta}_{f}(s), \gamma(s)\},
	\label{gradflow}
\end{eqnarray}
where $s \ge 0$ is a time parameter that models the flow of our gradient 
descent algorithm.
The following lemma is our main theoretical contribution in this section, the proof of which is summarized in Appendix \ref{sec:prooflemma}. The lemma presents an expression for the 
PINN NTK in terms of the pertinent 
differential operators in the PDE as well as other 
parameters that are computable during optimization.


\begin{lemma}\label{lem:ntk}
	Consider two neural networks $u(\mathbf{x},t,\pmb{\theta}_{u})$ and $f(\mathbf{x},t,\pmb{\theta}_{f})$ to approximate the solution $u$ and the source function $f$, respectively, and the gradient flow \eqref{gradflow}.
	Then along the path of this flow it holds that
	\begin{equation}
		\begin{bmatrix}
			\frac{d( \mathcal{F}[u(\mathbf{x}_{r},t_{r};\pmb{\theta}_{u}(s));\gamma(s)]- f(\mathbf{x}_{r},t_{r};\pmb{\theta}_{f}(s))}{ds}\\
			\frac{d \mathcal{B}[u(\mathbf{x}_{b},t_{b};\pmb{\theta}_{u}(s))}{ds} \\
			\frac{d\mathbb{\tau}[u(\mathbf{x}_{z},t_{z};\pmb{\theta}_{u}(s))]}{ds}     
		\end{bmatrix}= -\mathbf{K}(s)\cdot\begin{bmatrix}
			\mathcal{F}[u(\mathbf{x}_{r},t_{r};\pmb{\theta}_{u}(s));\gamma(s)]- f(\mathbf{x}_{r},t_{r}, \pmb{\theta}_{f}(s))\\
			\mathcal{B}[u(\mathbf{x}_{b},t_{b};\pmb{\theta}_{u}(s))]- h(\mathbf{x}_{b},t_{b})\\
			\mathbb{\tau}[u(\mathbf{x}_{z},t_{z};\pmb{\theta}_{u}(s))]- \mathbf{z}^{*}
		\end{bmatrix}
		\label{derivloss}
	\end{equation}
	where \begin{eqnarray}
		\mathbf{K}(s)= \begin{bmatrix}
			\frac{2\lambda_{r}}{N_r}\mathbf{K}_{rr}(s) &  \frac{2\lambda_{b}}{N_b}\mathbf{K}_{rb}(s) & \frac{2\lambda_{z}}{N_z}\mathbf{K}_{rz}(s)\\
			\frac{2\lambda_{r}}{N_r} \mathbf{K}_{br}(s) &  \frac{2\lambda_{b}}{N_b}\mathbf{K}_{bb}(s) & \frac{2\lambda_{z}}{N_z}\mathbf{K}_{bz}(s)\\
			\frac{2\lambda_{r}}{N_r}\mathbf{K}_{zr}(s) &  \frac{2\lambda_{b}}{N_b}\mathbf{K}_{zb}(s) & \frac{2\lambda_{z}}{N_z}\mathbf{K}_{zz}(s)
		\end{bmatrix}
		\label{ntkmatrix}
	\end{eqnarray}
	and 
	\begin{eqnarray*}
		\left\{
		\begin{array}{r c l}
			(\mathbf{K}_{rr})_{ij}(s)&=& \bigg<\begin{pmatrix}
				\frac{d\mathcal{F}[u(\mathbf{x}^{i}_{r},t^{i}_{r};\pmb{\theta}_{u}(s));\gamma(s)]}{d\pmb{\theta}_{u}}\\
				-\frac{df(\mathbf{x}^{i}_{r},t^{i}_{r}, \pmb{\theta}_{f}(s))\big)}{d\pmb{\theta}_{f}}\\
				\frac{d\mathcal{F}[u(\mathbf{x}^{i}_{r},t^{i}_{r};\pmb{\theta}_{u}(s));\gamma(s)]}{d\gamma}
			\end{pmatrix},\begin{pmatrix}
				\frac{d\mathcal{F}[u(\mathbf{x}^{j}_{r},t^{j}_{r};\pmb{\theta}_{u}(s));\gamma(s)]}{d\pmb{\theta}_{u}}\\
				-\frac{df(\mathbf{x}^{j}_{r},t^{j}_{r}, \pmb{\theta}_{f}(s))\big)}{d\pmb{\theta}_{f}}\\
				\frac{d\mathcal{F}[u(\mathbf{x}^{j}_{r},t^{j}_{r};\pmb{\theta}_{u}(s));\gamma(s)]}{d\gamma}
			\end{pmatrix}  \bigg>\\
			&=& (J_{rr}^{\pmb{\theta}_{u}}(s))(J_{rr}^{\pmb{\theta}_{u}}(s))^{T}+(J_{rr}^{\pmb{\theta}_{f}}(s))(J_{rr}^{\pmb{\theta}_{f}}(s))^{T}+ (J_{rr}^{\boldsymbol{\gamma}}(s))(J_{rr}^{\boldsymbol{\gamma}}(s))^{T} \\\\
			(\mathbf{K}_{bb})_{ij}(s)&=& \bigg<\frac{d\mathcal{B}[u(\mathbf{x}^{i}_{b},t^{i}_{b};\pmb{\theta}_{u}(s))]}{d\pmb{\theta}_{u}}, \frac{d\mathcal{B}[u(\mathbf{x}^{j}_{b},t^{j}_{b};\pmb{\theta}_{u}(s))]}{d\pmb{\theta}_{u}} \bigg>\\\\
			(\mathbf{K}_{zz})_{ij}(s)&=& \bigg< \frac{d\mathbb{\tau}_{i}[u(\mathbf{x}^{i}_{z},t^{i}_{z};\pmb{\theta}_{u}(s))]}{d\pmb{\theta}_{u}}  , \frac{d\mathbb{\tau}_{j}[u(\mathbf{x}^{j}_{z},t^{j}_{z};\pmb{\theta}_{u}(s))]}{d\pmb{\theta}_{u}}  \bigg>\\\\
			(\mathbf{K}_{br})_{ij}(s)&=&\bigg<\frac{d\mathcal{B}[u(\mathbf{x}^{i}_{b},t^{i}_{b};\pmb{\theta}_{u}(s))]}{d\pmb{\theta}_{u}},\frac{d\big(\mathcal{F}[u(\mathbf{x}^{j}_{r},t^{j}_{r};\pmb{\theta}_{u}(s));\gamma(s)]- f(\mathbf{x}^{j}_{r},t^{j}_{r}, \pmb{\theta}_{f}(s))\big)}{d\pmb{\theta}_{u}}  \bigg>\\\\
			(\mathbf{K}_{zr})_{ij}(s) &=& \bigg<\frac{d\mathbb{\tau}_{i}[u(\mathbf{x}^{i}_{z},t^{i}_{z};\pmb{\theta}_{u}(s))]}{d\pmb{\theta}_{u}}, \frac{d\big(\mathcal{F}[u(\mathbf{x}^{j}_{r},t^{j}_{r};\pmb{\theta}_{u}(s));\gamma(s)]- f(\mathbf{x}^{j}_{r},t^{j}_{r}, \pmb{\theta}_{f}(s))\big)}{d\pmb{\theta}_{u}} \bigg>\\\\
			(\mathbf{K}_{zb})_{ij}(s) &=& \bigg<\frac{d\mathbb{\tau}_{i}[u(\mathbf{x}^{i}_{z},t^{i}_{z};\pmb{\theta}_{u}(s))]}{d\pmb{\theta}_{u}} ,\frac{d\mathcal{B}[u(\mathbf{x}^{j}_{b},t^{j}_{b};\pmb{\theta}_{u}(s))]}{d\pmb{\theta}{u}}  \bigg>
		\end{array}
		\right.
	\end{eqnarray*}
	with $\langle \cdot,\cdot \rangle$ representing the Euclidean inner product in the neural network parameters space. 
	Here we have
	$\mathbf{K}_{rr}\in\mathbb{R}^{N_{r}\times N_{r}}, \mathbf{K}_{bb}\in\mathbb{R}^{N_{b}\times N_{b}}, \mathbf{K}_{zz}\in\mathbb{R}^{N_{z}\times N_{z}}$, $\mathbf{K}_{br}\in\mathbb{R}^{N_{b}\times N_{r}}$,
	$\mathbf{K}_{br}=(\mathbf{K}_{rb})^{T}$, $\mathbf{K}_{zr}\in\mathbb{R}^{N_{z}\times N_{r}}, \mathbf{K}_{zr}=(\mathbf{K}_{rz})^{T}, \mathbf{K}_{zb}\in\mathbb{R}^{N_{z}\times N_{b}}$ and $\mathbf{K}_{zr}=(\mathbf{K}_{rz})^{T}$.
\end{lemma}

It has been shown in \cite{wang2022and} that, the NTK $\mathbf{K}(s)$ is a positive semi-definite matrix and converges in probability to a deterministic kernel $\mathbf{K}^{*}\simeq \mathbf{K}(0)$ at initialization. Consequently, \eqref{derivloss} becomes 
\begin{eqnarray*}
	\begin{bmatrix}
		\frac{d\big( \mathcal{F}[u(\mathbf{x}_{r},t_{r};\pmb{\theta}_{u}(s));\gamma(s)]- f(\mathbf{x}_{r},t_{r};\pmb{\theta}_{f}(s))\big)}{ds}\\
		\frac{d \mathcal{B}[u(\mathbf{x}_{b},t_{b};\pmb{\theta}_{u}(s))}{ds} \\
		\frac{d\mathbb{\tau}[u(\mathbf{x}_{z},t_{z};\pmb{\theta}_{u}(s))]}{ds}    
	\end{bmatrix}\simeq - \mathbf{K}(0)
	\begin{bmatrix}
		\mathcal{F}[u(\mathbf{x}_{r},t_{r};\pmb{\theta}_{u}(s));\gamma(s)]- f(\mathbf{x}_{r},t_{r}, \pmb{\theta}_{f}(s))\\\\
		\mathcal{B}[u(\mathbf{x}_{b},t_{b};\pmb{\theta}_{u}(s))]- h(\mathbf{x}_{b},t_{b})\\\\
		\mathbb{\tau}[u(\mathbf{x}_{z},t_{z};\pmb{\theta}_{u}(s))]-\mathbf{z}^{*}
	\end{bmatrix}
\end{eqnarray*}
which implies that 
\begin{eqnarray*}
	\textbf{Q}\begin{pmatrix}
		
		\begin{bmatrix}
			\frac{d\big( \mathcal{F}[u(\mathbf{x}_{r},t_{r};\pmb{\theta}_{u}(s));\gamma(s)]- f(\mathbf{x}_{r},t_{r};\pmb{\theta}_{f})\big)}{ds}\\
			\frac{d  \mathcal{B}[u(\mathbf{x}_{b},t_{b};\pmb{\theta}_{u}(s))}{ds} \\
			\frac{dd\mathbb{\tau}[u(\mathbf{x}_{z},t_{z};\pmb{\theta}_{u}(s))]}{ds}     
		\end{bmatrix} -  \begin{bmatrix}
			0\\
			h(\mathbf{x}_{b},t_{b})\\
			\mathbf{z}^\ast
	\end{bmatrix}  \end{pmatrix}  \simeq - e^{\Gamma t} \textbf{Q} \begin{bmatrix}
		0\\
		h(\mathbf{x}_{b},t_{b})\\
		\mathbf{z}^\ast
	\end{bmatrix}
\end{eqnarray*}
since $\mathbf{K}(0)$ is symmetric positive semi-definite, and can be written as $\mathbf{K}(0)= \textbf{Q}^{T}\Gamma\textbf{Q}$ where $\textbf{Q}$ is an orthogonal matrix \cite{wang2022and}. This
means that the residual, boundary and data loss terms in the objective converges at the rates $e^{-\kappa_{r_{i}}s}, e^{-\kappa_{b_{i}}s}, e^{-\kappa_{z_{i}}s}$ respectively, where $\kappa_{r}, \kappa_{b}, \kappa_{z}$ are the eigenvalues of $\mathbf{K}_{rr}, \mathbf{K}_{bb}$ and $\mathbf{K}_{zz}$.\\

The NTK of PINNs was derived in  \cite{wang2022and}
and our calculations largely follow the methodology
of that work. However, the weighted adaptive method based on NTK of PINNs have been less explored on inverse problems and source inversion problems with parameter estimation as compared to forward problems. To the best of our knowledge, this is the first time this adaptive method is used to enhance PINNs training in such source inverse problems on advection-diffusion equations. We describe the details of this 
approach in the next section.

\section{Numerical results} \label{sec:NumResults}
In this section, we summarize our algorithms and present our numerical results for various 
synthetic instances of the ADE in the 2D and 3D for both variable and constant 
velocity and diffusion fields.

\subsection{Algorithm and setup summary} \label{sec:alg}

In the most general case 
we use four
separate neural networks to approximate the unknowns $u,V,D$ and $f$. At the early stage of the training the  weights
$\lambda_{r},\lambda_{b},\lambda_{z}$ are initialized to 1 before being adaptively updated
using the formula \eqref{weightloss} as given in \cite{wang2022and}: 
\begin{eqnarray}
	\lambda_{r} = \frac{Tr(\boldsymbol{K}(s))}{Tr(\boldsymbol{K}_{rr}(s))}, \quad
	\lambda_{b} = \frac{Tr(\boldsymbol{K}(s))}{Tr(\boldsymbol{K}_{bb}(s))},\quad
	\lambda_{z} = \frac{Tr(\boldsymbol{K}(s))}{Tr(\boldsymbol{K}_{zz}(s))}.
	\label{weightloss}
\end{eqnarray}
At this pre-training 
stage we use stochastic gradient descent. 
Once these weights are adapted and remain consistent, we fix their values and start the second phase of the training where L-BFGS is used to update the network parameters $\pmb{\theta}_{u},\pmb{\theta}_{f},\pmb{\theta}_{D},\pmb{\theta}_{V}$. This process 
is summarized in Algorithm~\ref{alg:1}.

\begin{algorithm}[htp]
	\caption{PINNs + NTK Training Algorithm for Source Inversion Problem }\label{alg:1}
	\begin{algorithmic}[1]
		\State \textbf{Initialize} 4 separate feedforward neural networks $u(\mathbf{x},t,\pmb{\theta}_{u}),f(\mathbf{x},t,\pmb{\theta}_{f}),V((\mathbf{x},t,\pmb{\theta}_{V})),D(\mathbf{x},t,\pmb{\theta}_{D})$ (when $V$ and $D$ are variable coefficients) or initialize $V$ and $D$ to 1.0 when they are constants.
		\State \textbf{Initialize} $\lambda_{r} = \lambda_{b} = \lambda_{z}=1$
		\State \textbf{Backpropagation} of $\mathcal{L}(\pmb{\Theta})$ with ADAM and learning rate $\eta$
		\For{$j = 1$ to $\max_{\rm iter}$}
		\State Compute the loss function $\mathcal{L}(\pmb{\Theta})$ in \eqref{loss}
		\State Compute the NTK $\pmb{K}$ in \eqref{ntkmatrix}
		\State $\pmb{\theta}_{u}^{j}\leftarrow \pmb{\theta}_{u}^{j-1}-\eta\nabla_{\pmb{\theta}_{u}}\mathcal{L}(\pmb{\Theta})$ 
		\State $\pmb{\theta}_{f}^{j}\leftarrow \pmb{\theta}_{f}^{j-1}-\eta\nabla_{\pmb{\theta}_{f}}\mathcal{L}(\pmb{\Theta})$ 
		\State $\pmb{\theta}_{V}^{j}\leftarrow \pmb{\theta}_{V}^{j-1}-\eta\nabla_{\pmb{\theta}_{V}}\mathcal{L}(\pmb{\Theta})$ 
		\State $\pmb{\theta}_{D}^{j}\leftarrow \pmb{\theta}_{D}^{j-1}-\eta\nabla_{\pmb{\theta}_{D}}\mathcal{L}(\pmb{\Theta})$
		\State Update the loss weights: \begin{eqnarray*}
			\lambda_{r} = \frac{Tr(\boldsymbol{K})}{Tr(\boldsymbol{K}_{rr})}, \quad
			\lambda_{b} = \frac{Tr(\boldsymbol{K})}{Tr(\boldsymbol{K}_{bb})},\quad
			\lambda_{z} = \frac{Tr(\boldsymbol{K})}{Tr(\boldsymbol{K}_{zz})},
		\end{eqnarray*}
		\EndFor
		\State \textbf{Update} 
		Keep $\lambda_r, \lambda_b, \lambda_z$ fixed 
		and use  L-BFGS to update network weights 
		$\pmb{\theta}_{u},\pmb{\theta}_{f},\pmb{\theta}_{V},\pmb{\theta}_{D}$
		
		\State \Return $\pmb{\theta}_{u},\pmb{\theta}_{f},\pmb{\theta}_{V},\pmb{\theta}_{D}$
	\end{algorithmic}
\end{algorithm}

In some of our experiments, we consider fixed parameters $V, D$ or constant fields in which case
the pertinent neural nets will be omitted or replaced with scalars. 
In our numerical experiments, we consider noisy observations on the solution $u$ 
inside the spacial domain $\Omega$, these will be 
described on a case by case basis below, and we assume  we have few noisy data of the solution at
$\partial\Omega$. We select a fraction $\frac{\rho}{1000}$ ($\rho>0$) of the total amount of boundary data, for training. We chose $\rho$ to be 11 in the 2D case with constants coefficients, and 1 in 2D and 3D case with variable coefficients.
All experiments were performed using the TorchPhysics library\cite{TorchPhysics}, on an NVIDIA RTX 5000 Ada Generation GPU.

\subsection{2D ADE with constant coefficients}
We consider the 2D ADE \eqref{ade} on $\Omega=[0,1]\times[0,1]$. We assume that the boundary of the domain is reflective to pollutants particles, therefore the boundary condition on \eqref{ade} becomes a Neumann boundary condition $ \nabla_{n}(u)(\textbf{x},t)= 0 \quad    t\in [0, 1], \quad \textbf{x}\in\partial\Omega$. We consider the initial condition $ u(\textbf{x},0)= 0 \quad \textbf{x}\in\Omega$ and assume constant diffusion $D$ and velocities coefficient $V$.

We assume that the source function is a mixture of two Gaussian functions: 
\begin{eqnarray*}
	f(x,y)= \exp\bigg(-\frac{(x-\mu_{1})^{2}}{2l_{1}^{2}}\bigg)+ \exp\bigg(-\frac{(y-\mu_{2})^{2}}{2l_{2}^{2}}\bigg)
	\label{sourcefunction}
\end{eqnarray*}
We first solved the above problem using FEM, with $V= (0.2, -0.2), D=0.01, \mu_{1}=0.25, \mu_{2}=0.75, l_{1}=0.06$ and $l_{2}=0.04$ to 
generate a ground truth solution. From the FEM solution, we derive two types of observation data that correspond to the following assumptions.
\begin{itemize}
	\item (Pointwise Observations) The observations result from sensors taking direct time series measurements 
	of the air pollutant concentration, that is, in \eqref{observ_operator} 
	we took $\tau_i(u) = u(\mathbf{x}_i, t_i)$ for a fixed location 
	$\mathbf{x}_i$ and time $t_i$.
	\item (Accumulative Observations) Observations result from sensors capturing average of air pollutants concentrations over a fixed time interval  of length $\delta_{t}$, i.e, $\tau_{i}(u)=\int_{t_{i}}^{t_{i}+\delta t}u(\mathbf{x}_{i},s) ds$. 
\end{itemize}

\subsubsection{Pointwise observations}
\begin{figure}[htp]
    \centering
    \setlength{\tabcolsep}{5pt}
    \renewcommand{\arraystretch}{1.2}
    \begin{tabular}{c c c c }
    \raisebox{40pt}{\rotatebox[origin=c]{90}{}}&
        \begin{overpic}[height=0.13\textheight] {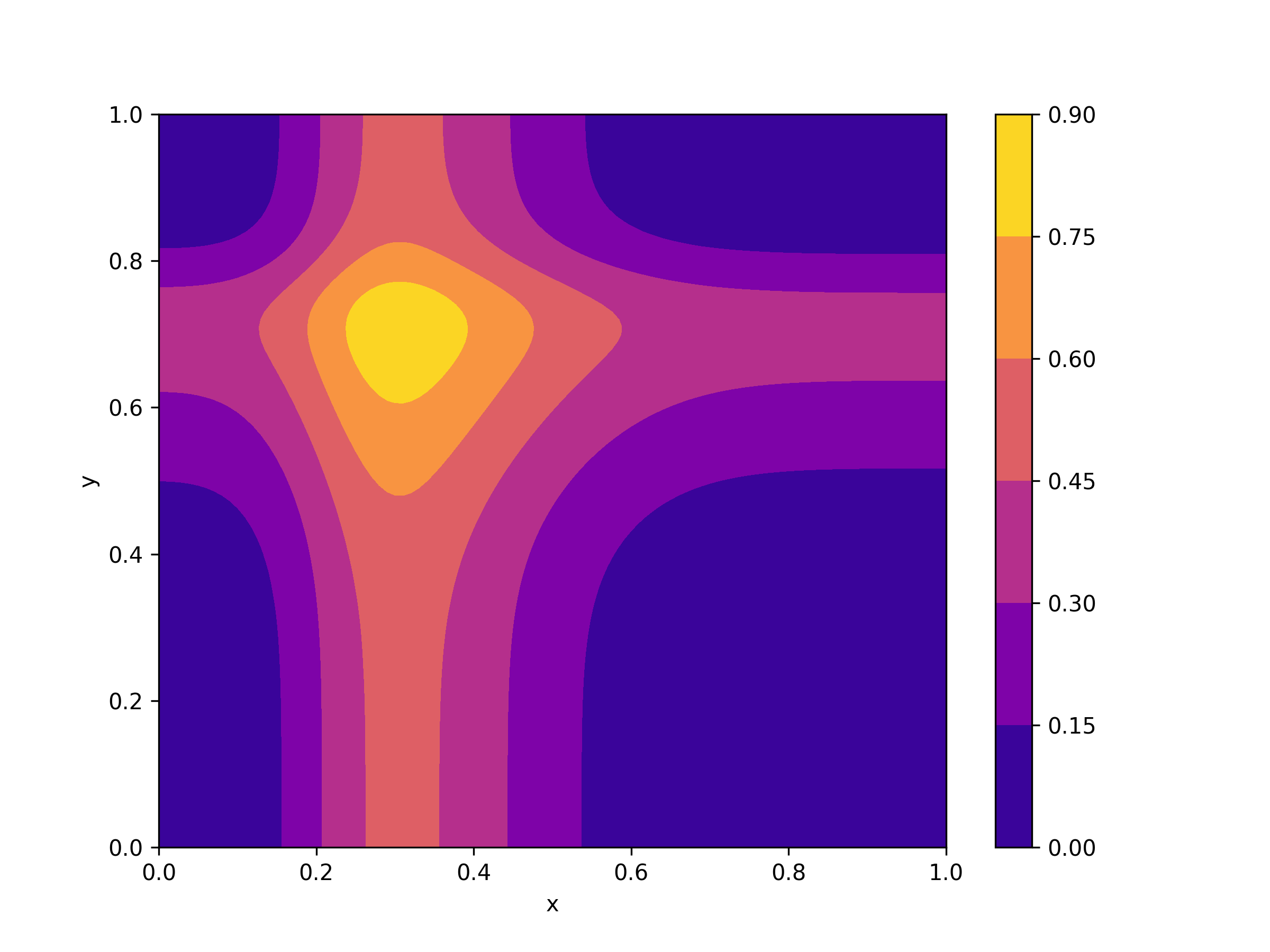}
            \put(20,74){\textbf{FEM solution $u$}}
            \end{overpic}
    \begin{overpic}[height=0.12\textheight] {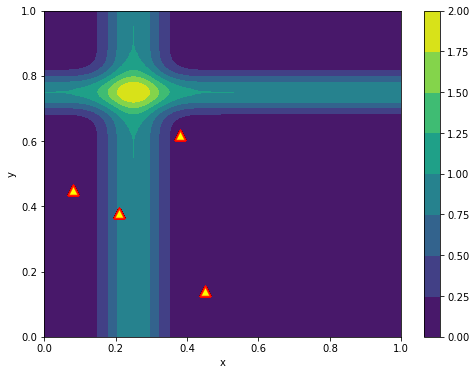}
            \put(10,85){\textbf{Exact $f$ with 4 obs}}
            \end{overpic}&
    \begin{overpic}[height=0.12\textheight] {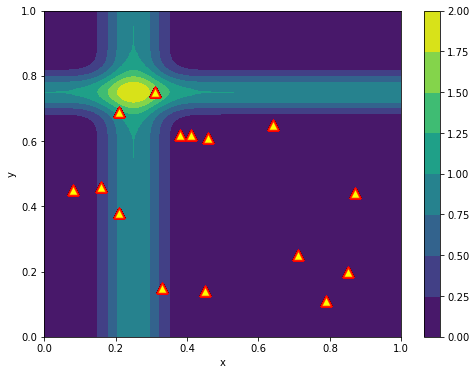}
            \put(10,85){\textbf{Exact $f$ with 15 obs}}
            \end{overpic}
    \end{tabular}
    \caption{\textbf{Contour Plots of the FEM solution at $t=1$(left), exact source function with 4 measurements locations(middle) and 15 measurement locations (right), with the the selected measurements locations represented by the triangles.}}
    \label{fig:ExactsolSource}
\end{figure}

We randomly choose 4 and 15 spatial locations at which we have noisy time series measurements of the FEM solution as shown in Figure~\ref{fig:ExactsolSource}.
We have considered two separate neural networks $u(\textbf{x},t;\pmb{\theta}_{u})$ with 3 hidden layers, 80 neurons per layer and $f(\textbf{x};\pmb{\theta}_{f})$ with 3 hidden layers, 100 neurons per layer to approximate $u$ and $f$ respectively with both 
networks using $\tanh$ activation functions. We applied Algorithm~\ref{alg:1} with
15000 ADAM optimizer steps and learning rate  $\eta= 10^{-3}$ followed by 20000 iterations of L-BFGS with learning rate $\eta= 0.1$.

\begin{figure}[htp]
    \centering
    \setlength{\tabcolsep}{5pt}
    \renewcommand{\arraystretch}{1.2}
    
    \begin{tabular}{c c c c c}
       \raisebox{30pt}{\rotatebox[origin=c]{90}{\textbf{$4$ obs}}} & 
        \begin{overpic}[height=0.09\textheight] {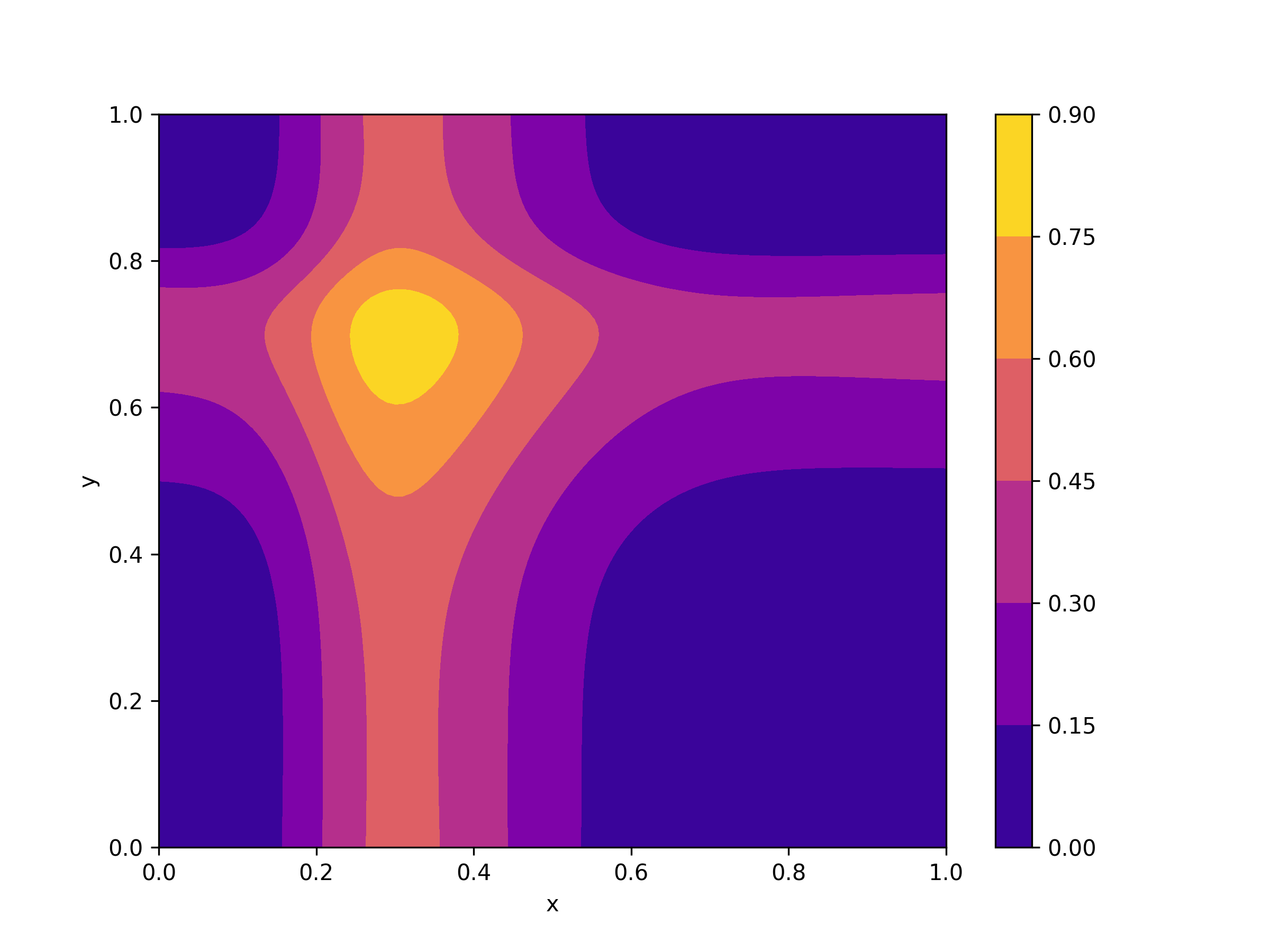}
            \put(20,74){\textbf{Predicted $u$}}
        \end{overpic} &
        \begin{overpic}[height=0.09\textheight]{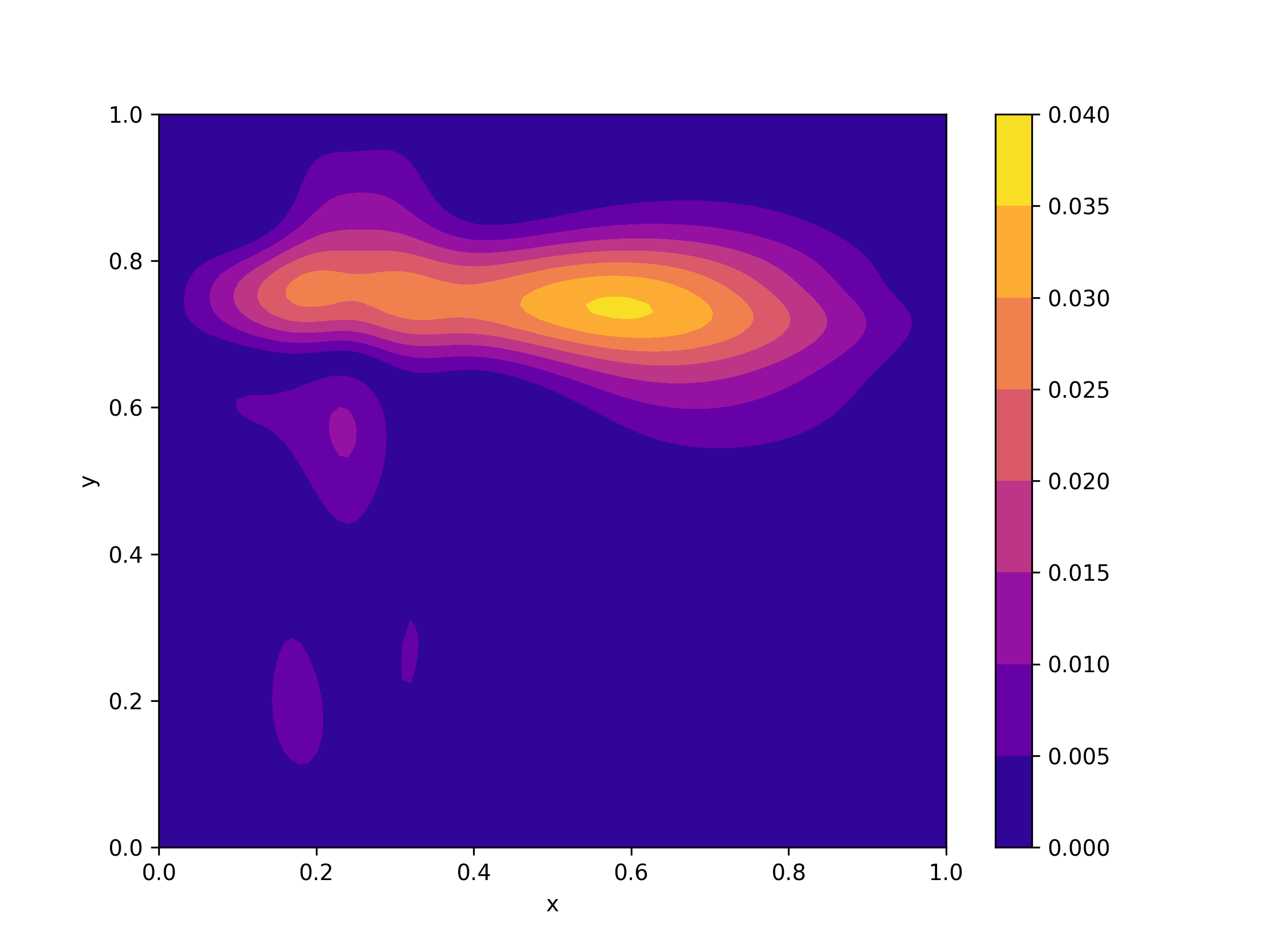}
            \put(10,74){\textbf{Error on $u$ }} 
        \end{overpic} &
        \begin{overpic}[height=0.09\textheight]{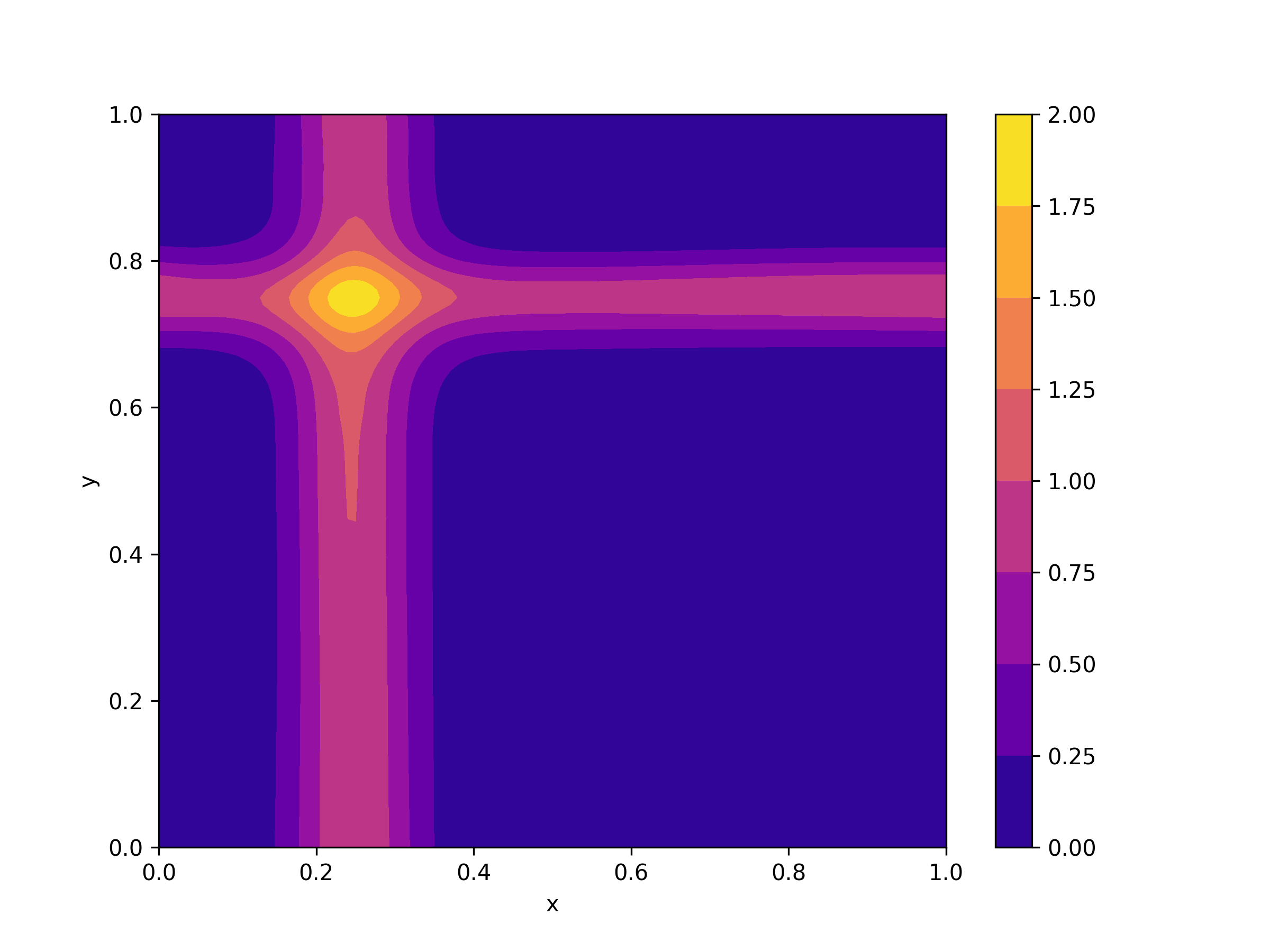}
            \put(20,74){\textbf{Predicted $f$}}
        \end{overpic} &
        \begin{overpic}[height=0.09\textheight]{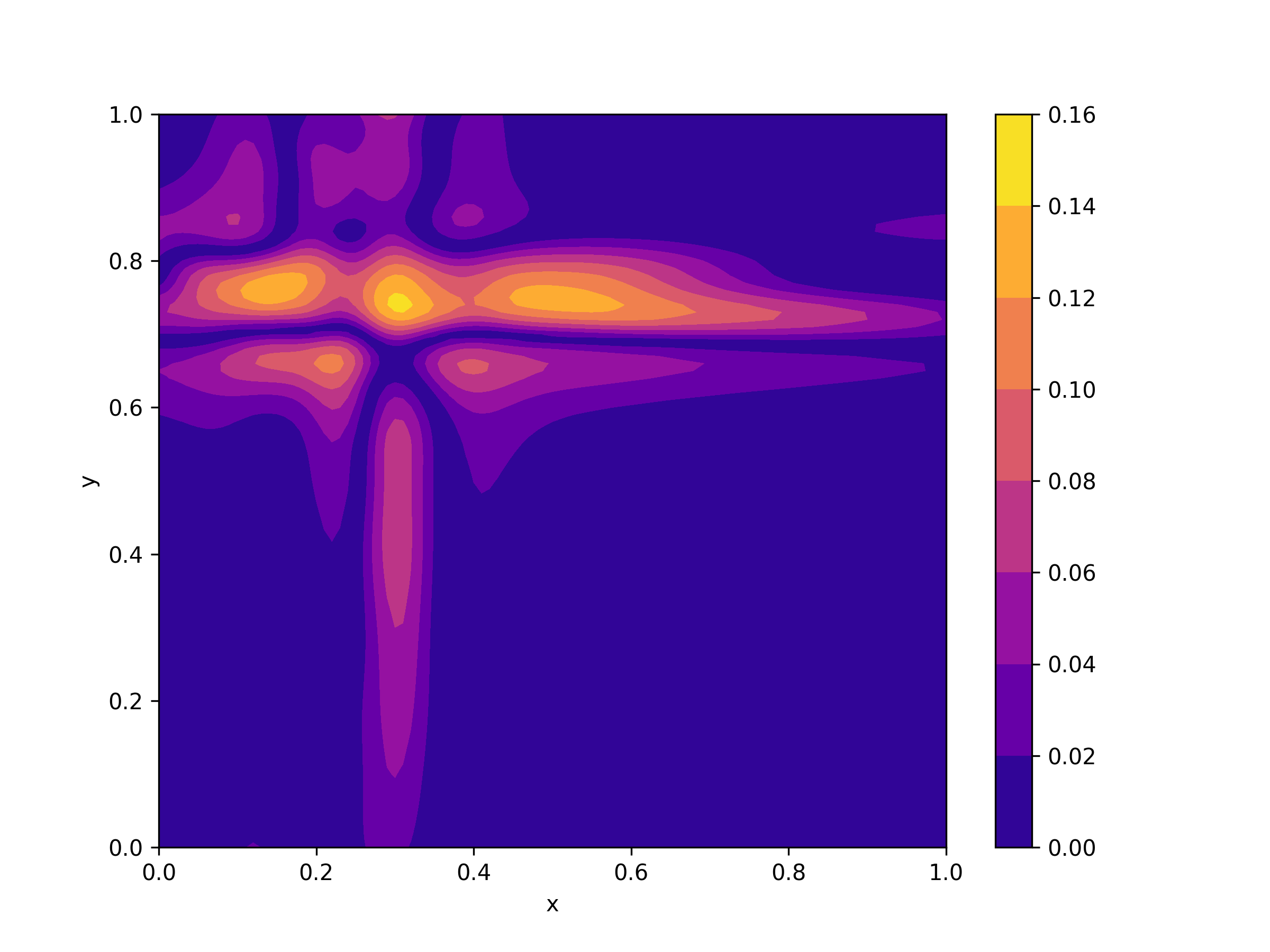}
           \put(20,74){\textbf{Error on $f$}}
        \end{overpic}\\
        \raisebox{30pt}{\rotatebox[origin=c]{90}{\textbf{$15$ obs}}} & 
        \begin{overpic}[height=0.09\textheight] 
        {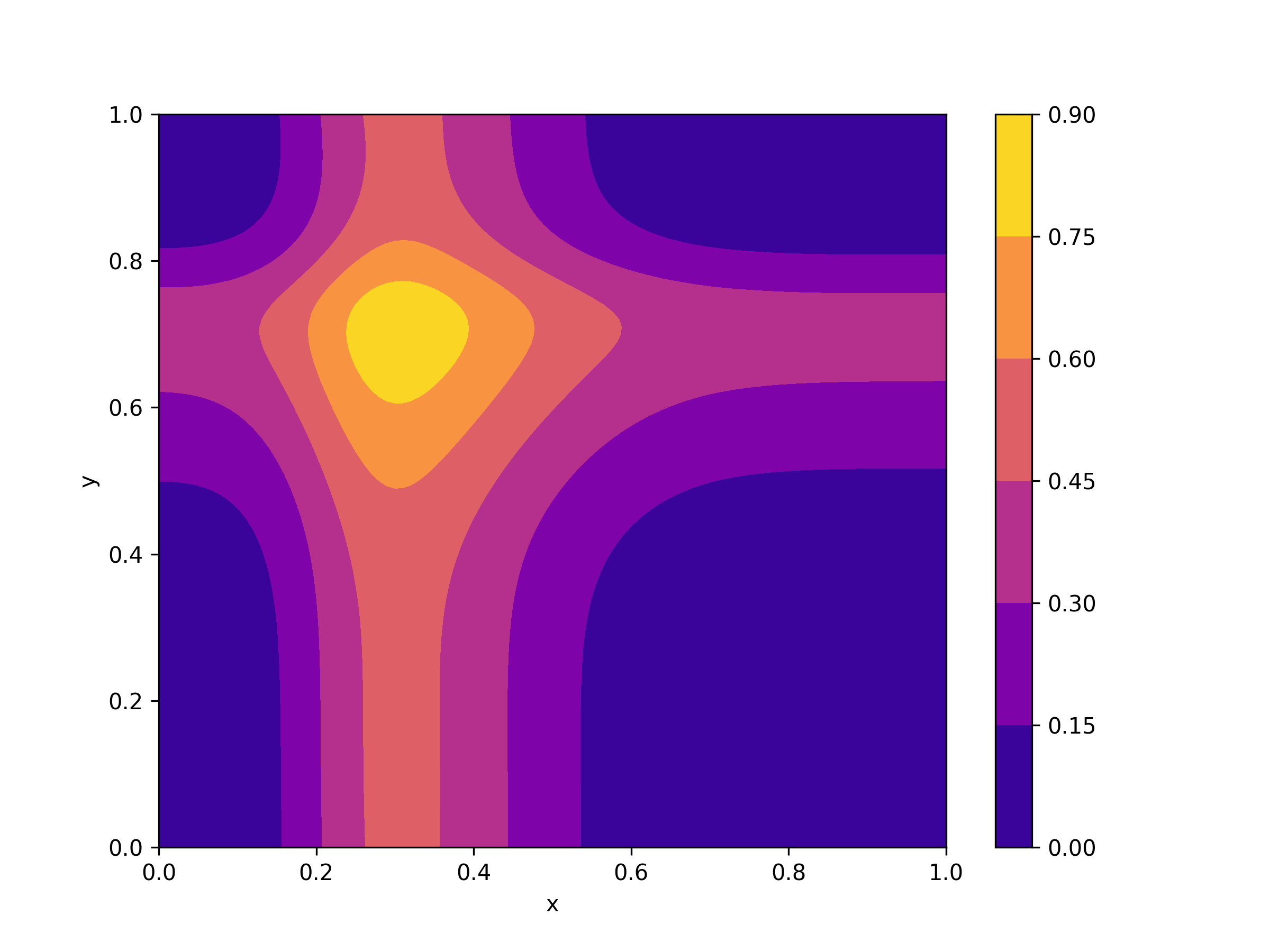}
        \end{overpic} &
        \begin{overpic}[height=0.09\textheight]
        {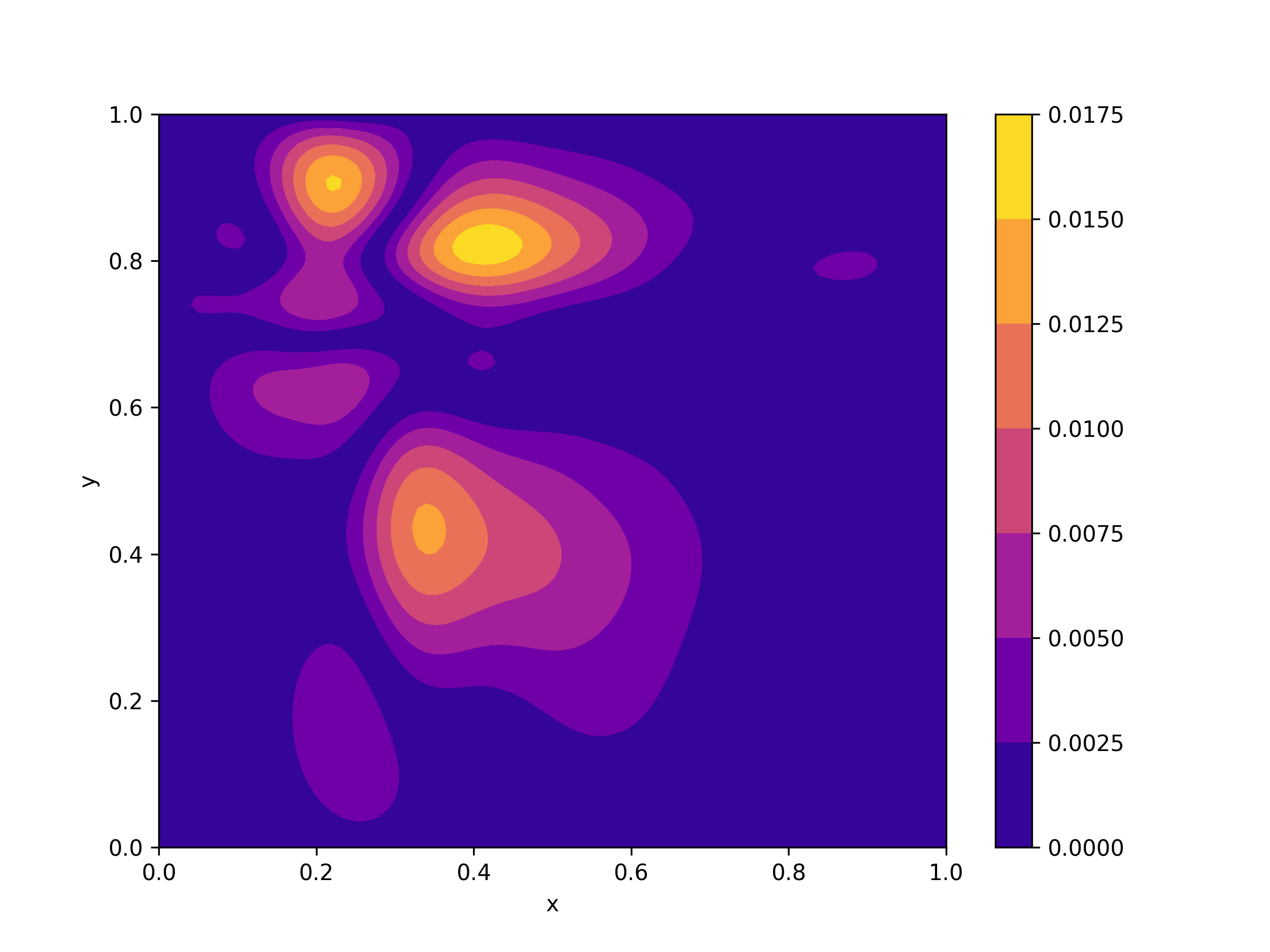}
        \end{overpic} &
        \begin{overpic}[height=0.09\textheight]
        {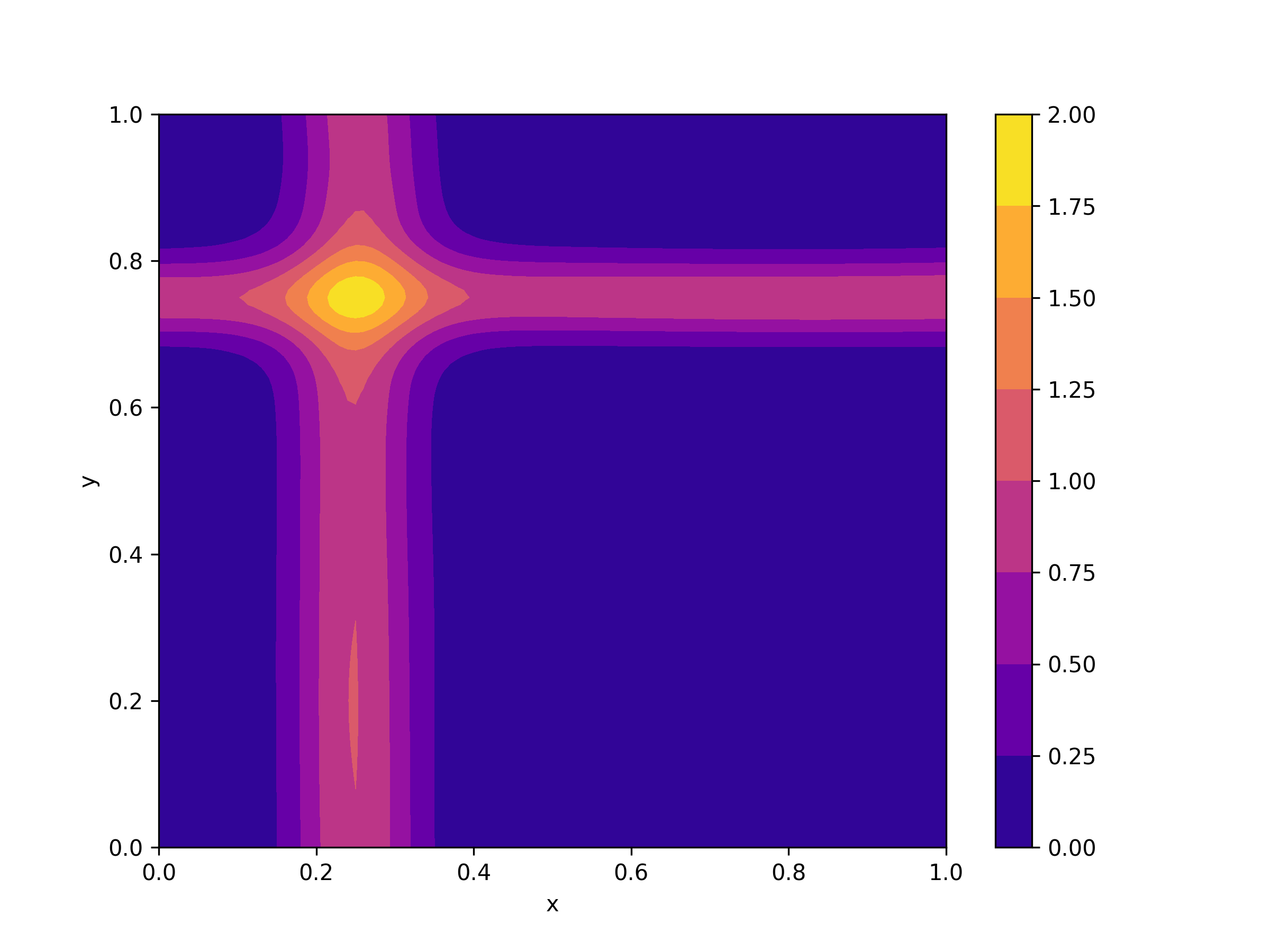}
        \end{overpic} &
        \begin{overpic}[height=0.09\textheight]
        {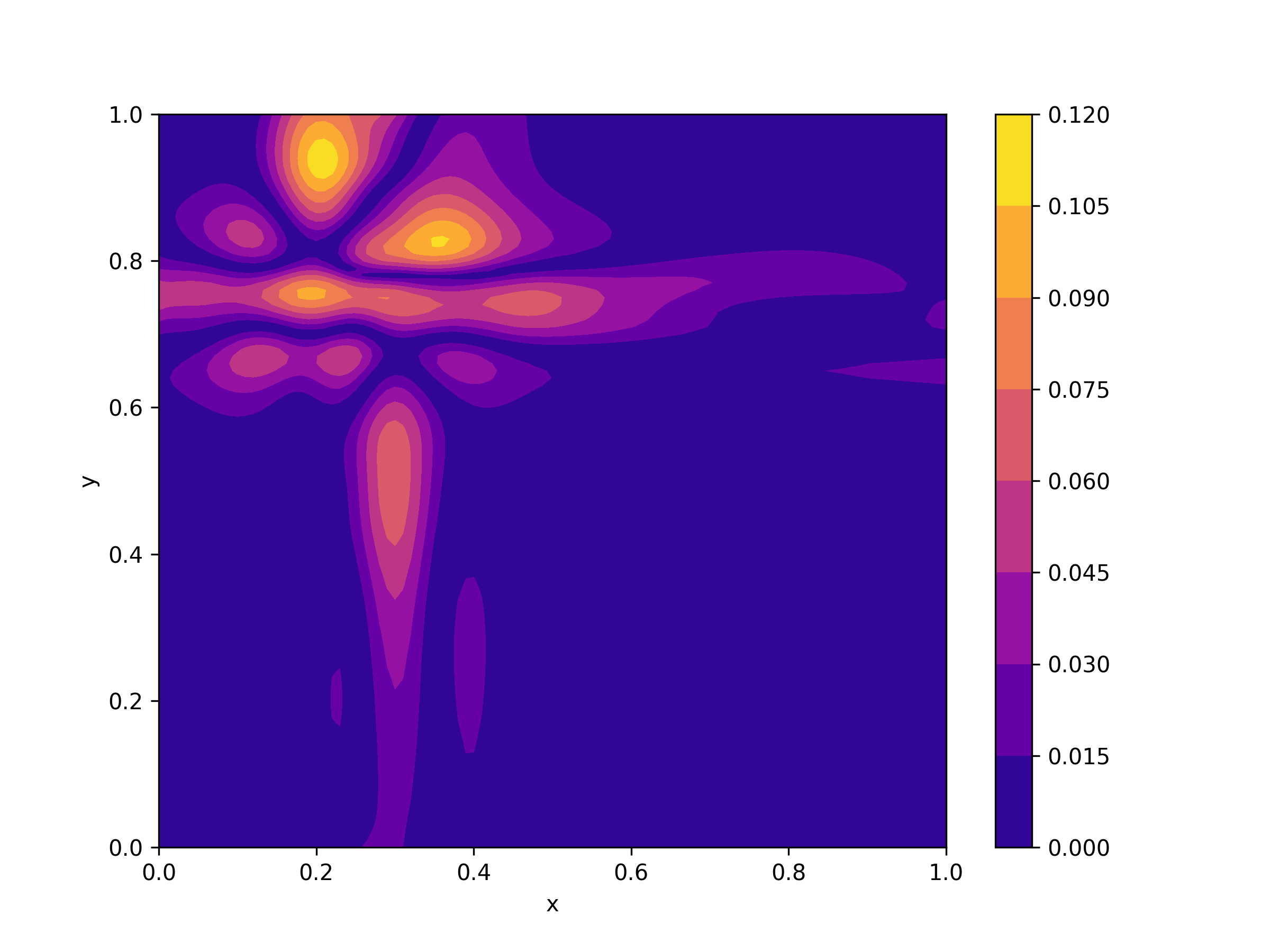}
        \end{overpic}
    \end{tabular}
    
    \caption{\textbf{Prediction and  absolute error of $u$ at $t=1$ and $f$, with $1\%$ of noise.}}
    \label{fig:1percsolSourceError}
\end{figure}

We observe in Figure ~\ref{fig:1percsolSourceError} (see also Figures~\ref{solID}  
and \ref{sourceId}) that despite having only 4 pointwise measurements, 
our method is able to obtain acceptable estimates of the  solution $u$ and 
the source $f$ at various noise levels. 
We also see that by increasing the number of observations to 15 points the recovery
of both $u$ and $f$ improved as expected. 
We consistently observed that the solution $u$ was recovered more accurately than $f$. 
This is natural due to the fact that we had access to direct observations of $u$.

We repeated our experiments 5 times by randomizing the noise  and report our 
average recovered values of the velocities and the diffusion coefficient in Table \ref{tab4ObsID}. We observed that our method is generally capable of 
recovering these unknown parameters from the scarce data while 
giving good recoveries of $u$ and $f$.

\begin{table}[htp]
\centering
\begin{tabular}{|c|c|c|c|c|c|c|}
    \cline{1-7} 
    \textbf{$4$ Obs}
     & \multicolumn{2}{c|}{\textbf{Vx}} 
     & \multicolumn{2}{c|}{\textbf{Vy}} 
     & \multicolumn{2}{c|}{\textbf{D}} \\
    \hline
    \textbf{Noise} &\textbf{ Predicted} &\begin{tabular}[c]{@{}c@{}}\textbf{Relative}\\\textbf{error}\end{tabular}  & \textbf{ Predicted} &\begin{tabular}[c]{@{}c@{}}\textbf{Relative}\\\textbf{error}\end{tabular}    & \textbf{Predicted}  & \begin{tabular}[c]{@{}c@{}}\textbf{Relative}\\\textbf{error}\end{tabular}  \\
   \hline
    $1\%$ &  0.20074  &0.37\%  & -0.2014 &0.7\% & 0.00944 & 5.6\% \\
    \hline
    $5\%$ & 0.20151  & 0.75\% &  -0.20174 &0.87\% & 0.00933 & 6.7\% \\
    \hline
    $10\%$ & 0.2038 & 1.9\% & -0.2058 & 2.9\ & 0.00917 & 8.3\% \\
    \hline
    \cline{1-7} 
    \textbf{15 Obs}
      & \multicolumn{2}{c|}{\textbf{Vx}} 
     & \multicolumn{2}{c|}{\textbf{Vy}} 
     & \multicolumn{2}{c|}{\textbf{D}} \\
    \hline
    \textbf{Noise} & \textbf{Predicted} &\begin{tabular}[c]{@{}c@{}}\textbf{Relative}\\\textbf{error}\end{tabular}  & \textbf{Predicted}  & \begin{tabular}[c]{@{}c@{}}\textbf{Relative}\\\textbf{error}\end{tabular}   & \textbf{Predicted}  &\begin{tabular}[c]{@{}c@{}}\textbf{Relative}\\\textbf{error}\end{tabular}  \\
   \hline
    $1\%$ & 0.19781  &1.1\%  & -0.19782 & 1.1\% & 0.00974 & 2.6\% \\
    \hline
    $5\%$ & 0.1962 & 1.94\% & -0.19496 & 2.58\% & 0.0096 &4.16 \% \\
    \hline
    $10\%$ & 0.19593 & 2\% & -0.19313 & 3.55\% & 0.00957 & 4.49\% \\
    \hline
    
\end{tabular}
\caption{\textbf{Predicted values of the velocities and diffusion coefficients with their relative error, assuming 4 and 15 observations locations with an identity observation operator}}
\label{tab4ObsID}
\end{table}

\begin{figure}[htp]
    \centering
    \setlength{\unitlength}{1mm} 
    \renewcommand{\arraystretch}{1.2}
    \begin{tabular}{c c  }
       \raisebox{30pt}{\rotatebox[origin=c]{90}{\textbf{}}} & 
        \begin{overpic}[height=0.15\textheight] {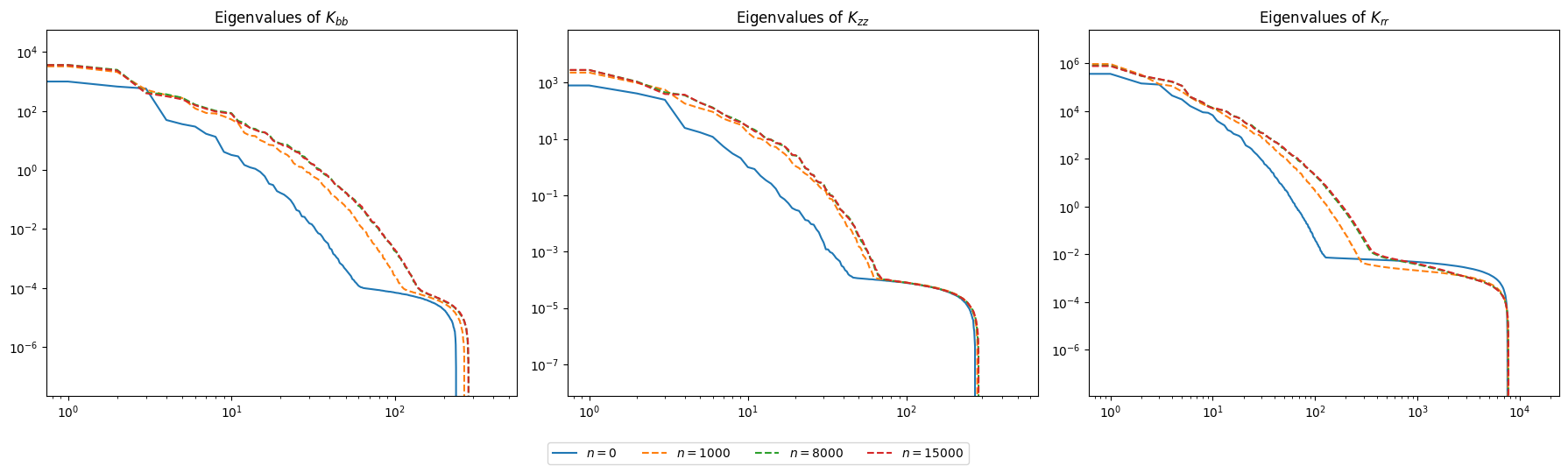}
             \put(15,2){\makebox(0,0){Index}} 
            \put(-2,10){\rotatebox{90}{Eigenvalue}} 
        \end{overpic}
    \end{tabular}
    \caption{ \textbf{Eigenvalues  in a source inversion problem of $\mathbf{K}_{bb}$,  $\mathbf{K}_{zz}$,  $\mathbf{K}_{rr}$ vs their index, with 15 observations on $u$.}}
    \label{fig:eigenvaluesinverse}
 \end{figure}

\begin{figure}[htp]
    \centering
    \setlength{\unitlength}{1mm} 
    \renewcommand{\arraystretch}{1.2}
    
    \begin{tabular}{c c c c }
       \raisebox{30pt}{\rotatebox[origin=c]{90}{\textbf{}}} & 
        \begin{overpic}[height=0.13\textheight] {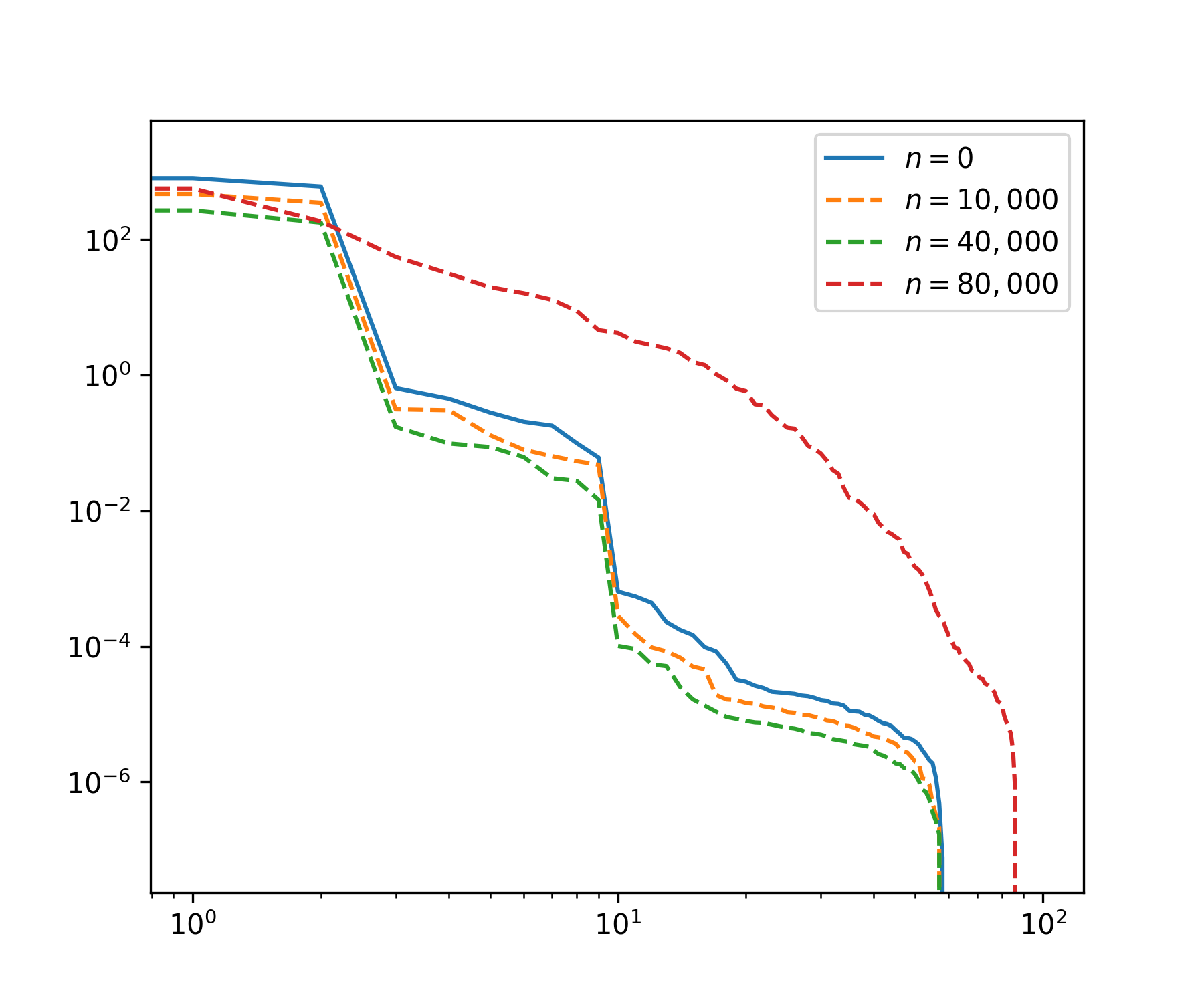}
             \put(50,0.5){\makebox(0,0){Index}} 
            \put(-4,20){\rotatebox{90}{Eigenvalue}} 
            \put(30,77){$\mathbf{K}_{i}$}
        \end{overpic} &
        \begin{overpic}[height=0.13\textheight]{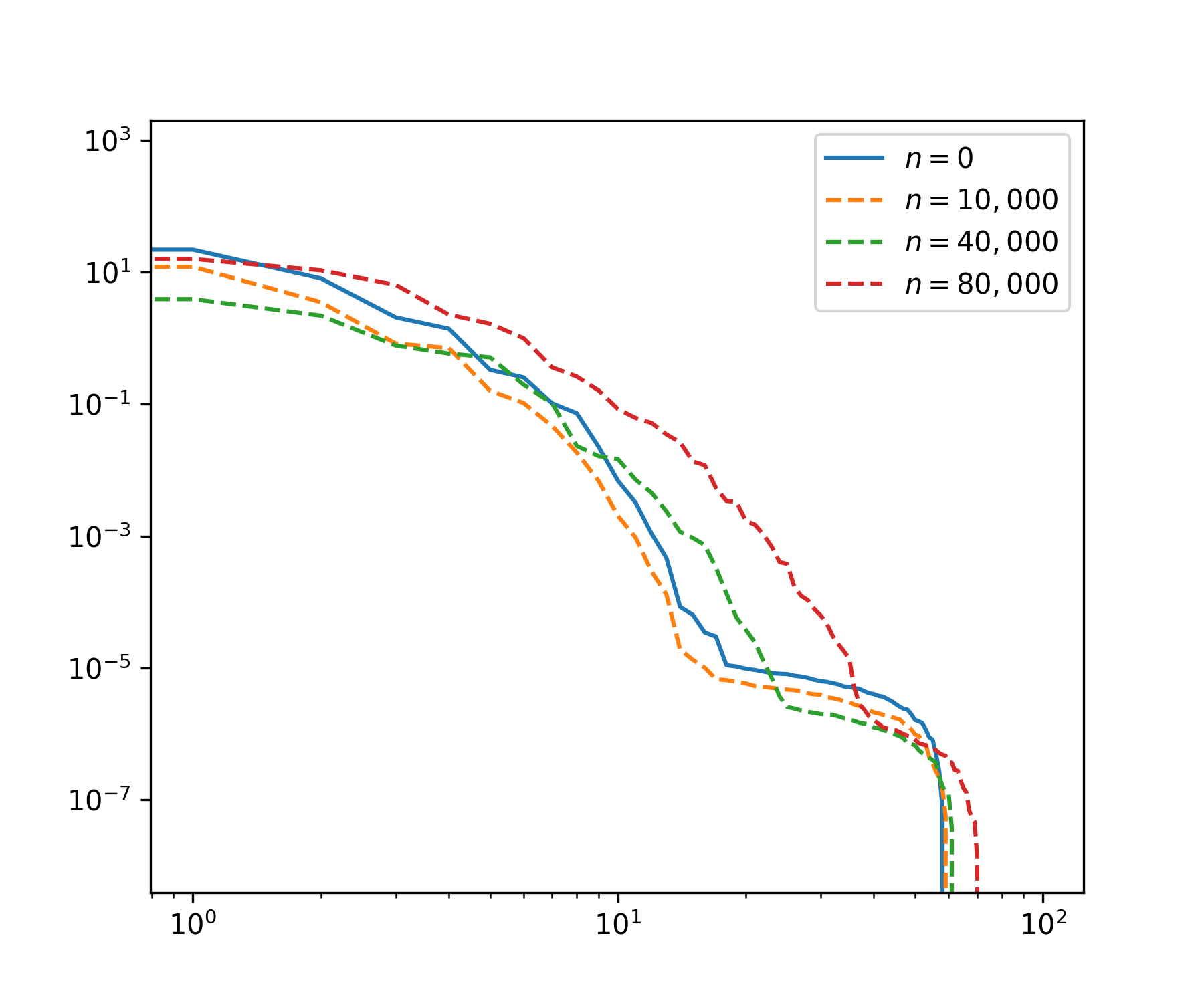}
            \put(50,0.5){\makebox(0,0){Index}} 
            \put(-4,20){\rotatebox{90}{Eigenvalue}} 
            \put(30,77){ $\mathbf{K}_{u_{x}}$}
        \end{overpic} &
        \begin{overpic}[height=0.13\textheight]{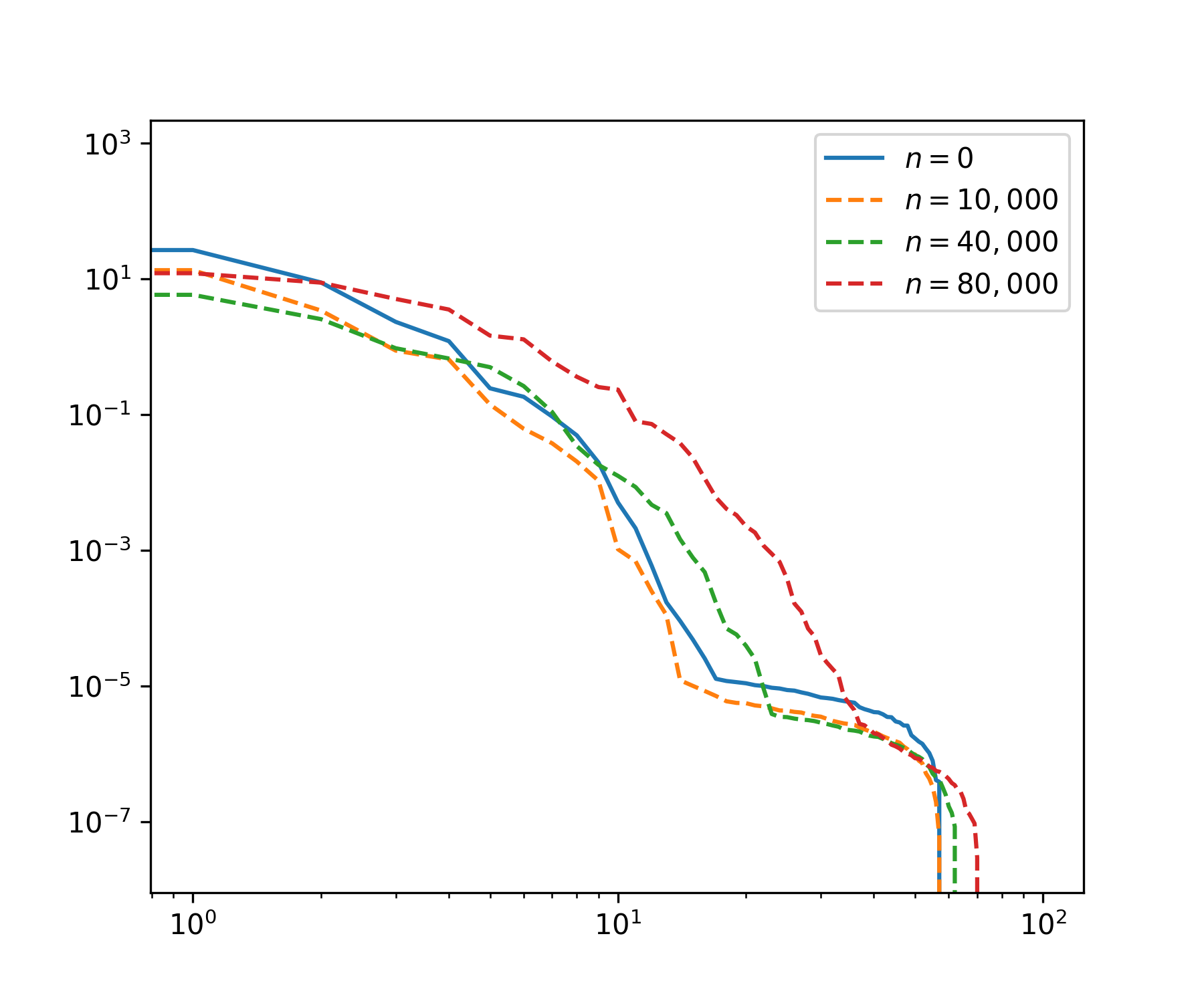}
            \put(50,0.5){\makebox(0,0){Index}} 
            \put(-4,20){\rotatebox{90}{Eigenvalue}} 
            \put(30,77){ $\mathbf{K}_{u_{y}}$}
        \end{overpic} \\\\
        \raisebox{30pt}{\rotatebox[origin=c]{90}{\textbf{}}} & 
        \begin{overpic}[height=0.13\textheight]{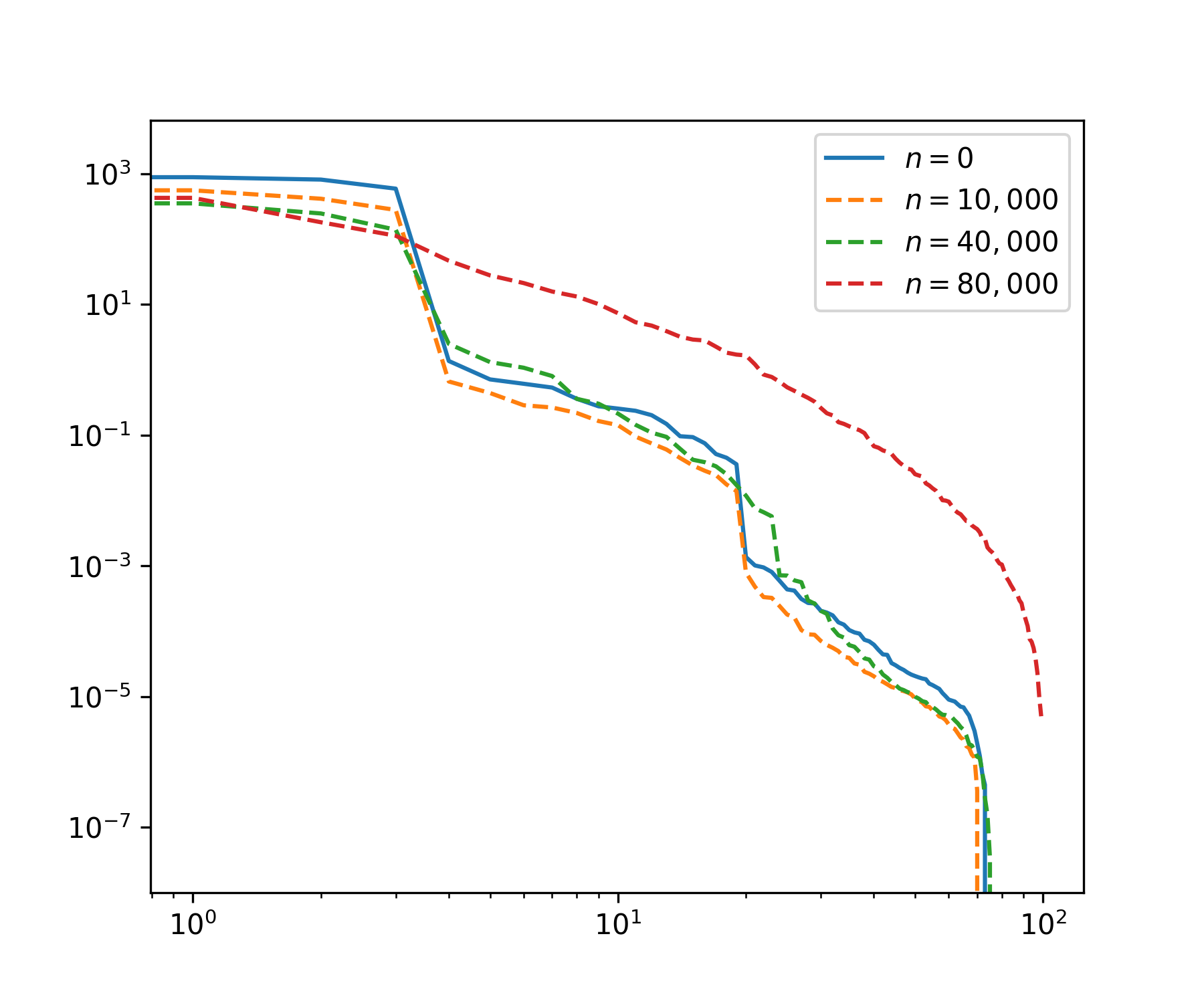}
           \put(50,0.5){\makebox(0,0){Index}} 
            \put(-4,20){\rotatebox{90}{Eigenvalue}} 
            \put(30,77){$\mathbf{K}_{zz}$}
        \end{overpic} &
        \begin{overpic}[height=0.13\textheight] 
        {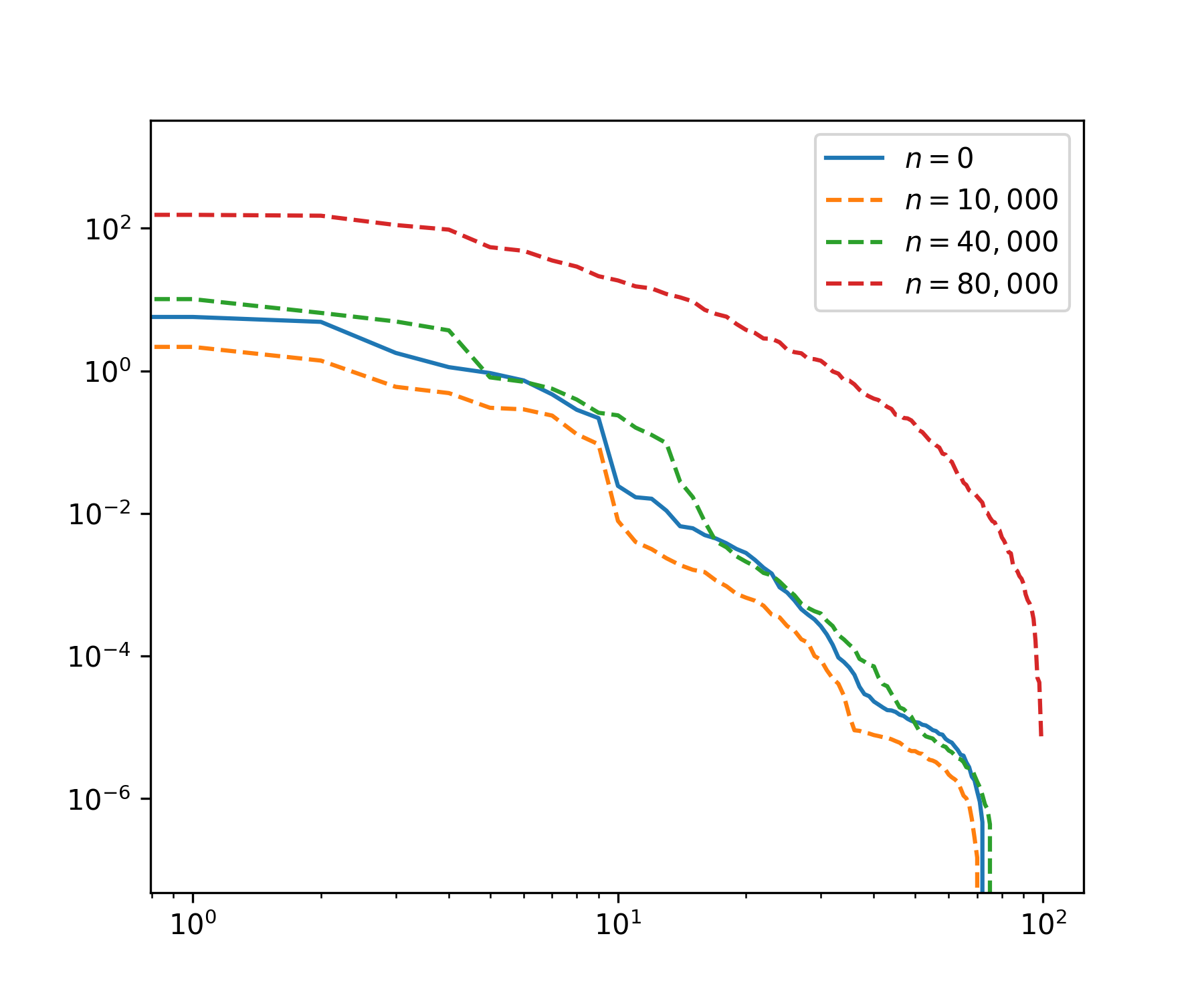}
            \put(50,0.5){\makebox(0,0){Index}} 
            \put(-4,20){\rotatebox{90}{Eigenvalue}} 
            \put(30,77){$\mathbf{K}_{rr}$}
        \end{overpic} 
    \end{tabular}
   \caption{\textbf{Eigenvalues in a forward context of $\mathbf{K}_{i}$,  $\mathbf{K}_{u_{x}}$,  $\mathbf{K}_{u_{y}}$, $\mathbf{K}_{zz}$,$\mathbf{K}_{rr}$ vs their index with $n$ representing the number of iterations.}}
   \label{fig:eigenvaluesforward}
\end{figure}
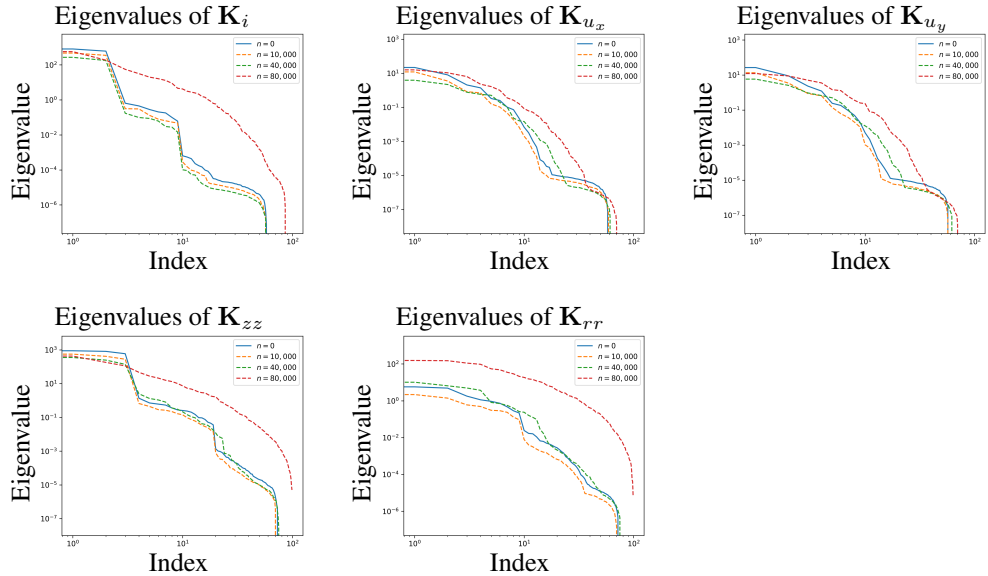

We observe form the Figures \ref{fig:eigenvaluesinverse} and \ref{fig:eigenvaluesforward} in a context of an inverse problem, at initialization, the residual dominates the data losses while we observe the opposite when solving the ADE in the forward context. In addition to the losses on the residual, initial and boundary conditions in the forward problem, we also added a data loss $\mathcal{L}_{z}(\pmb{\theta_{u}})$. Despite having 15 observations of $u$ in the forward case with up to 80000 iterations, we observe a very slow convergence of the residual as well as the data misfit term while their convergence is faster in the inverse context. This  simply means that when solving the forward problem, the algorithm first learns a solution that fits the initial and boundary conditions before the PDE residual is reduced. In the inverse problem case, the solution matching the PDE is learned first before it is constrained to fit the available data. We view this empirical observation as evidence that the NTK adaptation approach is better suited for inverse problems as it was also addressed by \cite{wang2022and}.

\subsubsection{
Accumulative observations
}
\begin{figure}[htp]
    \centering
    \setlength{\tabcolsep}{5pt}
    \renewcommand{\arraystretch}{1.2}
    
    \begin{tabular}{c c c c c}
       \raisebox{30pt}{\rotatebox[origin=c]{90}{\textbf{$4$ obs}}} & 
        \begin{overpic}[height=0.09\textheight] {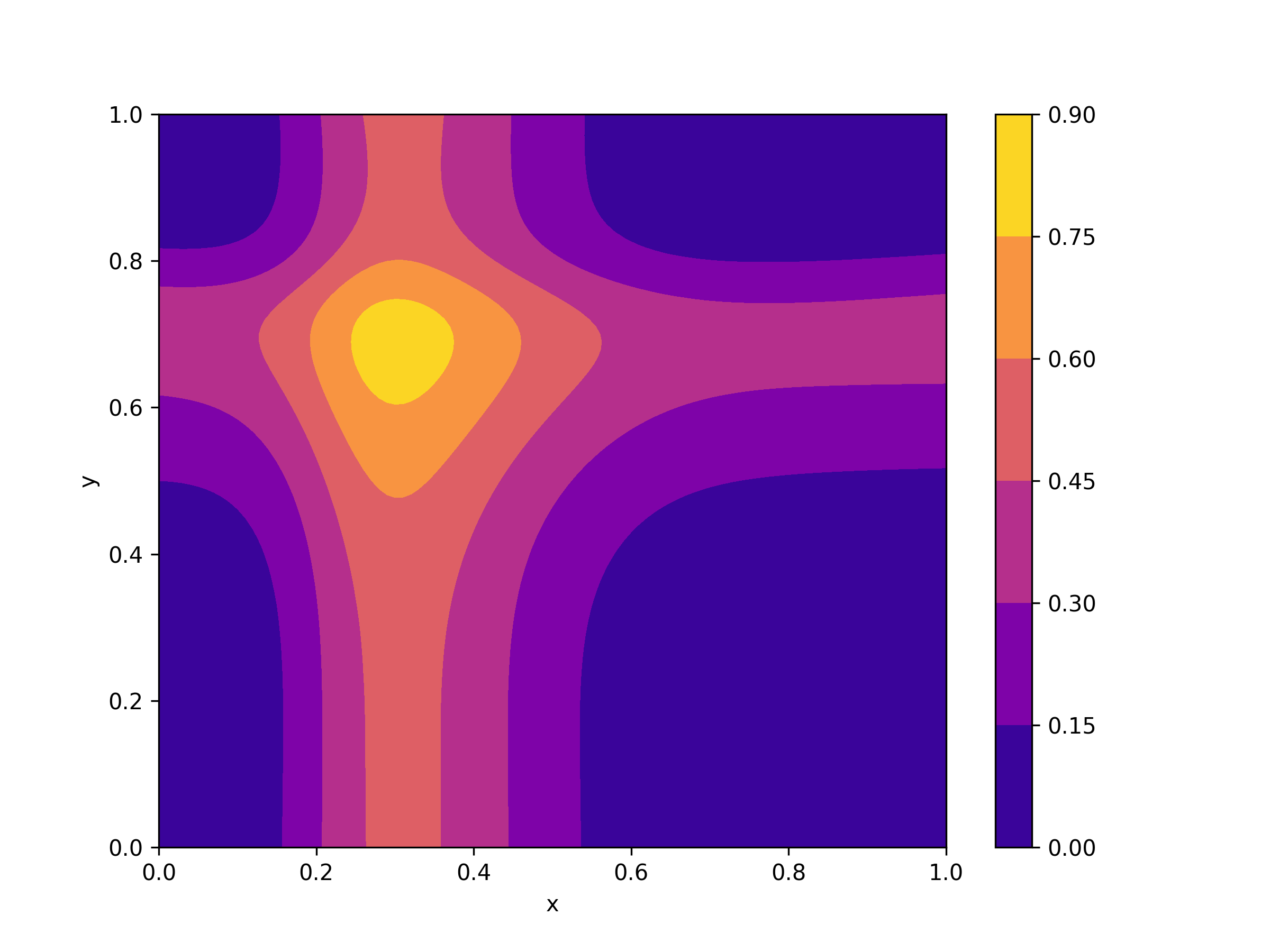}
            \put(20,74){\textbf{Predicted $u$}}
        \end{overpic} &
        \begin{overpic}[height=0.09\textheight]{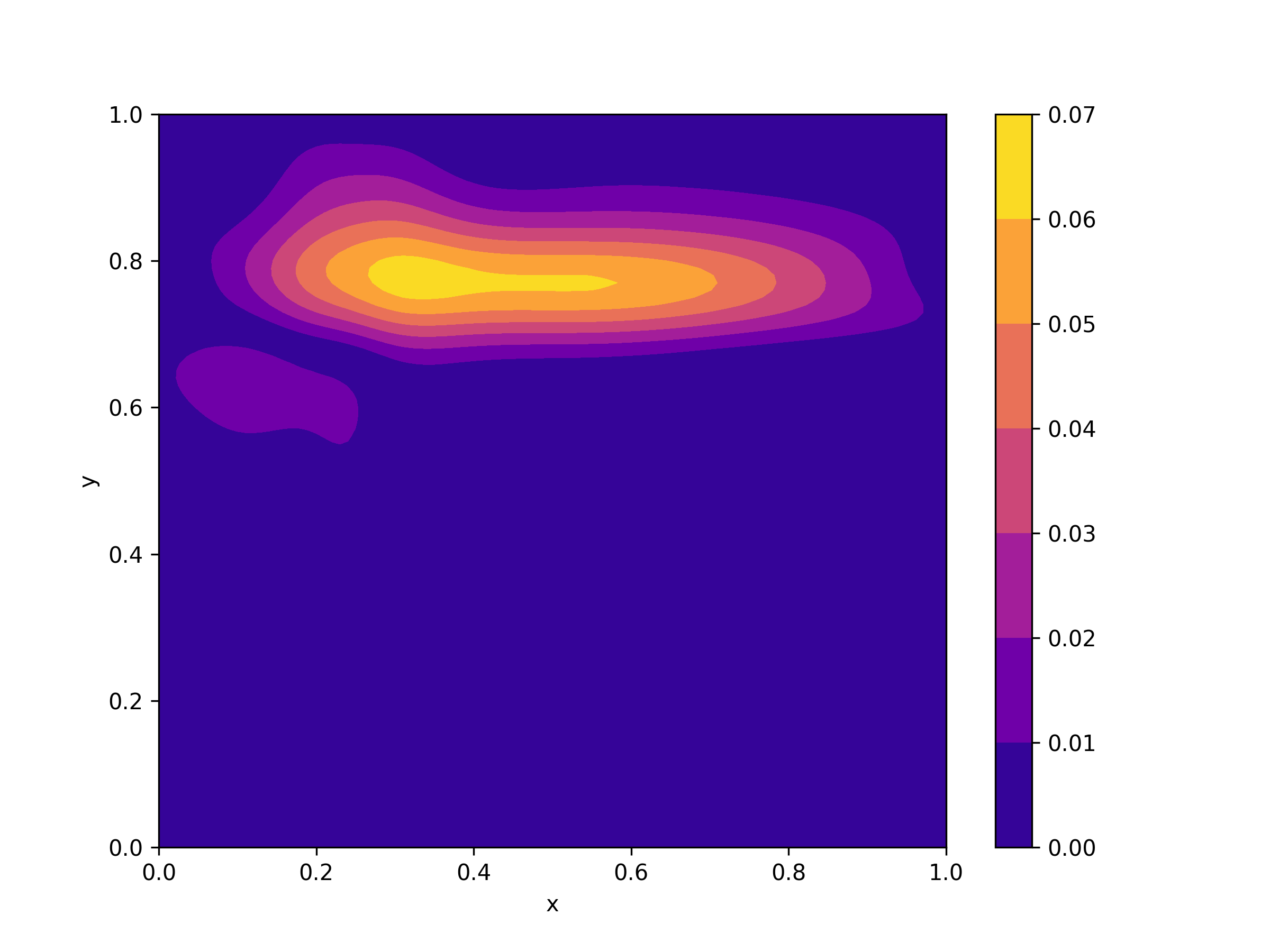}
            \put(10,74){\textbf{Error on $u$ }} 
        \end{overpic} &
        \begin{overpic}[height=0.09\textheight]{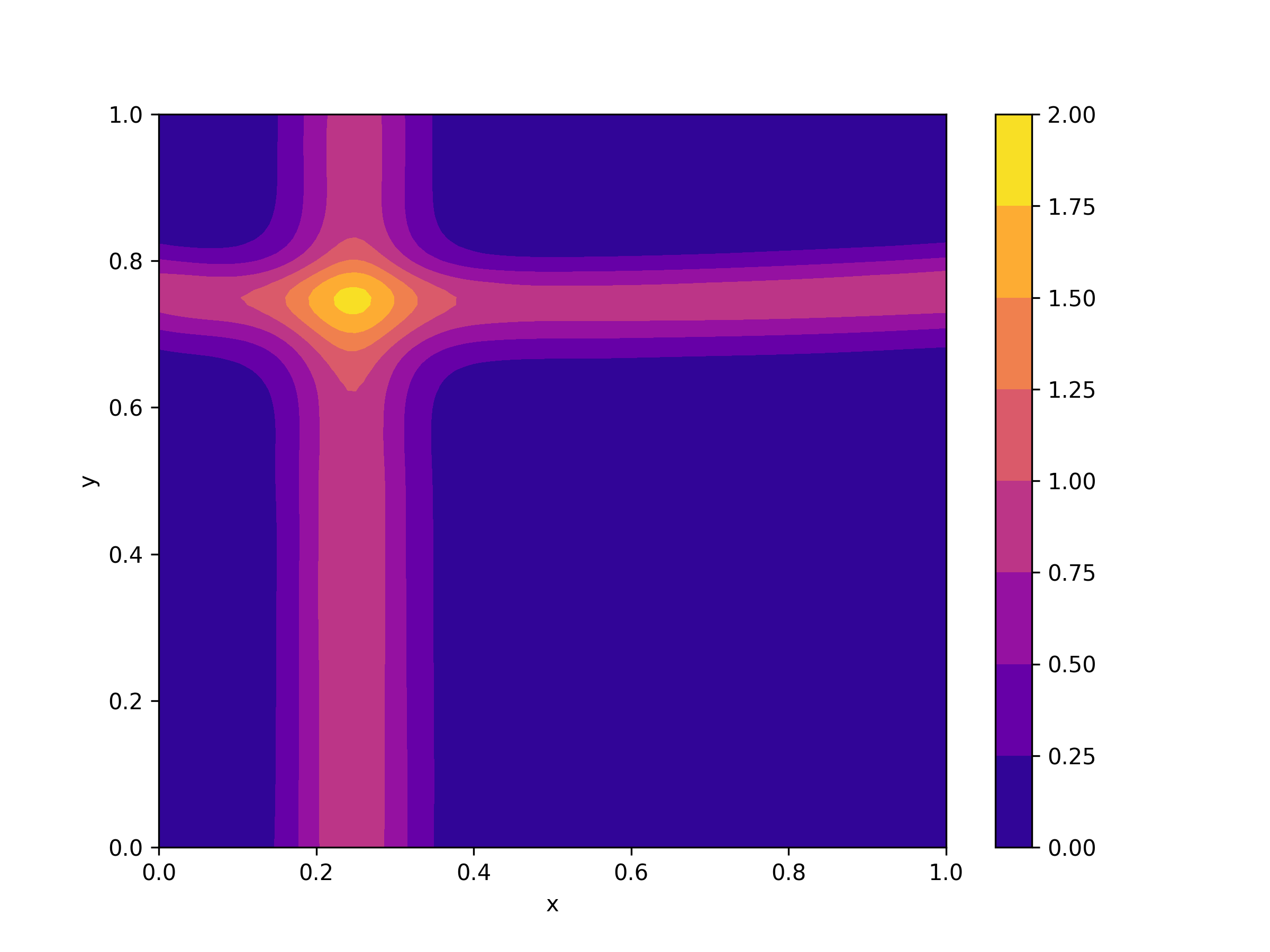}
            \put(20,74){\textbf{Predicted $f$}}
        \end{overpic} &
        \begin{overpic}[height=0.09\textheight]{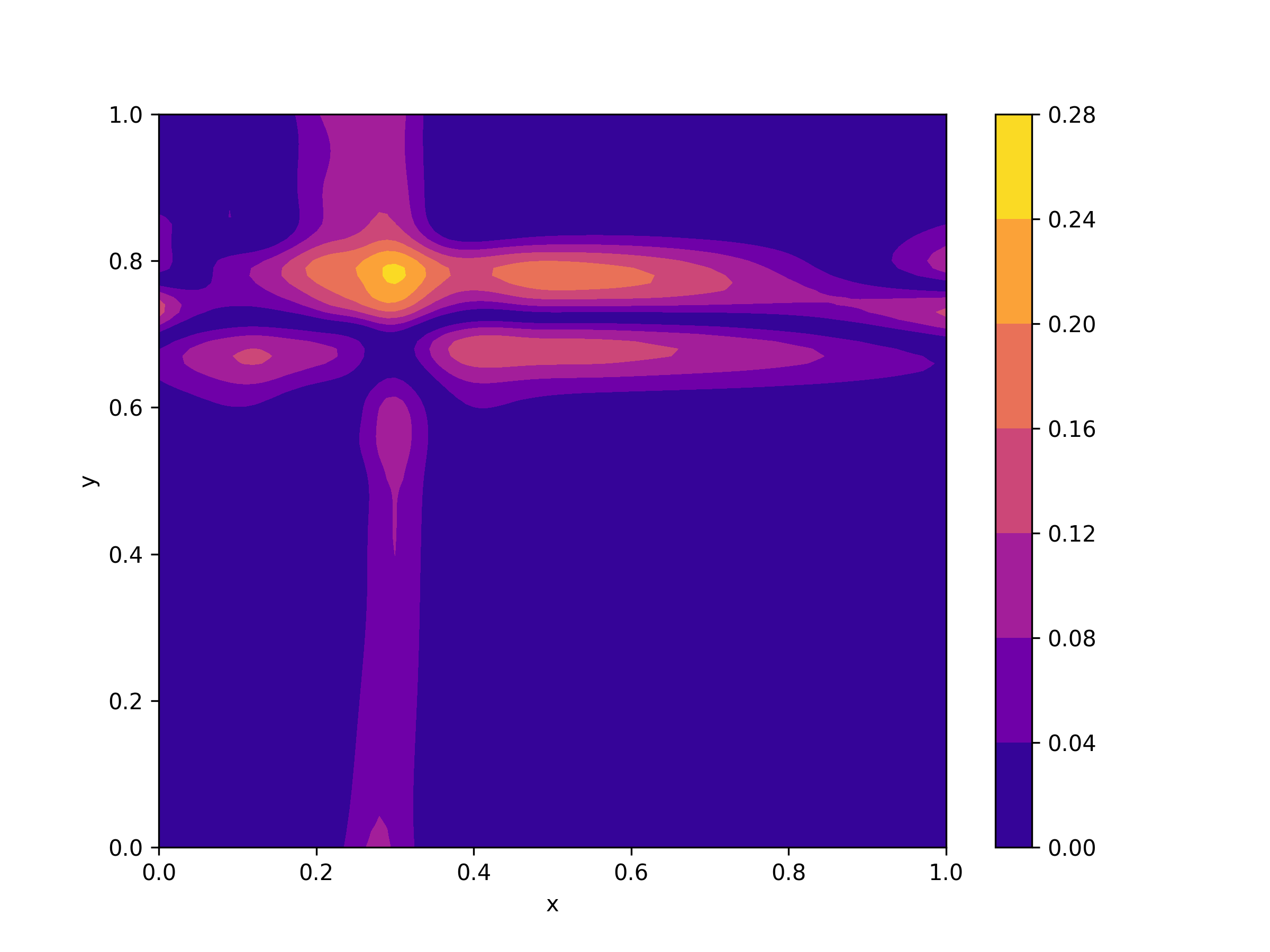}
            \put(20,74){\textbf{Error on $f$}}
        \end{overpic}\\
        \raisebox{30pt}{\rotatebox[origin=c]{90}{\textbf{$15$ obs}}} & 
        \begin{overpic}[height=0.09\textheight] 
        {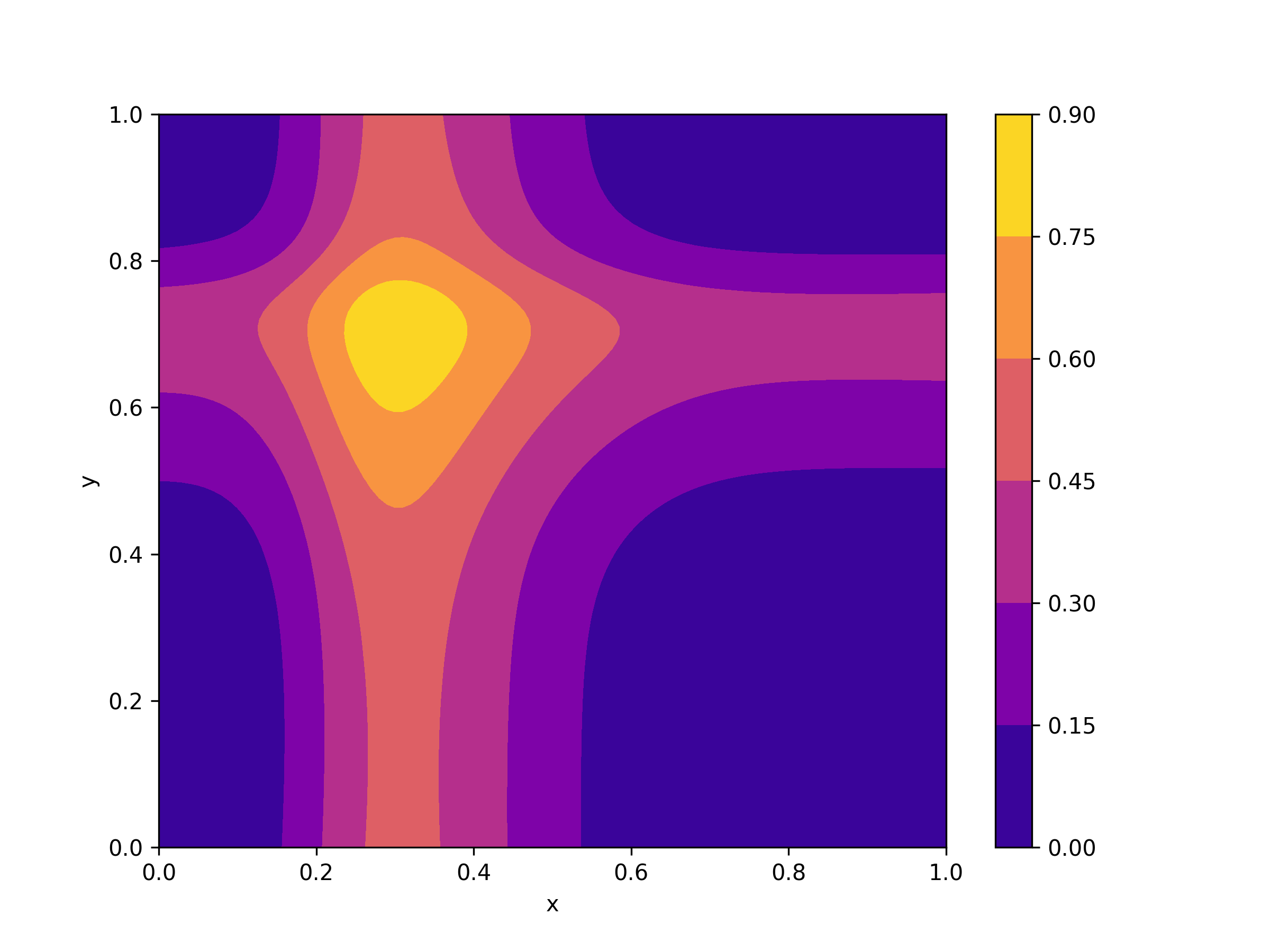}
        \end{overpic} &
        \begin{overpic}[height=0.09\textheight]
        {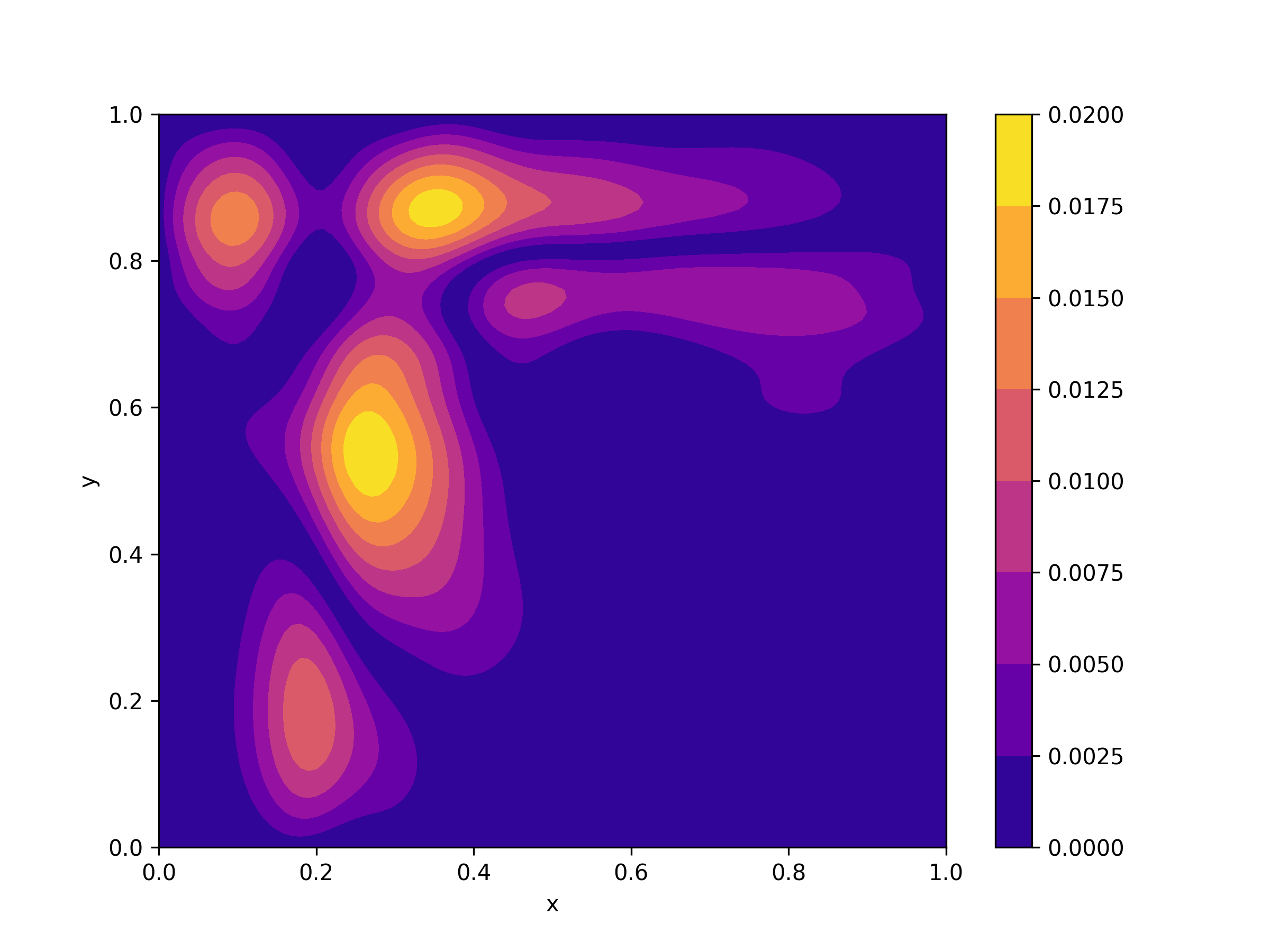}
        \end{overpic} &
        \begin{overpic}[height=0.09\textheight]%
        {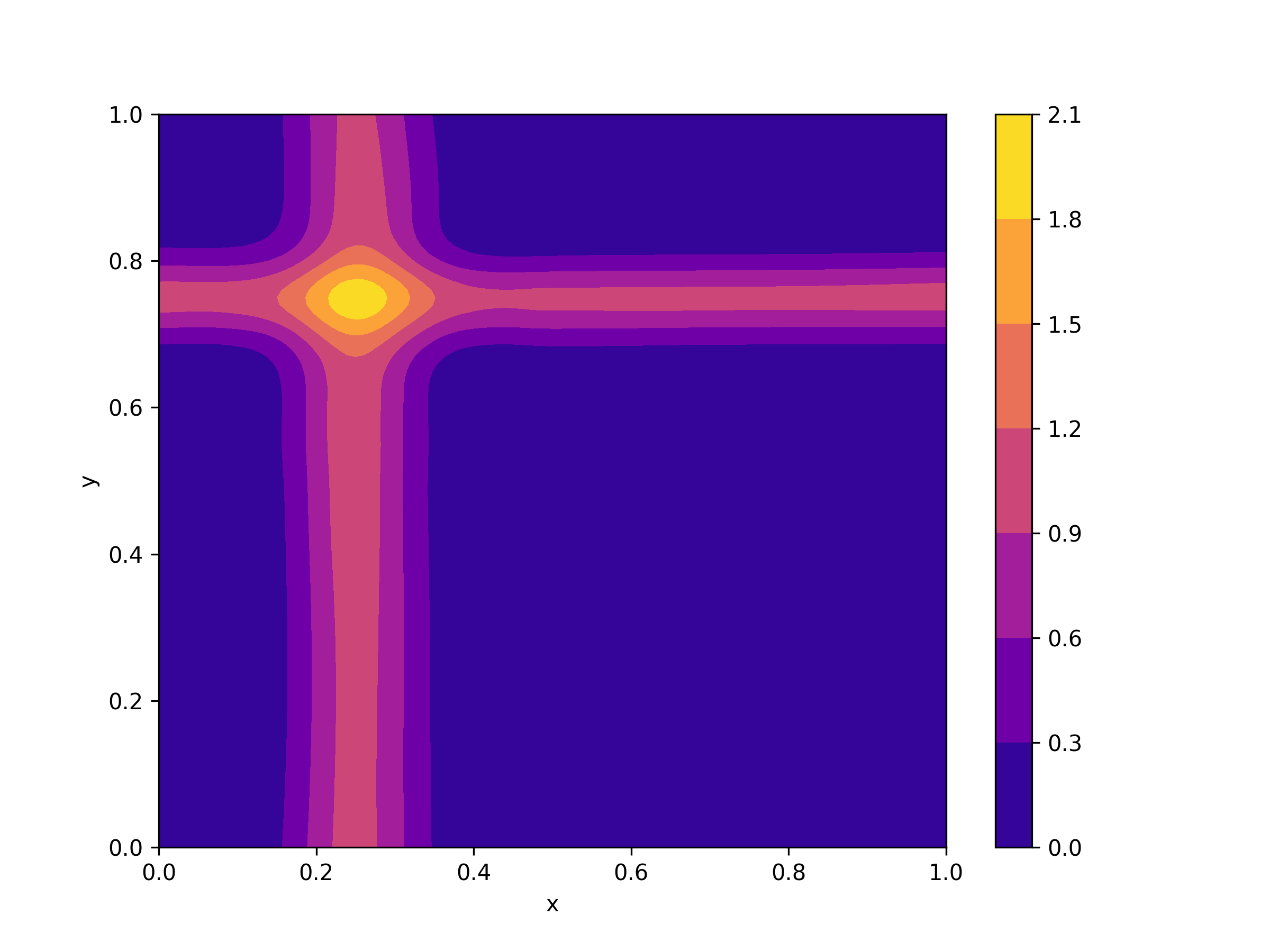}
        \end{overpic} &
        \begin{overpic}[height=0.09\textheight]
        {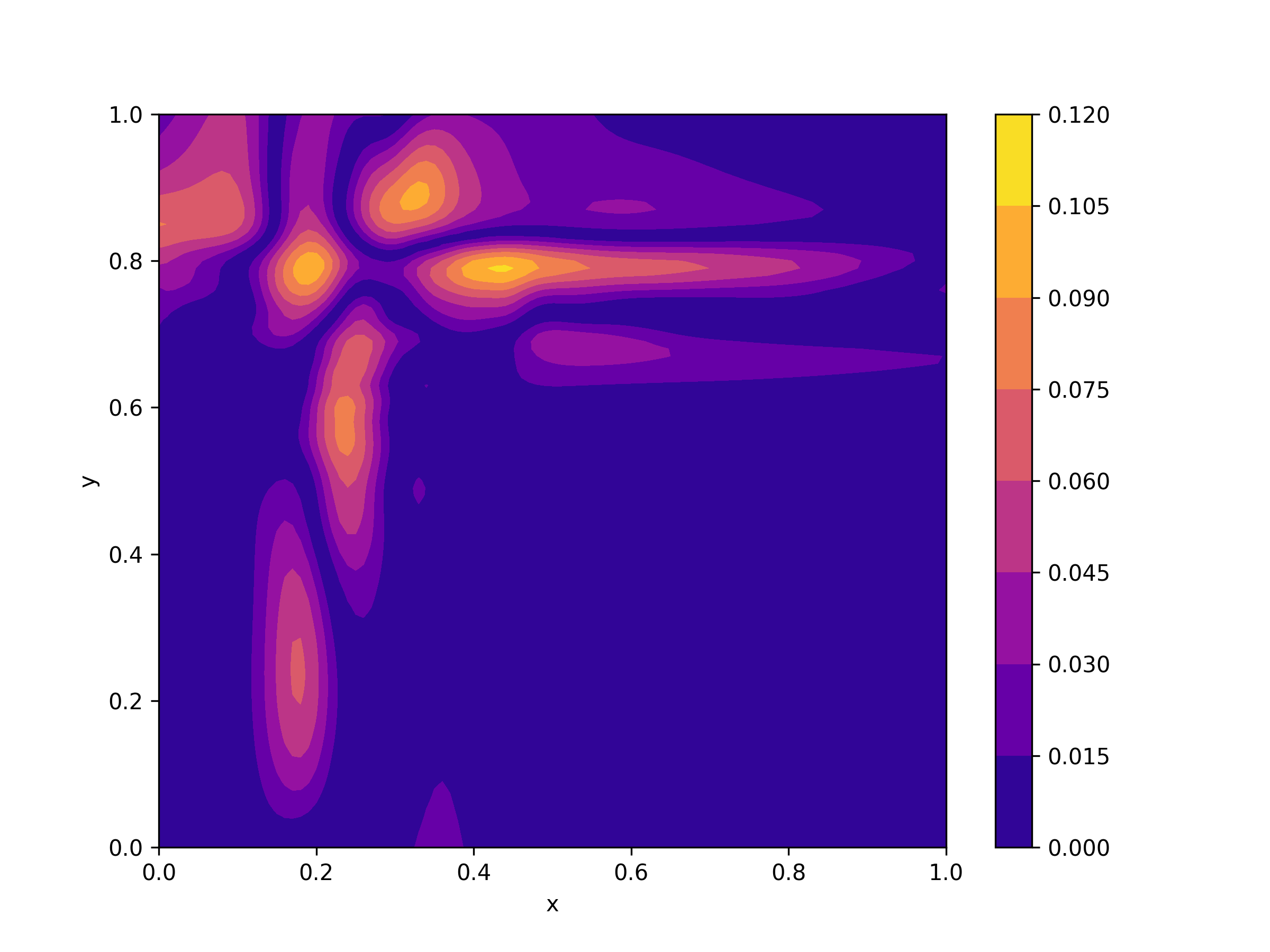}
        \end{overpic}
    \end{tabular}
    
    \caption{\textbf{Prediction and error of $u$ and $f$ with $1\%$ of noise and average observation operator over $\delta t=30$.}}
    \label{fig:1percAVGsolSourceError}
\end{figure}

In this scenario, we assume that at each measurement locations, we have the average of the solution $u$ over a fixed time interval of length $\delta t=30$. One can think of a scenario in which sensors are calibrated to capture air pollutants concentration over 30 seconds time interval, each concentration values representing the average of concentration over the corresponding time interval.\\ 
We have conducted the same experiments as in the previous scenario, with 4 and 15 measurement locations, using the same architectures for neural networks on $u$ and $f$, with positivity constraints imposed on $D$ and $f$.
The PINNs' model performance has slightly decreased in this case, compared to the scenario with pointwise observations, since the amount of measurement data has reduced. We observe from the  Figures \ref{fig:1percAVGsolSourceError} and \ref{ErrorAVG} that, our method produces good approximations of the solution, the source function and the PDE parameters despite the limited number of measurements data, the high ill-posedness of the problem as well as the noise level variation. However, we notice from the Table \ref{tab4AVGObs} a very poor approximation of the diffusion coefficient with 4 observations at $10\%$ of noise. This is due to the insufficient number of measurements data necessary for it approximation, suggesting that in this case the model might be trying to approximate $D$ from the noise since the chosen measurements locations are not well spatially distributed for a good coverage of the spatial domain. An increase of the number of measurements locations makes the model to give a better approximation of $D$ as expected, and as shown in Table \ref{tab4AVGObs}.

\begin{table}[htp]
\centering
\begin{tabular}{|c|c|c|c|c|c|c|}
    \cline{1-7} 
    \textbf{4 Obs}
     & \multicolumn{2}{c|}{\textbf{Vx}} 
     & \multicolumn{2}{c|}{\textbf{Vy}} 
     & \multicolumn{2}{c|}{\textbf{D}} \\
    \hline
    \textbf{Noise} &\textbf{ Predicted} &\begin{tabular}[c]{@{}c@{}}\textbf{Relative}\\\textbf{error}\end{tabular}  & \textbf{Predicted}  &\begin{tabular}[c]{@{}c@{}}\textbf{Relative}\\\textbf{error}\end{tabular}   & \textbf{Predicted}  &\begin{tabular}[c]{@{}c@{}}\textbf{Relative}\\\textbf{error}\end{tabular}  \\
   \hline
    $1\%$ & 0.19912  &0.44\%  & -0.20525 & 2.62\% & 0.008643 & 13.57\% \\
    \hline
    $5\%$ & 0.19469 & 2.65\% & -0.15938 & 20.31\% & 0.00538 & 46.2\% \\
    \hline
    $10\%$ & 0.18404 & 7.98\% & -0.23356 & 16.78\% & 0.0021 & 79\% \\
    \hline
    \cline{1-7} 
    \textbf{15 Obs}
     & \multicolumn{2}{c|}{\textbf{Vx}} 
     & \multicolumn{2}{c|}{\textbf{Vy}} 
     & \multicolumn{2}{c|}{\textbf{D}} \\
    \hline
    \textbf{Noise} & \textbf{Predicted} &\begin{tabular}[c]{@{}c@{}}\textbf{Relative}\\\textbf{error}\end{tabular}  & \textbf{Predicted}  & \begin{tabular}[c]{@{}c@{}}\textbf{Relative}\\\textbf{error}\end{tabular}  & \textbf{Predicted}  &\begin{tabular}[c]{@{}c@{}}\textbf{Relative}\\\textbf{error}\end{tabular}  \\
   \hline
    $1\%$ & 0.19713  &1.44\%  & -0.19977 & 0.11\% & 0.009849 & 1.53\% \\
    \hline
    $5\%$ & 0.19009 & 4.95\% & -0.1942522 & 2.87\% & 0.007424 & 25.76\% \\
    \hline
    $10\%$ &  0.18095 & 9.52\% & -0.20362 & 1.81\% & 0.005604 & 43.96\% \\
    \hline
\end{tabular}
\caption{\textbf{Predicted values of the velocities and diffusion coefficients with their relative error, assuming 4 and 15 observations locations with an average observation operator on $u$ over time intervals of length $\delta t$=30.}}
\label{tab4AVGObs}
\end{table}

\subsection{2D ADE with non-constant velocity and diffusion coefficients}
\label{sec:variable-coeff}
In this section we consider the same 2D advection diffusion equation 
\eqref{ade}, with a time-dependent velocity vector, and a time-independent 
diffusion coefficient. We consider the same source function as in 
(\ref{sourcefunction}), and we solve the problem using Finite Element Method 
with $V(t)=(V_{1}(t), V_{2}(t))= (0.2+0.1\sin(2\pi t), -0.2-0.1\cos(2\pi t)$  
and $D(x,y)= 0.1(1+\sin(\pi x)\cos(\pi y)/2)$. We randomly select 30 locations 
at which we assume we have noisy observations $\mathbf{z}^\ast$ and velocity 
data. We used 4 separate Neural Networks  $u(\mathbf{x},t,\pmb{\theta}_{u}), 
f(\mathbf{x},\pmb{\theta}_{f})$,\\
$V(t,\pmb{\theta}_{V}),D(\mathbf{x},\pmb{\theta}_{D})$ to approximate $u,f, V$ and $D$ respectively. In addition to the PDE constraint, we impose a positivity constraint on the source $f$ in the PINN model, by parameterizing it using an exponential transformation. 

\begin{figure}[htp]
    \centering
    \includegraphics[width=.7\textwidth]{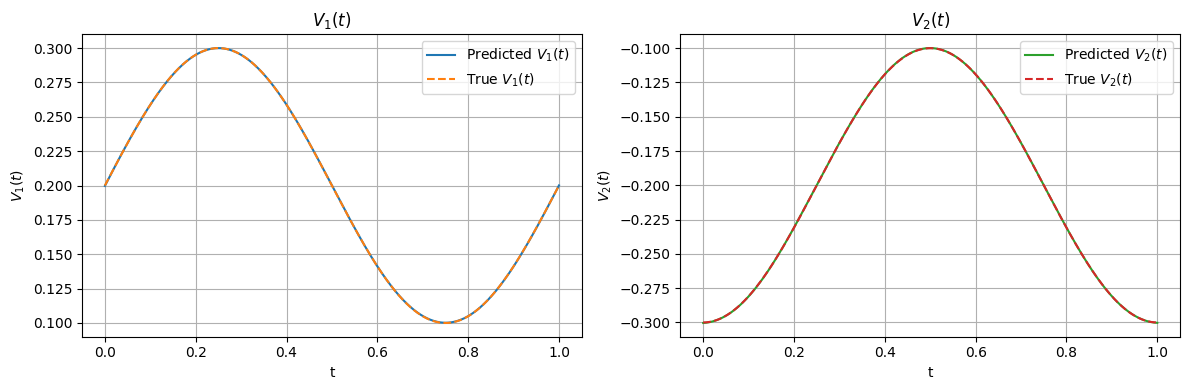}
    \caption{\textbf{Plot of predicted vs exact velocity component over time.}}
    \label{fig:predicted_Velocityfunction}
\end{figure}

\begin{figure}[htp]
    \centering
    \setlength{\tabcolsep}{5pt}
    \renewcommand{\arraystretch}{1.2}
    
    \begin{tabular}{c c c c }
        \begin{overpic}[width=0.3\textwidth] {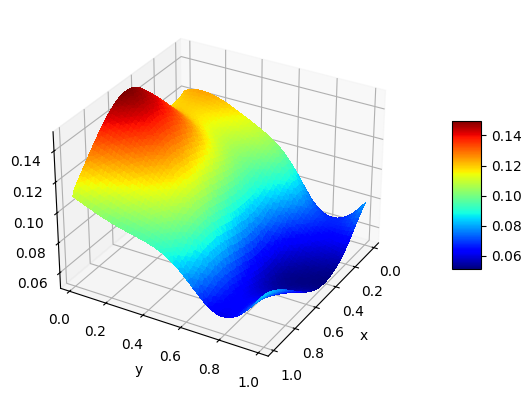}
            \put(20,74){\textbf{Predicted $D$}}
        \end{overpic} &
     \begin{overpic}[width=0.3\textwidth] {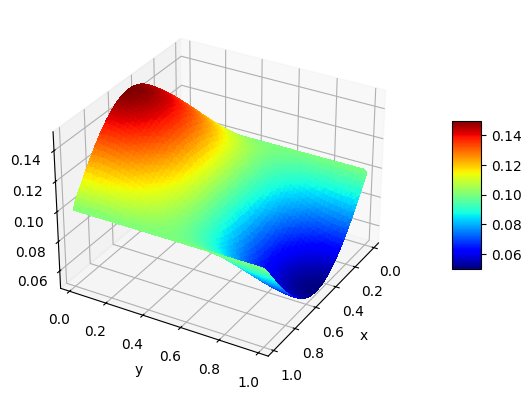}
            \put(20,74){\textbf{Exact $D$}}
        \end{overpic} &
        \begin{overpic}[width=0.3\textwidth] {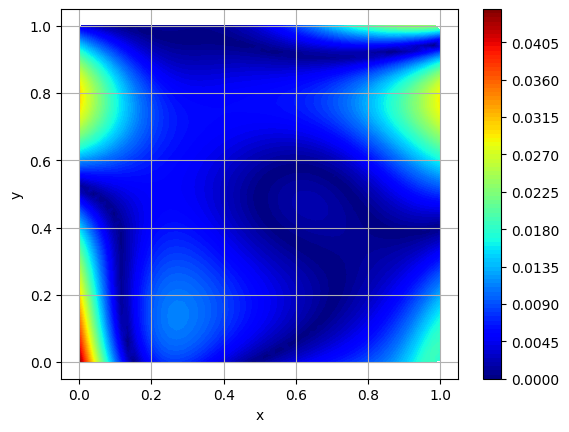}
            \put(10,78){\textbf{Absolute error on $D$}}
        \end{overpic} &
    \end{tabular}
        \caption{\textbf{\textit{Left:} surface plots of the predicted diffusion.  \textit{Middle:} exact diffusion \textit{Right:} Contour plot of the absolute error on the diffusion.}}
    \label{fig:predicted_Difffunction}
\end{figure}

\begin{figure}[htp]
    \centering
    \setlength{\tabcolsep}{5pt}
    \renewcommand{\arraystretch}{1.2}
    
    \begin{tabular}{c c c c }
        \begin{overpic}[width=0.42\textwidth] {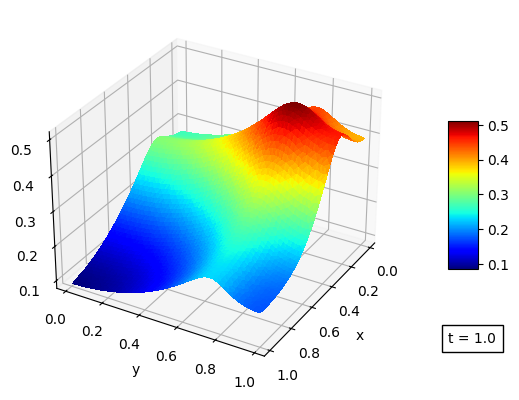}
            \put(40,74){\makebox(0,0){\textbf{Predicted solution $u$}}} 
        \end{overpic} &
     \begin{overpic}[width=0.4\textwidth] {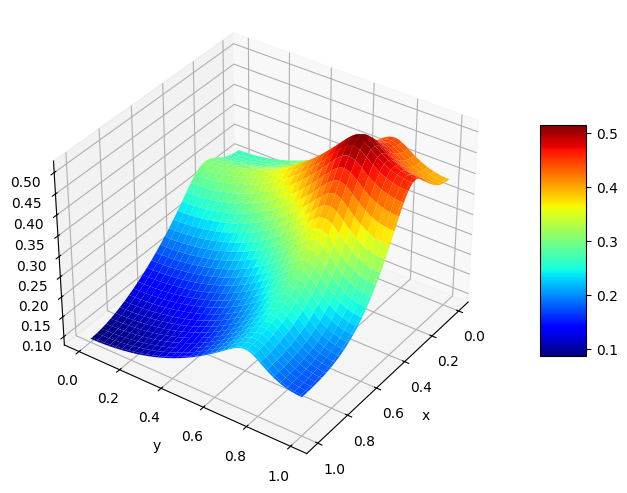}
            \put(40,78){\makebox(0,0){\textbf{FEM solution}}}
        \end{overpic} &
    \end{tabular}
        \caption{\textbf{Surface plots of the predicted vs FEM solution $u$ at $t=1.0$}}
   \label{fig:surfpredicted_Solution}
\end{figure}

\begin{figure}[htp]
    \centering
    \setlength{\tabcolsep}{5pt}
    \renewcommand{\arraystretch}{1.2}
    
    \begin{tabular}{c c c c }
        \begin{overpic}[width=0.3\textwidth] {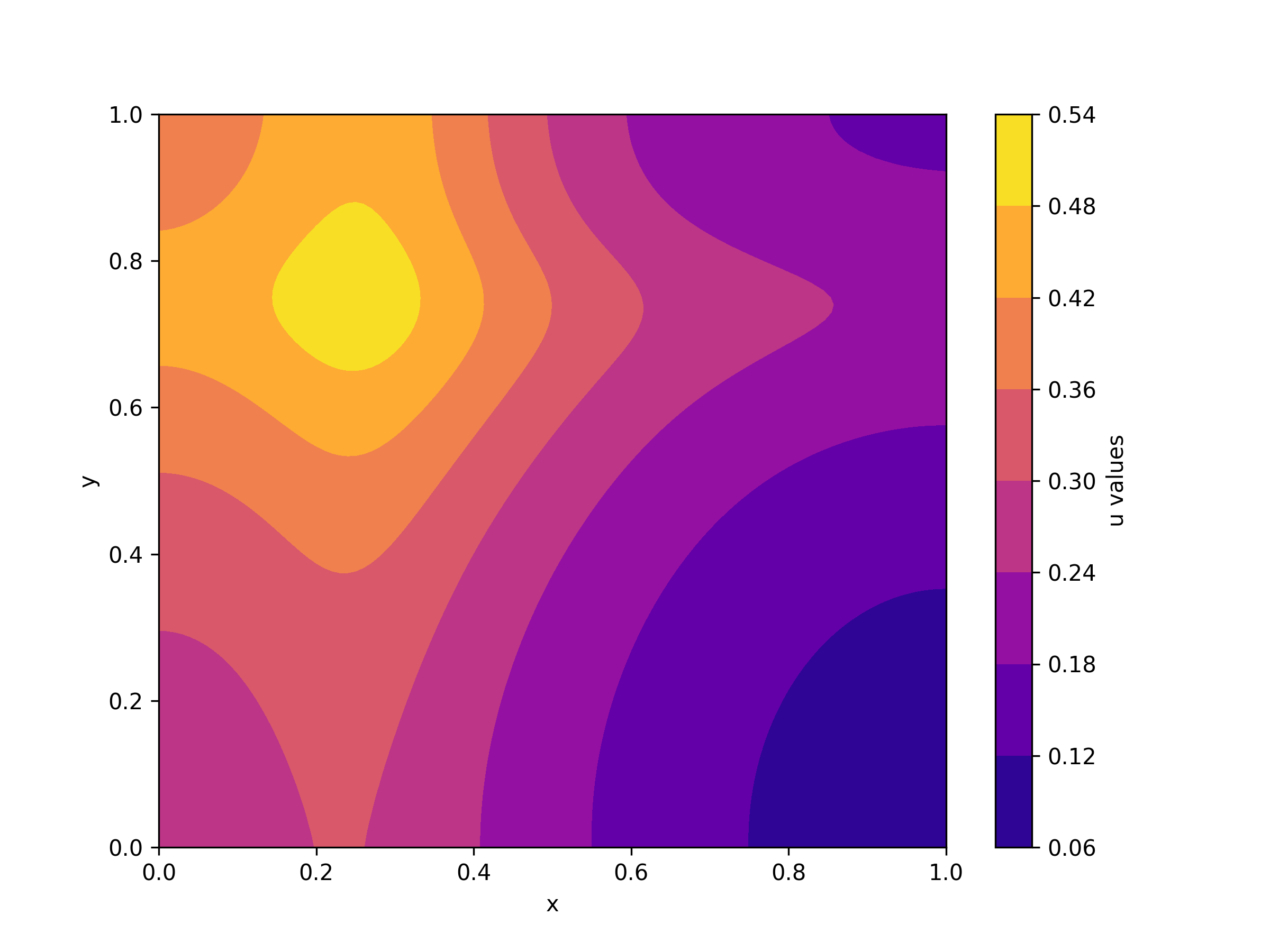}
            \put(18,74){\textbf{Predicted solution}}
        \end{overpic} &
     \begin{overpic}[width=0.3\textwidth] {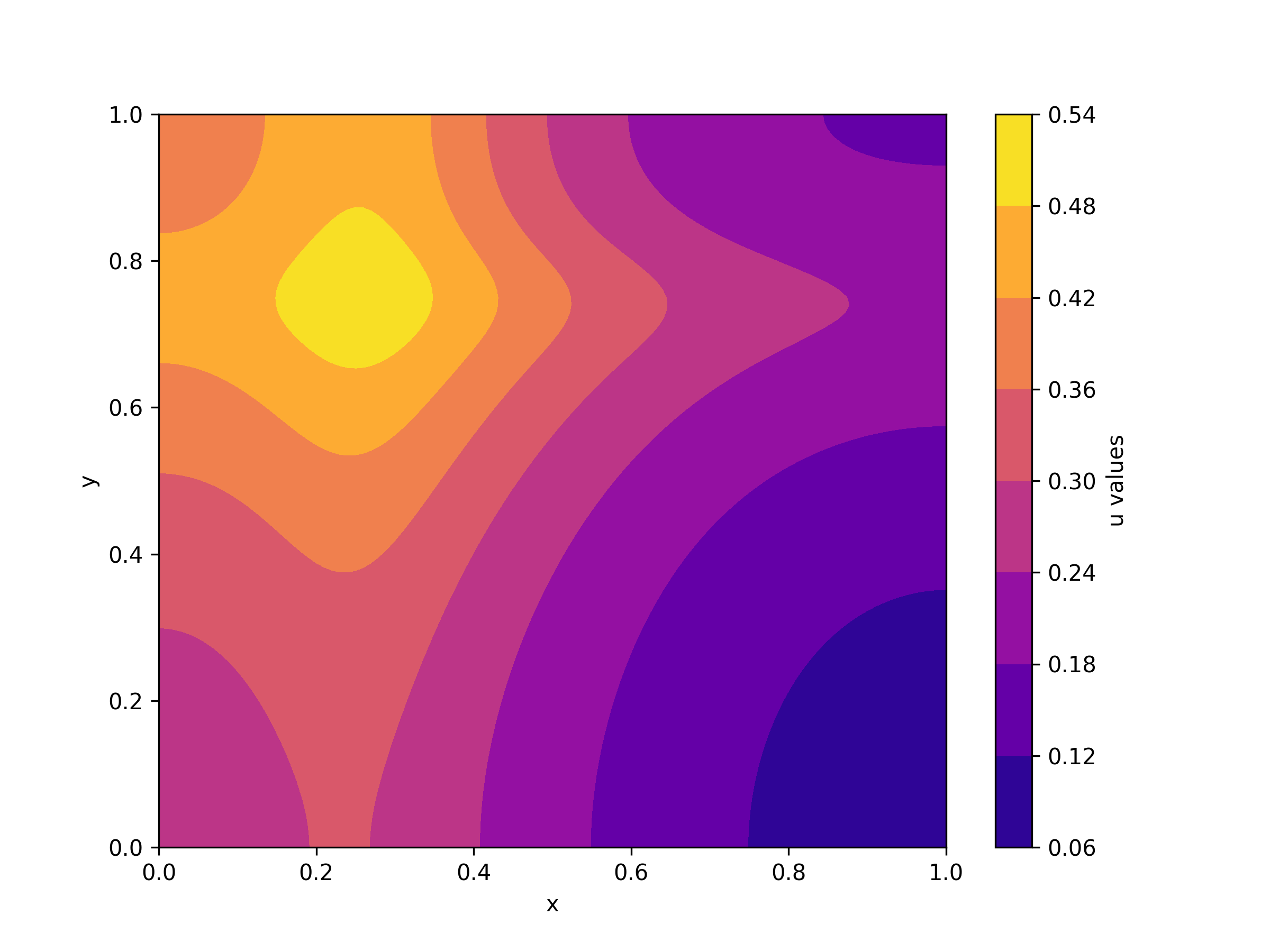}
            \put(20,74){\textbf{FEM solution}}
        \end{overpic} &
        \begin{overpic}[width=0.3\textwidth] {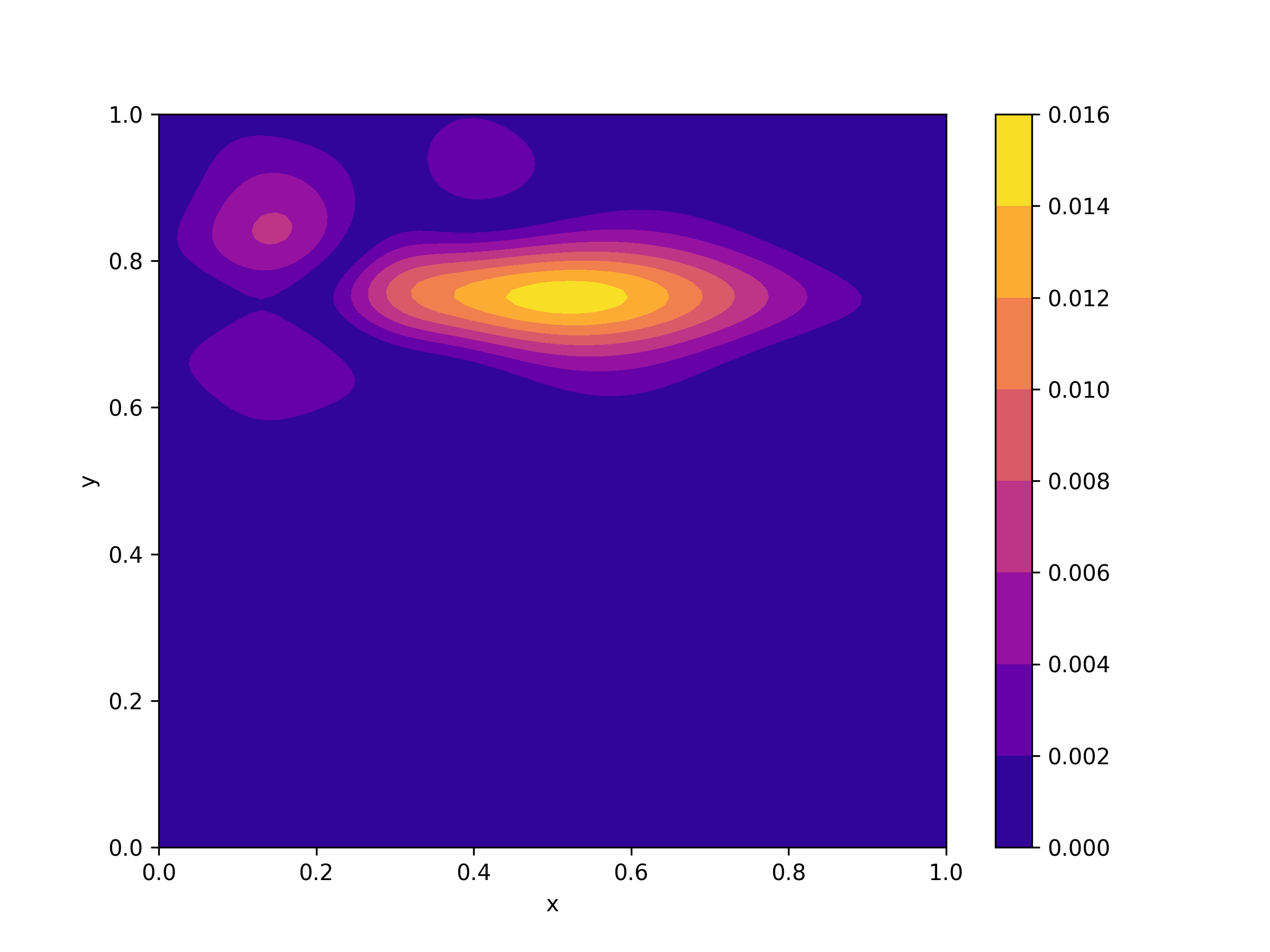}
            \put(12,74){\textbf{Absolute error on $u$}}
        \end{overpic} &
    \end{tabular}
         \caption{\textbf{Contour plots of the predicted (left) vs FEM solution(middle) and the absolute error (right) on $u$ at $t=1$.}}
    \label{fig:Contpredicted_Solution}
\end{figure}

\begin{figure}[htp] 
    \centering
    \setlength{\tabcolsep}{5pt}
    \renewcommand{\arraystretch}{1.2}
    
    \begin{tabular}{c c c c }
        \begin{overpic}[width=0.3\textwidth] {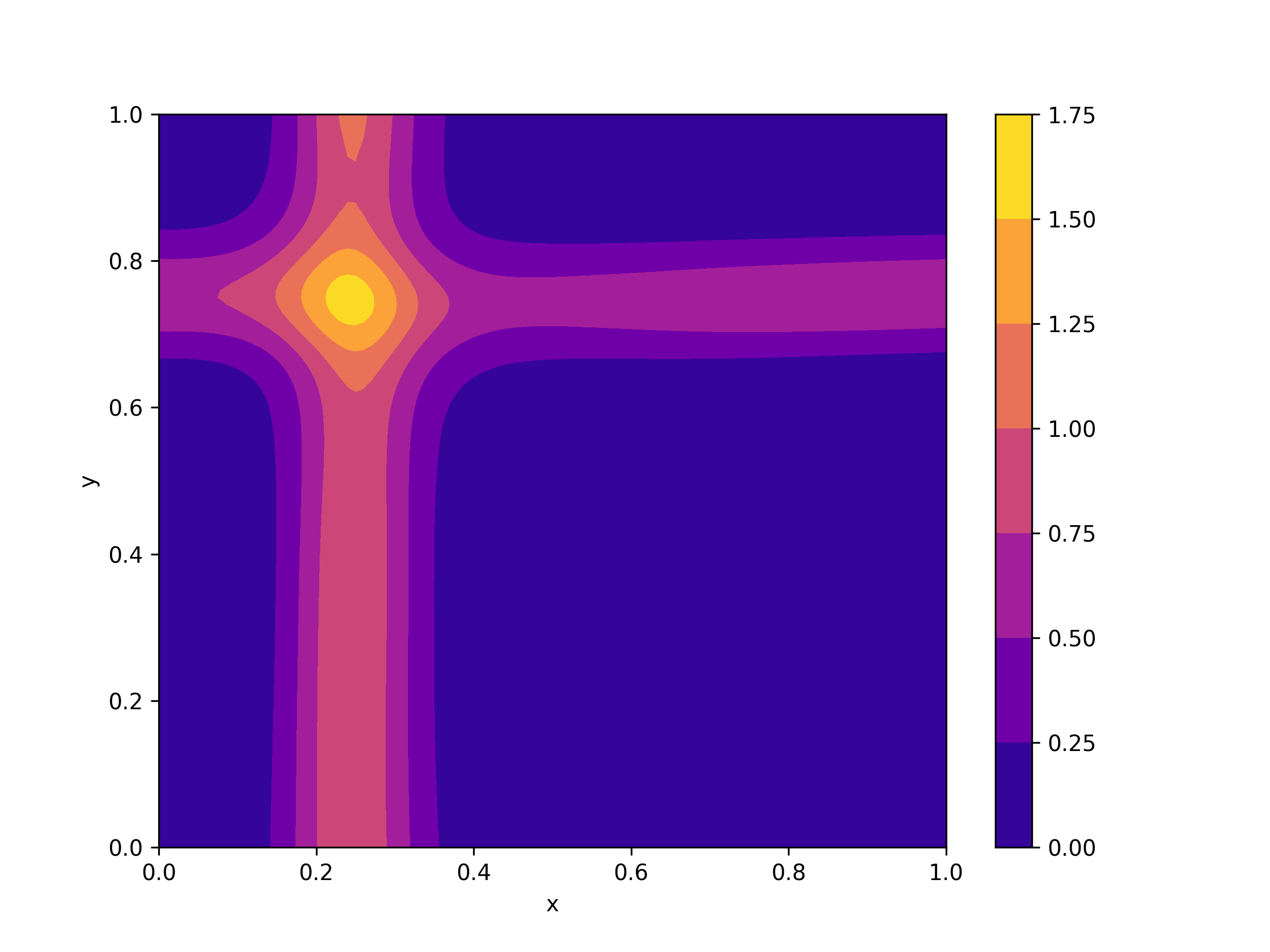}
            \put(20,74){\textbf{Predicted $f$}}
        \end{overpic} &
     \begin{overpic}[width=0.3\textwidth] {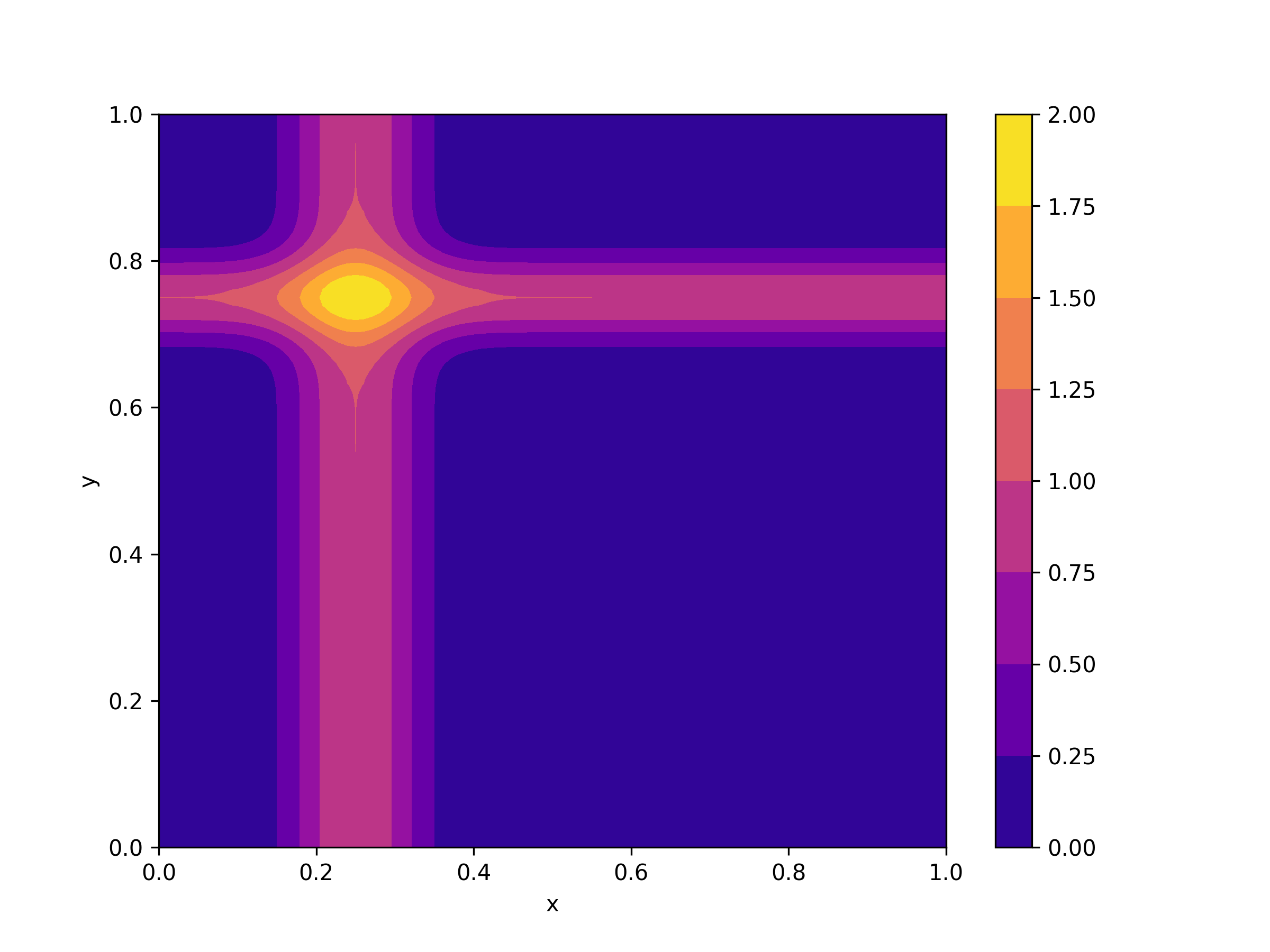}
            \put(20,74){\textbf{Exact $f$}}
        \end{overpic} &
        \begin{overpic}[width=0.3\textwidth] {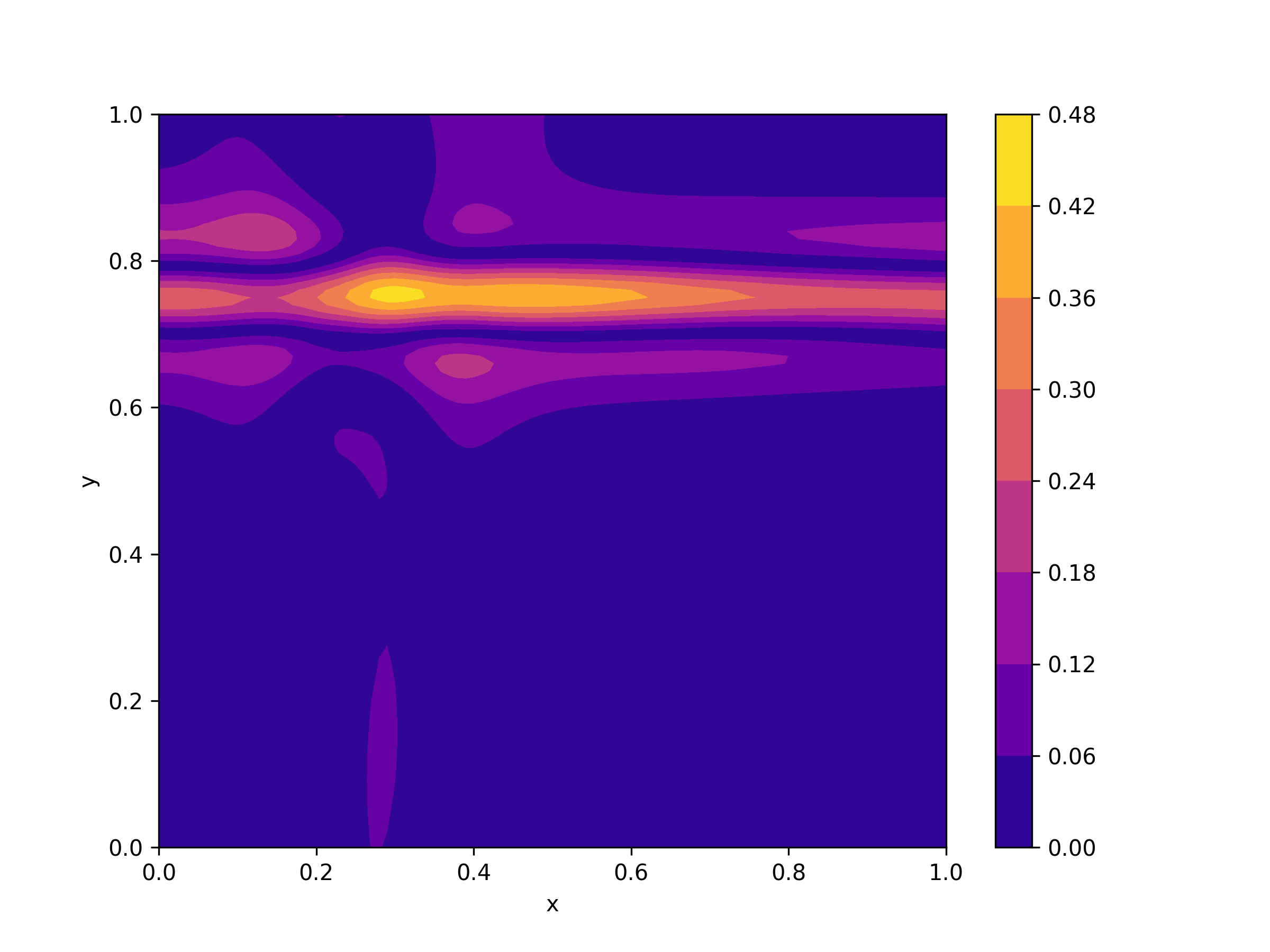}
            \put(22,76){\textbf{Absolute error on $f$}}
        \end{overpic} &
    \end{tabular}
         \caption{\textbf{Contour plots of the predicted (left) vs exact (middle) source function with its absolute error (right).}}
     \label{fig:Contpredicted_Source}
\end{figure}

Once again, we simultaneously trained these neural networks using the weighted adaptive method based on NTK for PINNs while using 20000 iterations on ADAM ( with learning rate $10^{-3}$) followed by 30000 iterations with L-BFGS optimizer  with learning rate $0.1$ .
Although we have a limited number of measurements, the weighted adaptive method on PINNs accurately approximates the velocity, and the solution $u$ as we can observe on  Figure \ref{fig:predicted_Velocityfunction}, \ref{fig:surfpredicted_Solution} and \ref{fig:Contpredicted_Solution} respectively. We observe good approximation of the diffusion coefficients with a Mean Absolute Error (MAE) of $6.45e-03$, a relative $L2$ error of $9.1427e-02$ and larger error values on the absolute point wise error of the diffusion Figure \ref{fig:predicted_Difffunction}, because no boundary on initial condition on $D$ were given to the model. This does not influence the approximation of the source function since, on the Figure \ref{fig:Contpredicted_Source}, the absolute error on $f$ is larger at the locations at which the approximation error of the solution $u$ also yields the highest absolute error value. The small amount of observations on $u$ might not cover enough the spatial domain, making it hard for the PINNs to capture correctly the shape of the diffusion component.

\subsection{3D advection-diffusion equation with height-dependent velocity and diffusion parameters}
\begin{figure}[htp]
    \centering
    \includegraphics[width=0.3\textwidth]{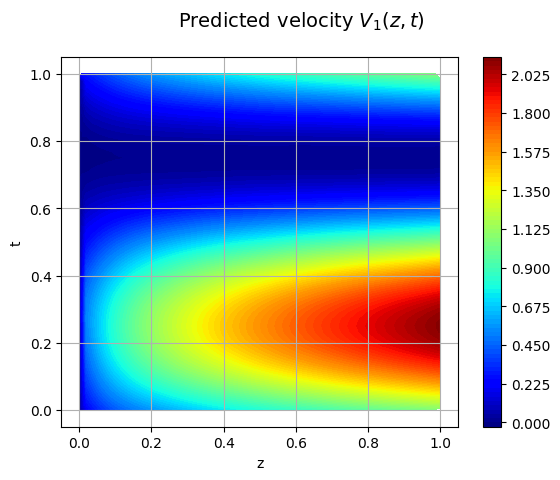}
    \includegraphics[width=0.3\textwidth]{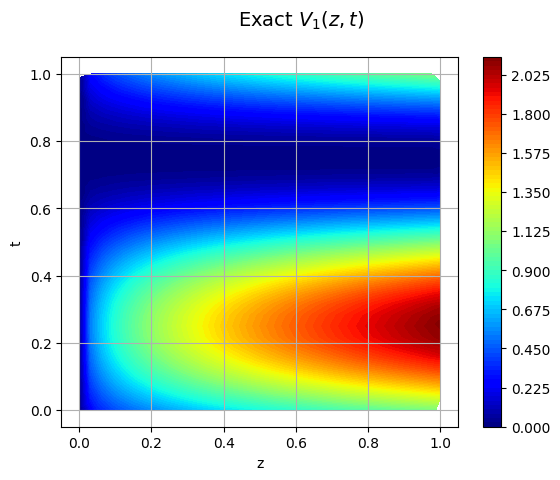}
    \includegraphics[width=0.3\textwidth]{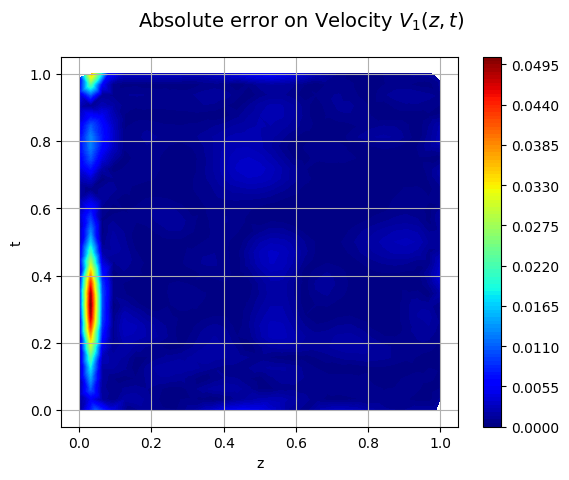}
     \includegraphics[width=0.3\textwidth]{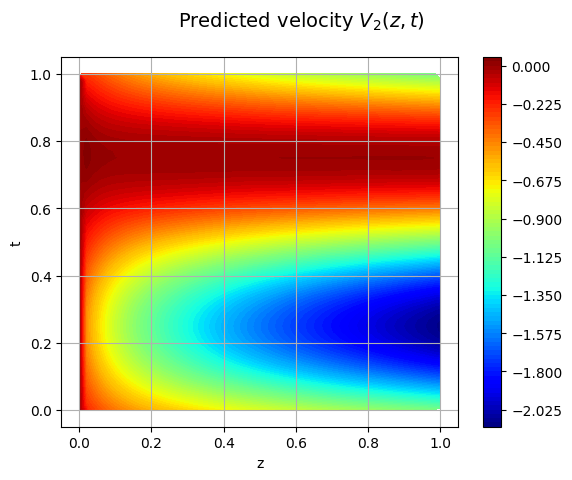}
    \includegraphics[width=0.3\textwidth]{figures/3DpredictedVy.png}
    \includegraphics[width=0.3\textwidth]{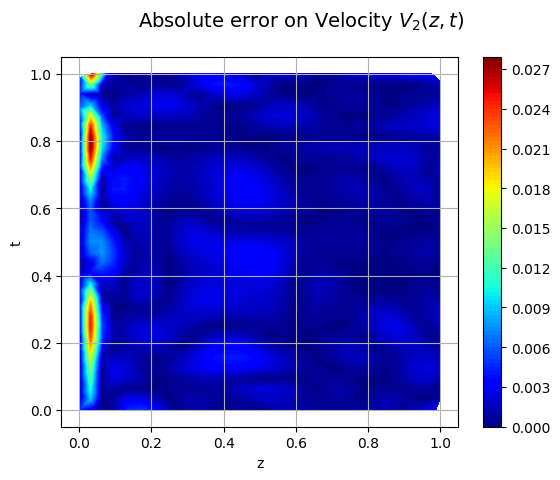}
    \caption{\textbf{\textit{Top row: }Contour plots of predicted (left), exact (middle) and absolute error (right) of the velocity function $V_{1}(z,t)$. \textbf{\textit{Bottom row: }Contour plots of predicted (left), exact (middle) and absolute error (right) of the velocity function $V_{2}(z,t)$.}}}
    \label{fig:3Dpredicted_VelocityX}
\end{figure}

\begin{figure}[htp]
    \centering
    \includegraphics[width=0.3\textwidth]{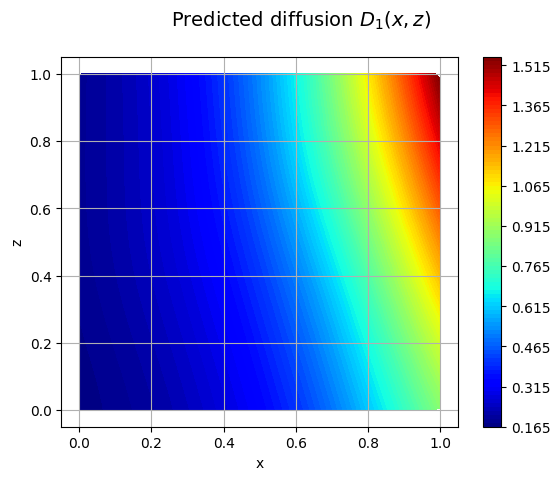}
    \includegraphics[width=0.3\textwidth]{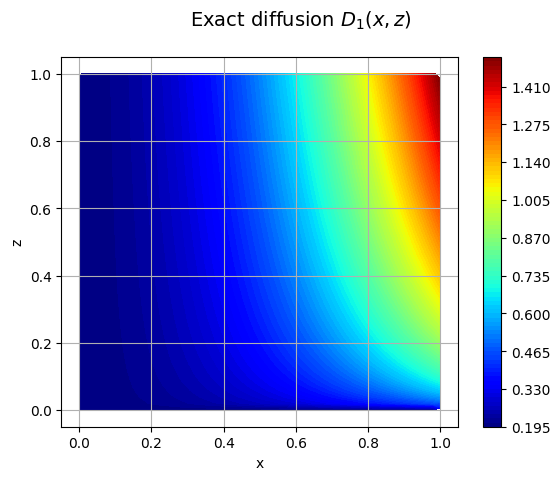}
    \includegraphics[width=0.3\textwidth]{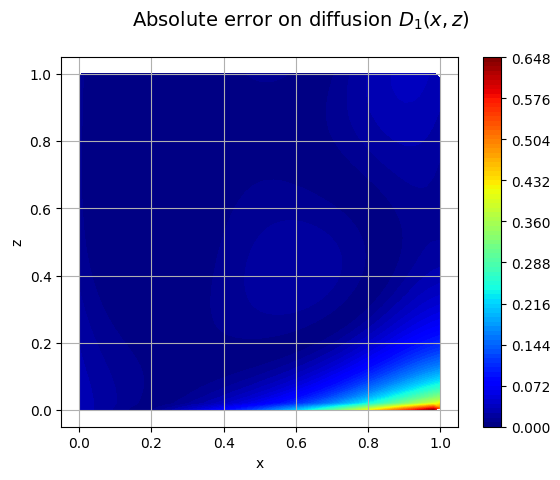}
    \includegraphics[width=0.3\textwidth]{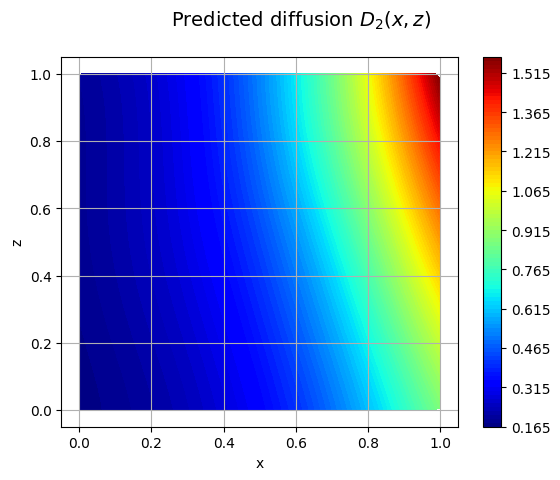}
    \includegraphics[width=0.3\textwidth]{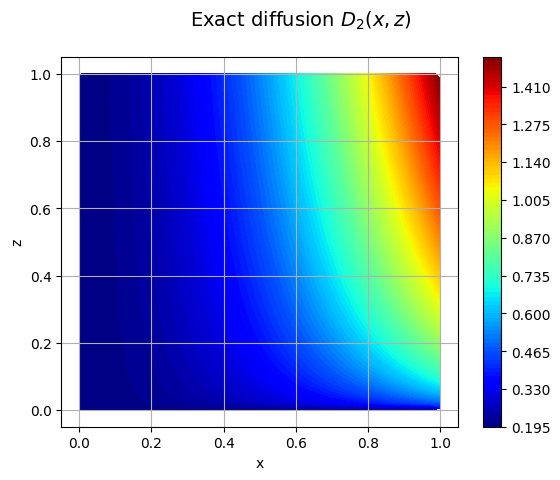}
    \includegraphics[width=0.3\textwidth]{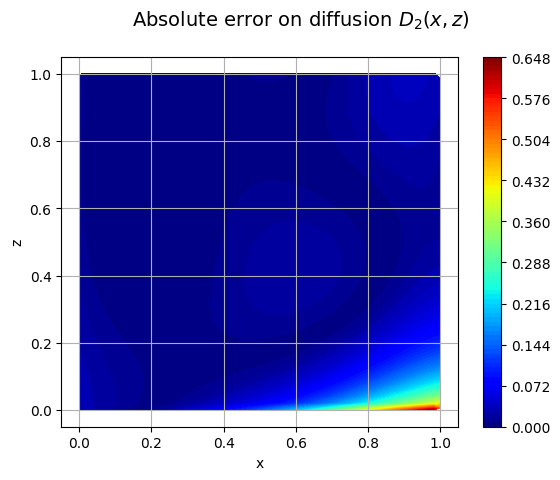}
    \includegraphics[width=0.3\textwidth]{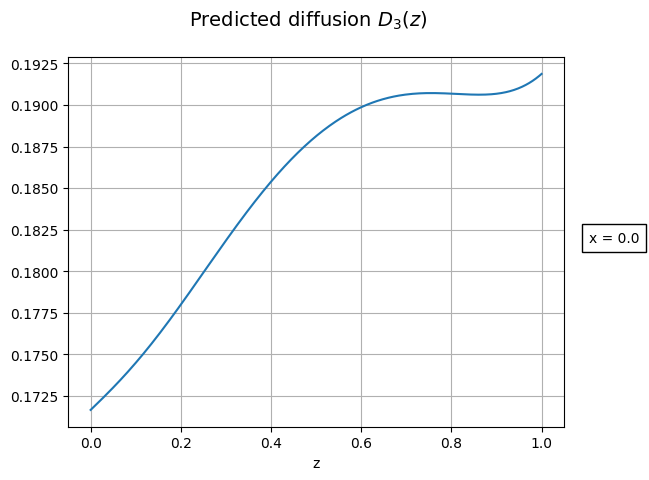}
    \includegraphics[width=0.3\textwidth]{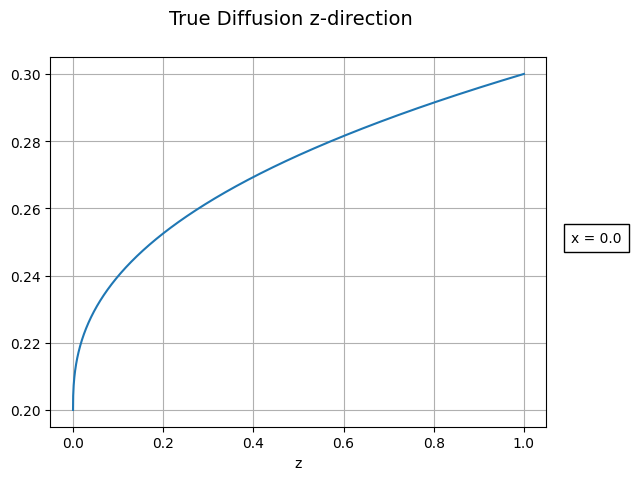}
    \caption{\textbf{\textit{Top and middle row: }contour plots of predicted (left), exact (middle) and absolute error (right) of the diffusion functions $D_{1}(x,z)$ and $D_{2}(x,z)$ respectively.} \textbf{\textit{Bottom row:} predicted vs exact diffusion $D_{3}(z)$.}}
    \label{fig:3Dpredicted_Diffusion}
\end{figure}

\begin{figure}[htp]
    \centering
    \setlength{\tabcolsep}{5pt}
    \renewcommand{\arraystretch}{1.2}
    
    \begin{tabular}{c c c c }
        \begin{overpic}[width=0.3\textwidth] {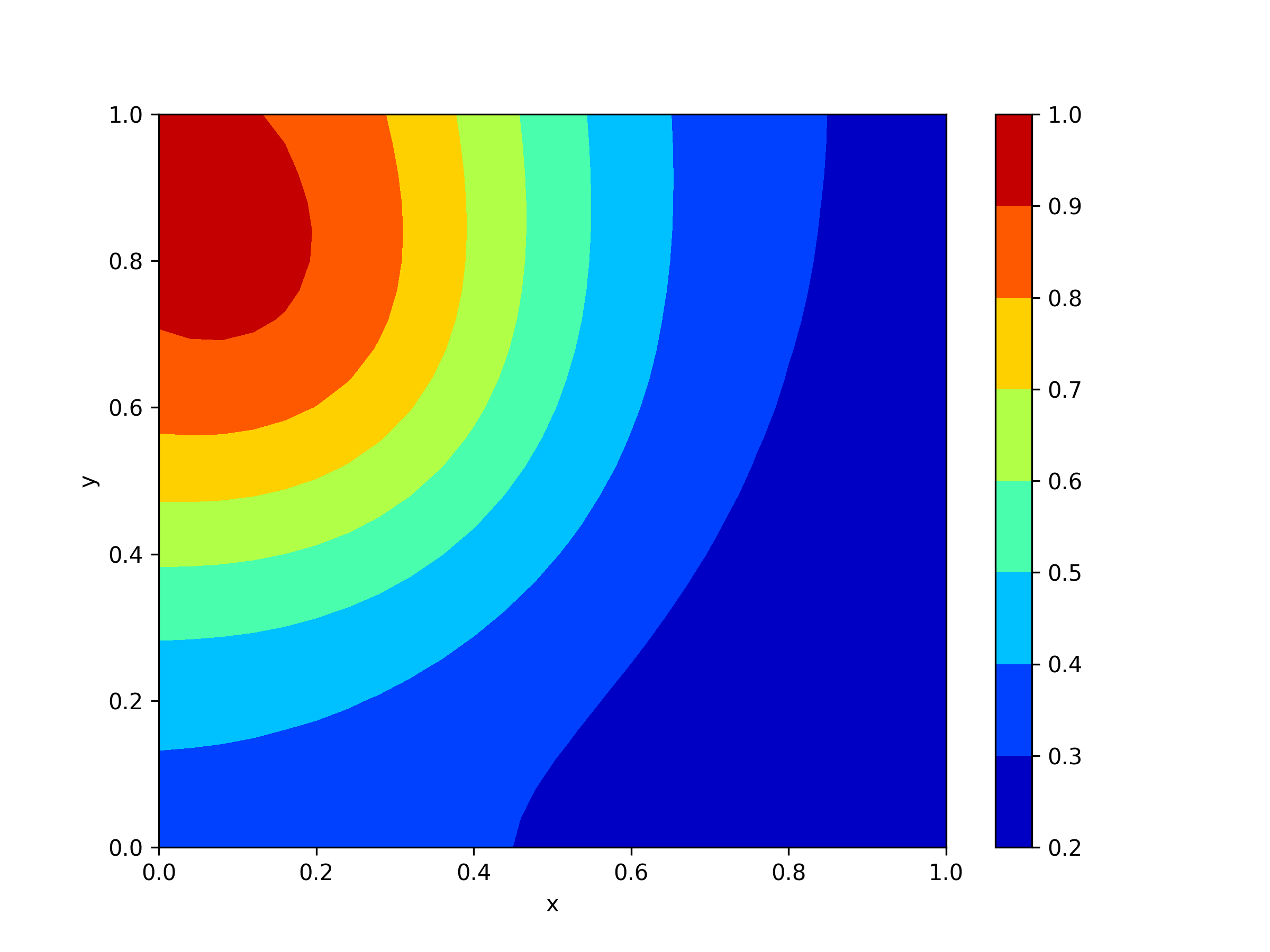}
            \put(20,71){\textbf{Predicted solution}}
        \end{overpic} &
     \begin{overpic}[width=0.3\textwidth] {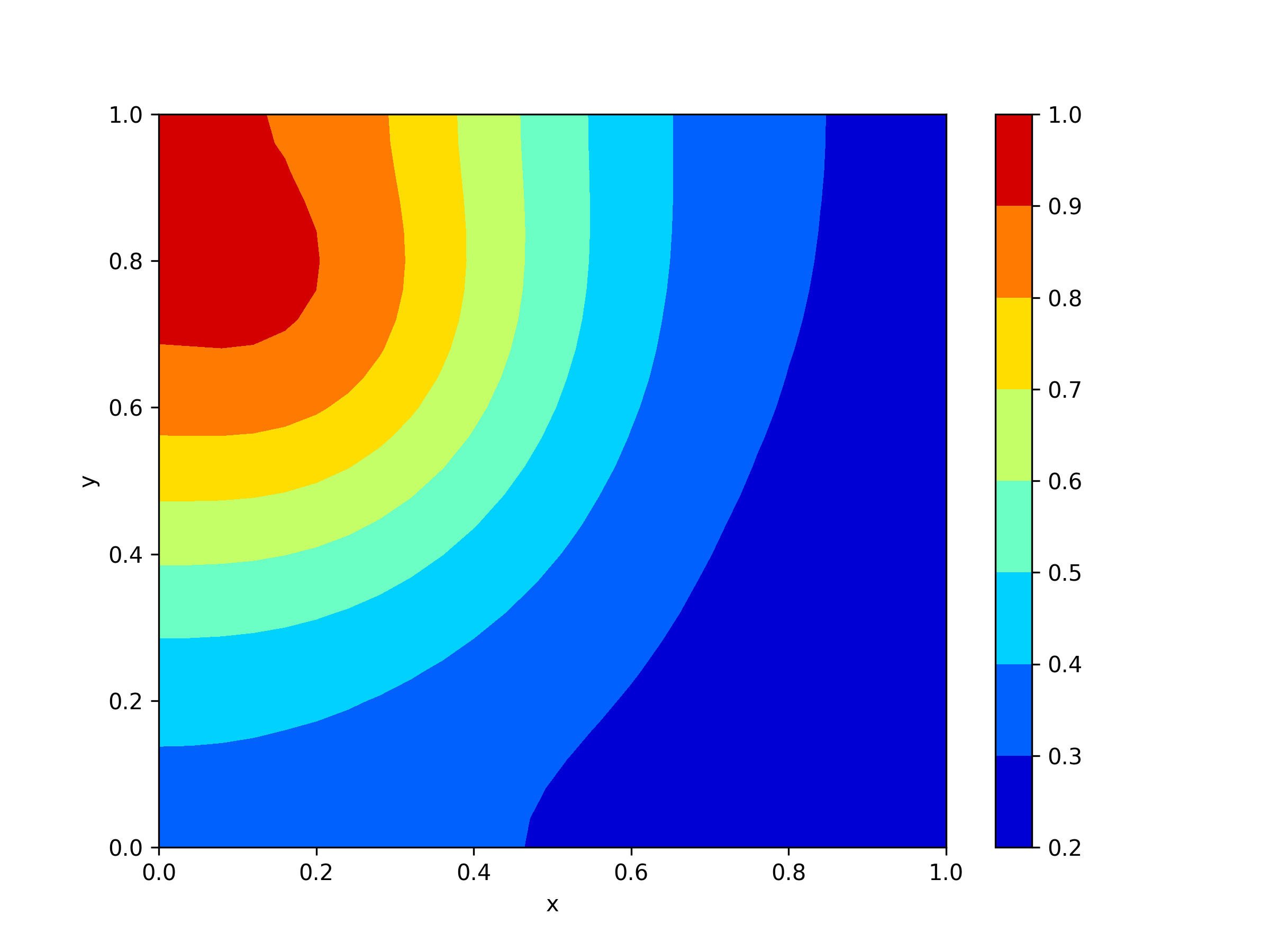}
            \put(20,71){\textbf{FEM solution}}
        \end{overpic} &
        \begin{overpic}[width=0.3\textwidth] {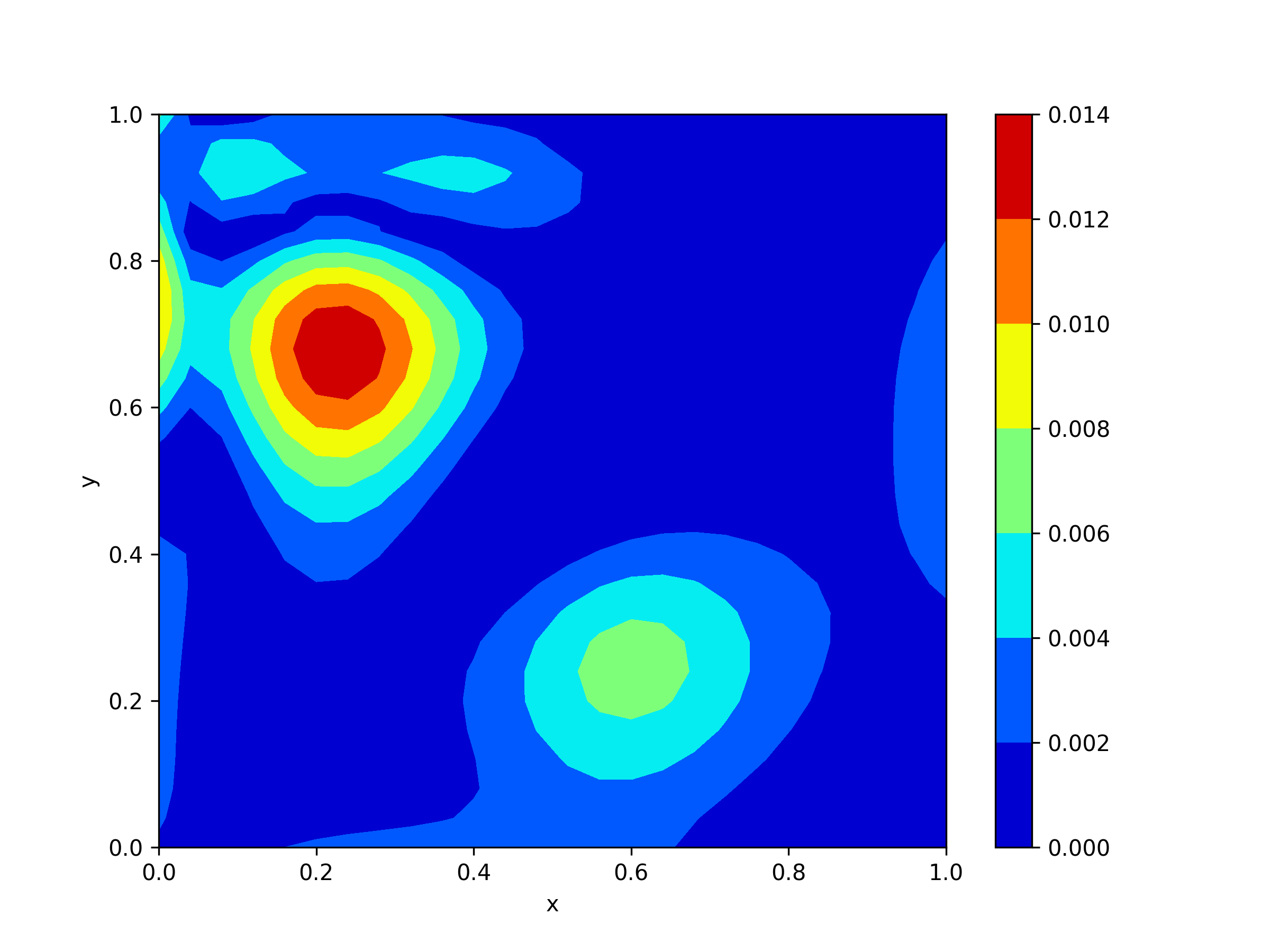}
            \put(0,71){\textbf{Absolute error on solution}}
        \end{overpic} &
    \end{tabular}
         \caption{\textbf{Contour Plots of the Predicted (left) vs FEM Solution(middle) and the absolute error(right) of $u$ at $t=1$.}}
    \label{fig:3DContpredicted_Solution}
\end{figure}

\begin{figure}[htp]
    \centering
    \setlength{\tabcolsep}{5pt}
    \renewcommand{\arraystretch}{1.2}
    
    \begin{tabular}{c c c c }
        \begin{overpic}[width=0.3\textwidth] {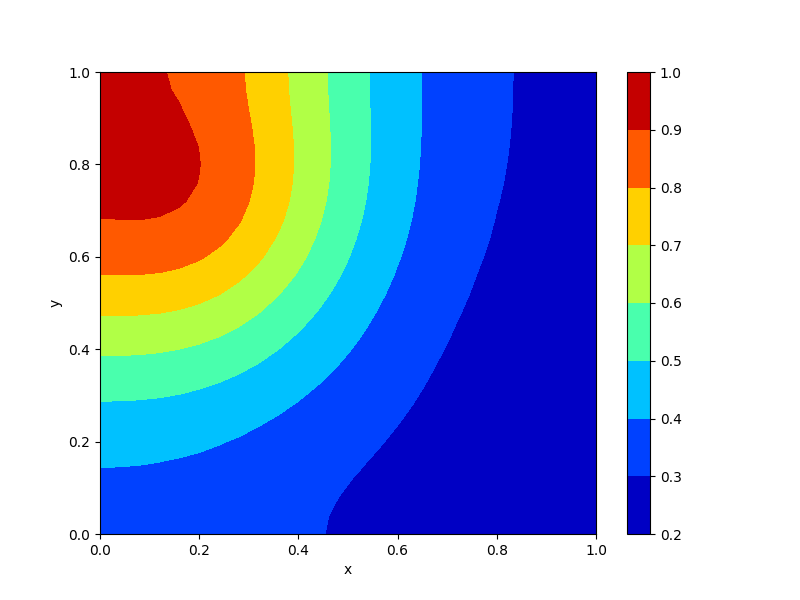}
            \put(20,72){\textbf{Predicted solution}}
        \end{overpic} &
     \begin{overpic}[width=0.3\textwidth] {figures/3DNCExact_u_t1.00.png}
            \put(20,72){\textbf{FEM solution}}
        \end{overpic} &
        \begin{overpic}[width=0.3\textwidth] {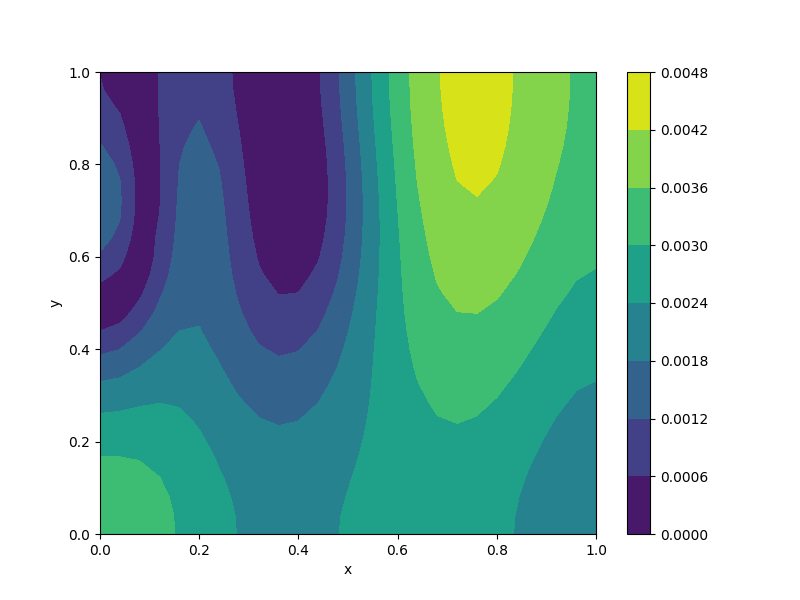}
            \put(0,72){\textbf{Absolute error on solution}}
        \end{overpic} &
    \end{tabular}
         \caption{\textbf{Comparison of FEM solutions at $t=1$ with the FEM solution obtained using the predicted diffusion from the model (left)), the solution with the exact diffusion (middle), with their absolute error (right).}}
    \label{fig:3DComparisonFEM_Solution}
\end{figure}

\begin{figure}[htp]
    \centering
    \setlength{\tabcolsep}{5pt}
    \renewcommand{\arraystretch}{1.2}
    
    \begin{tabular}{c c c c }
     \begin{overpic}[width=0.3\textwidth]{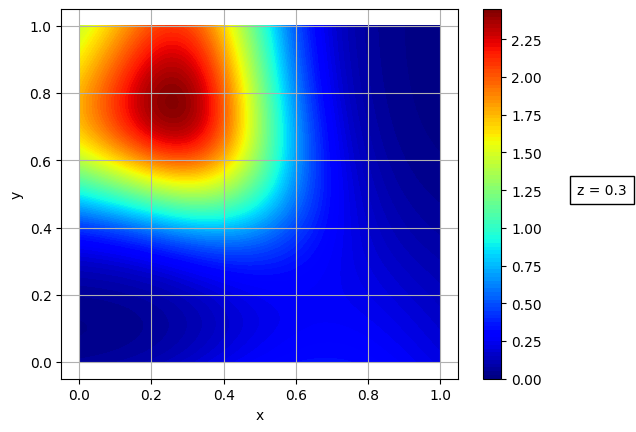}
     \put(10,69){\textbf{Predicted $f$ at $z=0.3$}}
     \end{overpic}
    \begin{overpic}[width=0.3\textwidth]{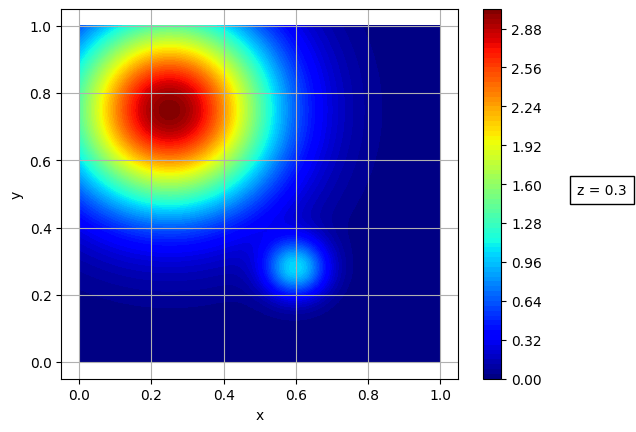}
    \put(10,69){\textbf{Exact $f$ at $z=0.3$}}
    \end{overpic}
    \begin{overpic}[width=0.3\textwidth]{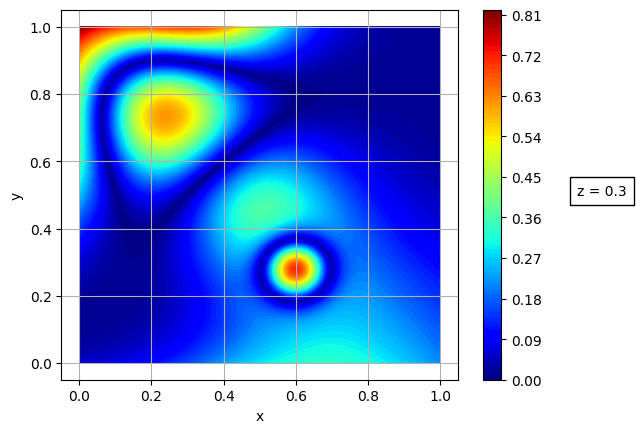}
         \put(10,69){\textbf{Absolute error on $f$ }}
    \end{overpic}\\\\
     \begin{overpic}[width=0.3\textwidth]{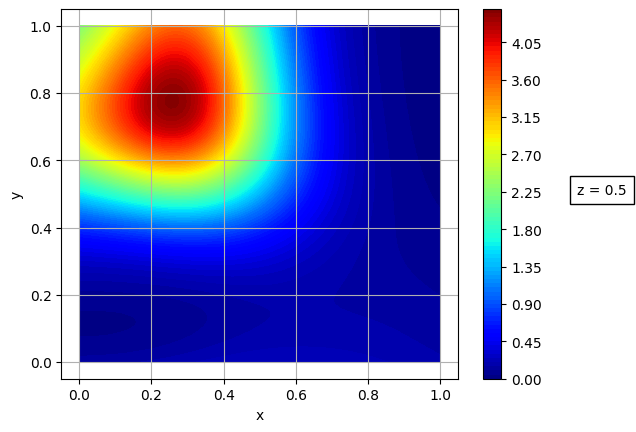}
      \put(10,69){\textbf{Predicted $f$ at $z=0.5$}}
     \end{overpic}
    \begin{overpic}[width=0.3\textwidth]{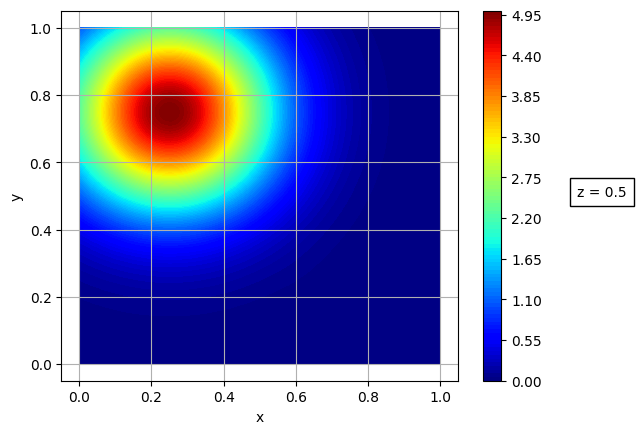}
    \put(10,69){\textbf{Exact $f$ at $z=0.5$}}
    \end{overpic}
    \begin{overpic}[width=0.3\textwidth]{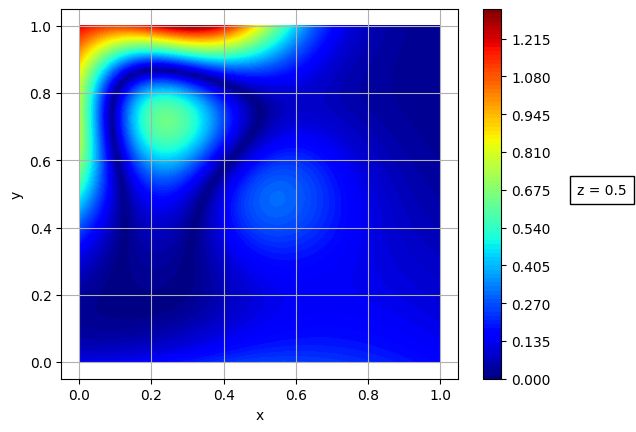}
     \put(10,69){\textbf{Absolute error on $f$ }}
    \end{overpic}
    \end{tabular}
    \caption{\textbf{\textit{Top row:} Contour plots of predicted (left), exact (middle) and absolute error (right) of the source function at $z=0.3$. \textit{Bottom row:} \textit{Middle row:}Contour plots of predicted (left), exact (middle) and absolute error (right) of the source function (bottom) at $z=0.5$.}}
    \label{fig:3Dpredicted_ExactSource}
\end{figure}

We dive into a higher-dimensional case where we consider a 3D spatial domain. In this experimental case, we consider a 3D ADE on a unit cube $\Omega =[0,1]^{3}$ with a Neumann boundary condition at the top boundary layer $z=1$ (a reflective boundary layer at $z=1$) and a Dirichlet condition at the boundary of the domain except the top boundary $z=1$. Therefore, we have the initial and boundary conditions $ u(\mathbf{x},0)= 0 \quad \mathbf{x}\in\Omega , \quad D_{z}\frac{\partial u}{\partial z}(\mathbf{x},t)=0, \quad  \text{at } z=1,$ and
$u(\mathbf{x},t) = 0 \quad  \text{for } \mathbf{x} \in \partial\Omega\text{\textbackslash}\{z=1\}$ respectively.
We assume that the wind speed $V$ is time and height dependent, that is $V(z,t)= (V_{1}(z,t),V_{2}(z,t), V_{3})$ with $V_{3}=V_{set}$ the setting velocity given by the Stokes law \cite{stockie2011mathematics, lin1996analytical, hosseini2016airborne} as $V_{set}= \frac{\rho g d^{2}}{18\mu},$
where $ \rho=1500 \text{[kg/$m^3$]}, d=2e^{-6}[m],  g = 9.8 [m/s^{2}], \mu = 1.8 \times 10^{-5} [kg/m s]$ represent the air pollutants density, the air pollutants diameter,  the gravitational acceleration and the viscosity of air respectively.
We also assume that $V_{1}$ and $V_{2}$ are derived from the power law correlation \cite{arya1999air} at the reference altitude of $z=0.5$, with the reference velocity denoted as $V_{ref}$. Therefore, we have $V_{1}(z,t)= \big(\frac{z}{0.5}\big)^{n}V_{ref}$ with $V_{ref}=0.8(1+\sin(2\pi t))$ and $V_{2}(z,t)=-V_{1}(z,t)$. We assume $n=0.4$, which corresponds to a rough surface domain. We assume that we have 3 different locations of emission sources \\
$S=\{(0.25,0.75, 0.5), (0.52,0.8,0.8), (0.6,0.28,0.3)\}$ with their respective emission rate $r_{1}=5, r_{2}=3, r_{3}=1$. The exact source function is the sum of the Gaussian kernel functions at those locations with the respective standard deviation $\sigma_{1}=0.2, \sigma_{2}=0.06, \sigma_{3}=0.08$ and consequently, $f$ is written as
\begin{eqnarray*}
	f(x,y,z)= \sum_{i=1}^{3}r_{i}\exp\Bigg(-\frac{\big((x-x_{i})^{2}+(y-y_{i})^{2}+(z-z_{i})^{2}\big)}{2\sigma_{i}^{2}}\Bigg), \quad (x_{i}, y_{i}, z_{i})\in S
\end{eqnarray*}
We assume the the first two components of the diffusion coefficient to be isotropic in the XZ-plane and the third component height dependent only with a boundary layer $H=1$ i.e $D(x,z)=(D_{1}(x,z),D_{2}(x,z),D_{3}(z))=(0.2+x^{2}(z/0.5)^{n},0.2+x^{2}(z/0.5)^{n}, 0.2+0.1(\frac{z}{H})^{n})$.
We also assume we have 30 measurements locations at which we have noisy data and noisy wind speed data $V$ at those locations. Therefore, the loss function is similar to (\ref{loss}) with $\textbf{x}=(x,y,z)$.\\
Under the above-mentioned assumptions, we randomly choose 30 measurements locations and initialize the neural networks $u(\mathbf{x},t,\pmb{\theta}_{u}), f(\mathbf{x},\pmb{\theta}_{f})$,\\ $V(t,\pmb{\theta}_{V}),D(\mathbf{x},\pmb{\theta}_{D})$ with [4,80,80,80,80,80,1], [3,100,100,100,1], [2,80,80,80,60,60], [2,80,80,80,80,80,1] respectively. The training process of these neural networks is spited into 3 phases:\\
\textbf{Pretraining phase:} we first train the model with 10000 ADAM optimizer  while minimizing the data losses of \eqref{loss}, that is,
\begin{eqnarray*}
	\frac{1}{N_{b}}\sum_{i=1}^{N_{b}}(\mathcal{B}[u(\textbf{x}^{i}_{b}, t_{b}^{i};\pmb{\theta}_{u})]- h(\textbf{x}^{i}_{b},t^{i}_{b}))^{2}+ \frac{1}{N_{z}}\sum_{i=1}^{N_{z}}(\tau_{i}[u(\textbf{x}^{i}_{z}, t_{z}^{i};\pmb{\theta}_{u})]- \mathbf{z}^\ast_{i})^{2}\\
	+\frac{1}{N_{v}}\sum_{i=1}^{N_{v}}(v(t_{v}^{i};\pmb{\theta}_{V})]- V^{*}(t^{i}_{v}))^{2}
\end{eqnarray*}
The loss weights are not updated in this phase.\\
\textbf{Training phase while updating the loss weights:} we train the PINN model with 20000 ADAM optimizer using the total loss \eqref{loss}. The neural networks parameters $\pmb{\Theta}$, as well as the loss weights $\lambda_{r}, \lambda_{b}, \lambda_{v}$ and $\lambda_{z}$ are being updated during the training.\\
\textbf{Fine tuning with L-BFGS:} the third phase consists in updating $\pmb{\Theta}$ to minimize \eqref{loss} using L-BFGS over 30000 iterations.\\
The numerical results shown bellow indicate that once again the PINNs model is able to achieve an accurate approximation on the velocity vector (Figure \ref{fig:3Dpredicted_VelocityX}) with an absolute error that is nearly zero throughout most of the domain except near the left boundary of the domain. This can be explained by the non specification of initial or boundary conditions on the velocity to the model and therefore, when $z=0$ the model might be learning only from the noise, resulting in an approximation with larger error. In addition, the model approximates $V_{3}=2.25e-5$ with a higher amplitude compared to the exact value $V_3 = 2.893518518518519e-07$ since $V_3$ itself is extremely small. The model is still able to approximate well the solution $u$ with a maximum approximation error (MAE) of $2.4e-2$ as we observe in Figure \ref{fig:3DContpredicted_Solution}, achieving at the final time $t=1.0$ an L2-error of $5.6e-3$ and a relative L2-error of $1e-2$. Furthermore, the PINN model gives a good approximation of diffusion in the xz-plane Figure \ref{fig:3Dpredicted_Diffusion}, while the approximation of the diffusion in the z direction $D_{3}(z)$ is inherently difficult once again owing to the ill-posedness of the problem and the scarcity in the measurements data. Errors arising from the approximation of $D_{3}(z)$ propagate through the model and directly affect the accuracy in the approximation of the source function as illustrated in the figure \ref{fig:3Dpredicted_ExactSource}. Despite the scaling difference observed on $D_3$, on may question the reliability of the predicted diffusion $D_3$. By considering the predicted diffusion from the model as the true diffusion, and after solving the ADE with finite element method, the resulting solution successfully reconstructs the original FEM solution with a relative L2-error of $4.94e-3$ as illustrated on the figure \ref{fig:3DComparisonFEM_Solution}. Thus, this demonstrate how the model preserves the essential characteristics of the true diffusion function and is still able to approximate the source function with a reasonable accuracy.  

\section{Error estimate of PINNs in a source inversion problem on ADE } \label{generroranalysis}
The analysis presented here follows the abstract framework of \cite{mishra2022estimates}, adapted to our source inversion setting. Our analysis relies on two key steps: First, we establish a conditional Lipschitz stability estimate for recovering the source function from interior observations, based on the Carleman estimate of \cite{yamamoto2009carleman}; Second, we use standard parabolic regularity theory to give error bounds on $u$ and $f$ in terms of training and discretization errors and the observation noise level.

Although the adaptive PINN method used in this work simultaneously estimates the source term $f$, the solution $u^\ast$ and the PDE parameters $V$ and $D$ in a joint optimization framework, the theoretical analysis presented here is restricted to the case of known parameters $V$ and $D$. This restriction is motivated by the absence, to the best of our knowledge, of conditional stability results for the joint recovery of the source term, the solution, and PDE coefficients simultaneously from sparse observations in parabolic PDEs.
The extension of conditional stability analysis to this type of joint optimization problem constitutes an important and challenging open problem that we leave for future work. Nevertheless, our numerical results provide strong empirical evidence for the practical viability of the joint optimization problem  beyond the theory outlined below.

Let $\Omega \subset \mathbb{R}^{d}, (d\geq 1)$ be an open and bounded domain with Lipschitz boundary $\partial\Omega$. We consider a source inversion problem of the ADE \eqref{ade} with homogeneous Neumann boundary condition.
We note that \eqref{ade} would be well posed only if the initial condition is specified, thus, for simplicity, we assume that $u(\mathbf{x},0)=0$. We further assume a time independent/stationary source function $f\in L^{2}(\Omega)$ and  $V,D\in C^{\infty}(\Omega)$  sufficiently smooth functions that 
are known. We also assume that $u\in H^{1}(0,T;L^{2}(\Omega))\cap L^{2}(0,T;H^{2}(\Omega))$ is the solution to \eqref{ade} in a weak sense.

\subsection{Conditional stability}
Let $\omega\subset \Omega$ a compact subdomain of spatial measurement locations $\{\mathbf{x}_{i}\}_{i=1}^{N_{\mathbf{z}}}$ at which we have the measurements data. We assume we have observation $z$ of $u$ on $\omega\times [0,T]$ we will estimate $f$ and $u$ based on observation $u|_{\omega\times]t_0-\eta,t_0+\eta[}$ and  we set $Q=\Omega\times(0,T))$ and $Q_{\omega}=\omega\times]t_0-\eta,t_0+\eta[$. The following theorem gives the conditional stability of a source inversion problem   with known PDE coefficients.

\begin{theorem} \label{cond_stab}
	Let  $(u,f)$ be the pair of solution to the source inverse problem on \eqref{ade} with $u\in H^{1}(0,T;L^{2}(\Omega))\cap L^{2}(0,T;H^{2}(\Omega))$  then there exists a constant $C>0$ depending on $\omega, t_{0}, \Omega$ such that
	\begin{eqnarray}
		||f||_{L^{2}(\Omega)}\leq C \bigg[||\partial_{t}u||_{L^2(Q_{\omega})} + ||Au(\cdot, t_{0})||_{L^{2}(\Omega)} \bigg]
		\label{condstabF}
	\end{eqnarray}
	where we introduced the shorthand notation $A:u \mapsto  V\cdot\nabla u -\nabla(D\nabla u)$.
\end{theorem}
\begin{proof}
	See Appendix \ref{proof::theo1AppendixC}.
\end{proof}
\subsection{Error estimate of PINNs in reconstructing the solution and the source function}
We consider  two separate neural networks $\pmb{\theta}_{u}$ and $\pmb{\theta}_{f}$ that approximate using PINNs, the solution and the source function  respectively. In contrast to the standard loss often used in PINNs
\begin{eqnarray*}
	\mathcal{L}(\pmb{\theta}_{u},\pmb{\theta}_{f})&=& \lambda_{r}||\mathcal{R}_{int,\pmb{\theta}_{u},\pmb{\theta}_{f}}||^{2}_{L^2(Q)}+ \lambda_{b}||\mathcal{R}_{bc;\pmb{\theta}_{u}}||^{2}_{L^{2}(\partial Q))} \nonumber\\
	&+&\lambda_{z}||\mathcal{R}_{z,{\pmb\theta_{u}}}||^{2}_{L^2(\Omega))}
	+ ||\mathcal{R}_{ic,\pmb{\theta}_{u}}||^{2}_{L^{2}(\Omega)},
	\label{contloss}
\end{eqnarray*}
we define the following loss function, motivated by the theory on conditional stability Theorem \ref{cond_stab}
\begin{eqnarray}
	\mathcal{L}_{\lambda}(\pmb{\theta}_{u},\pmb{\theta}_{f})&=& \lambda_{r}||\mathcal{R}_{int,\pmb{\theta}_{u},\pmb{\theta}_{f}}||_{H^{1}(0,T;L^2(Q))}+ \lambda_{b}||\mathcal{R}_{bc;\pmb{\theta}_{u}}||_{H^{2}(0,T;L^{2}(\partial\Omega))} \nonumber\\
	&+&\lambda_{z}||\mathcal{R}_{z,{\pmb\theta_{u}}}||_{H^{2}(\Omega)}
	+ ||\mathcal{R}_{ic,\pmb{\theta}_{u}}||_{H^{2}(\Omega)}
	\label{contloss}
\end{eqnarray}
where $\lambda=\{\lambda_{r},\lambda_{b},\lambda_{z}\}$ are the loss weight hyperparameters as usual; while these 
are adaptive in our algorithm.
We consider our noisy measurements data $z^{\delta}$ at a fixed time $t_0$ in $\omega$, with spatially varying noise level, defined by:
$$
z_{i}^{\delta}(\mathbf{x})= u(\mathbf{x}_{i},t_0)(1+\delta_{i}\cdot \mathcal{N}(0,1)), i=1,...,N_{z}
$$
with $\delta_i$ the noise level of the measurements data at the location $\mathbf{x}_{i}$ and $N_{z}$ is the number of measurements locations. Then various $\mathcal{R}$ terms 
in \eqref{cond_stab} are defined as 
\begin{align*}
	\mathcal{R}_{int,\pmb{\theta}_{u},\pmb{\theta}_{f}}(\mathbf{x},t)&:= \partial _{t}u_{\pmb{\theta}_{u}}(\mathbf{x},t) + V(\mathbf{x})\cdot \nabla u_{\pmb{\theta}_{u}}(\mathbf{x},t)\\
	 &-\nabla(D(\mathbf{x})\nabla  u_{\pmb{\theta}_{u}}(\mathbf{x},t)) -f_{\pmb{\theta}_{f}}(\mathbf{x}) \quad \text{in } Q \\
	\mathcal{R}_{z,{\pmb{\theta}_{u}}}(\mathbf{x},t)&:= u_{\pmb{\theta}_{u}}(\mathbf{x},t)-z^{\delta}(\mathbf{x}) \quad \forall (\mathbf{x},t) \in Q_{\omega}\\
	\mathcal{R}_{bc,\pmb{\theta}_{u}}(\mathbf{x},t)&:= \nabla_{n}  u_{\pmb{\theta}_{u}} (\mathbf{x},t) \quad \forall (\mathbf{x},t)\in \partial Q\\
	\mathcal{R}_{ic,\pmb{\theta}_{u}}(\mathbf{x})&:= u(\mathbf{x},0), \quad \mathbf{x}\in\Omega,
\end{align*}
which, in order, are the interior PDE residual, data misfit, boundary 
residual, and the initial condition residual.

We choose to approximate the integrals in \eqref{contloss} by quadrature, and thus we consider the training set $S\subset \bar{\Omega}\times (0,T)$ such that $S=S_{int}\cup S_{bc}\cup S_{ic}\cup S_{z}$ with \\
$S_{int}=\{(\mathbf{x}_n, t_n)\}$ for $1\leq n\leq N_{int}$ with each $(\mathbf{x}_n,t_n)\in \Omega\times (0,T)$\\
$S_{bc}=\{(\mathbf{x}_n,t_n)\}$ for $1\leq n\leq N_{bc}$ with each $(\mathbf{x}_n,t_n)\in \partial\Omega \times (0,T)$\\
$S_{ic}=\{\mathbf{x}_n\}$ for $1\leq n\leq N_{ic}$ with each  $\mathbf{x}_n\in \Omega$.\\
$S_{z}= \{(\mathbf{x}_{n})\}$ for $1\leq n\leq N_{z}$ with $\mathbf{x}_n\in \omega $\\
We define the discrete  PINN loss:
\begin{eqnarray}
	\mathcal{L}^{N}_{\lambda}(\pmb{\theta}_{u},\pmb{\theta}_{f}; S)&=& \lambda_{r}\mathcal{E}_{int}^{T}(\pmb{\theta}_{u},\pmb{\theta}_{f}; S_{int})^{2}+ \lambda_{b} \mathcal{E}_{bc}^{T}(\pmb{\theta}_{u}; S_{bc})^2\nonumber\\
	&+& \lambda_{z}\mathcal{E}_{z}^{T}(\pmb{\theta}_{u}; S_{z})^2 + \mathcal{E}_{ic}^{T}(\pmb{\theta}_{u}; S_{ic})^2 
	\label{discloss}
\end{eqnarray}
with the training errors
\begin{align*}
	\mathcal{E}_{int}^{T}(\pmb{\theta}_{u},\pmb{\theta}_{f}; S_{int})^{2}&= \sum_{n=1}^{N_{int}}w_{n}^{int}|\mathcal{R}_{int,\pmb{\theta}_{u},\pmb{\theta}_{f}}(\mathbf{x}_{n}^{int},t_{n}^{int})|^2 \\
	\mathcal{E}_{ic}^{T}(\pmb{\theta}_u; S_{ic})^{2}&= \sum_{n=1}^{N_{ic}}w_{n}^{ic}|\mathcal{R}_{ic,\pmb{\theta}_{u}}(\mathbf{x}_{n}^{ic},t_{n}^{ic} )|^2 \end{align*}
\begin{align*}
	\mathcal{E}_{bc}^{T}(\pmb{\theta}_{u}; S_{bc})^{2}&= \sum_{n=1}^{N_{bc}}w_{n}^{bc}|\mathcal{R}_{bc,\pmb{\theta}_{u}}( \mathbf{x}_{n}^{bc},t_{n}^{bc})|^2 \\
	\mathcal{E}_{z}^{T}(\pmb{\theta}_{u}; S_{z})^{2}&= \sum_{n=1}^{N_{z}}w_{n}^{z}|\mathcal{R}_{z,\pmb{\theta}_{u}}( \mathbf{x}_{n}^{z})|^2 
\end{align*}
with $w_{n}^{int}, w_{n}^{ic},w_{n}^{bc},w_{n}^{z}$ denoting 
quadrature weights. With this setup 
we recall the following proposition that
characterizes the discretization error 
of the PINN loss:

\begin{prop}\label{prop}
	The difference between $\mathcal{L}_{\lambda}(\pmb{\theta}_{u},\pmb{\theta}_{f})$ and $\mathcal{L}^{N}_{\lambda}(\pmb{\theta}_{u},\pmb{\theta}_{f};S)$  satisfies the error estimate
	\begin{equation}\label{eq:quadrature_error}
		\left|\mathcal{L}_{\lambda}(\pmb{\theta}_{u},\pmb{\theta}_{f})
		-\mathcal{L}^{N}_{\lambda}(\pmb{\theta}_{u},\pmb{\theta}_{f};S)\right| 
		\leq C \max
		\left\{
		N_z^{-\beta_{z}}, \quad
		N_{int}^{-\beta_{int}}, \quad
		N_{ic}^{-\beta_{ic}}, \quad
		N_{bc}^{-\beta_{bc}}
		\right\},
	\end{equation}
	where the orders of convergence $\beta > 0$ and $C>0$ is a constant depending on the adaptive loss weights $\lambda$ and the supremum bound of the residual terms under the continuous norm $\|\cdot\|_{C(\Omega)}$.
\end{prop}
\begin{proof}
	For a proof see {\cite[Sec.~2.2]{mishra2022estimates},\cite[Prop.~3.1]{zhang2025convergence}}.
	
\end{proof}

Write $(u^\ast, f^\ast):=(u^{\ast}_{\pmb{\theta}^{\ast}_{u}},f^{\ast}_{\pmb{\theta}^{\ast}_{f}})$ for the optimal pair of solutions of the minimization problem:
\begin{equation*}
	\arg\min_{\pmb{\theta}_{u},\pmb{\theta}_{f}} \mathcal{L}^{N}_{\lambda}(\pmb{\theta}_{u},\pmb{\theta}_{f}; S),
\end{equation*}
and define the generalization error of PINNs for approximating $u$ and $f$ as follows:
\begin{eqnarray}
	\mathcal{E}_{u}:=||u^\ast -u||_{C([0,T];L^{2}(\Omega))} \qquad \mathcal{E}_{f}:=||f^\ast -f||_{L^{2}(\Omega)}
\end{eqnarray}
\begin{theorem}\label{thm::generrorbounds}
	We have the following error estimate
	\begin{align}
		\mathcal{E}_{f} 
		&\leq C\Bigl(
		\mathcal{E}_{z}^{T}(\pmb{\theta}_{u}^\ast;S_z) 
		+ \mathcal{E}_{int}^{T}(\pmb{\theta}_{u}^\ast,\pmb{\theta}_{f}^\ast; S_{int})
		+ \mathcal{E}_{bc}^{T}(\pmb{\theta}_{u}^\ast; S_{ic}) + \mathcal{E}^{T}_{ic}(\pmb{\theta}_{u}^\ast; S_{bc})  \nonumber\\
		&\quad + C_q^{\frac{1}{2}} N^{-\frac{\beta}{2}} 
		+ \delta
		\Bigr)
		\label{genError:f}
	\end{align}
	\begin{align}
		\mathcal{E}_{u} 
		&\leq \bar{C}\Bigl(
		\mathcal{E}_{z}^{T}(\pmb{\theta}_{u}^\ast;S_z) 
		+ \mathcal{E}_{int}^{T}(\pmb{\theta}_{u}^\ast,\pmb{\theta}_{f}^\ast; S_{int})
		+ \mathcal{E}_{ic}^{T}(\pmb{\theta}_{u}^\ast; S_{ic}) + \mathcal{E}^{T}_{bc}(\pmb{\theta}_{u}^\ast; S_{bc})  \nonumber\\
		&\quad + C_q^{\frac{1}{2}} N^{-\frac{\beta}{2}}  + \delta
		\Bigr)
		\label{genError:u}
	\end{align}
	Here, the constants $C$ and $\bar{C}$ depend on the adaptive weights $\lambda$ and the underlying geometry $T,u^\ast, u,f,f^\ast$, and 
	\begin{align*}
		N &:= \min\left\{N_z,\, N_{int},\, N_{bc},\, N_{ic}\right\},\\[6pt]
		\beta &:= \min\left\{
		\beta_{z},\, \beta_{int},\, \beta_{bc},\,
		\beta_{ic}
		\right\},\\[6pt]
		C_q &:= \max\left\{
		C_{q,z},\, C_{q,int},\, C_{q,bc},\, C_{q,ic}
		\right\},
	\end{align*}
	where the quadrature constants are defined as follows:
	\begin{align*}
		C_{q,z} &= C_{q,z}\!\left(
		\|\mathcal{R}_{z,\pmb{\theta}_{u}^*}\|_{C(\Omega)}
		\right), 
		&C_{q,int} &= C_{q,int}\!\left(
		\|\mathcal{R}_{int,\pmb{\theta}_{u}^*,\pmb{\theta}_{f}^*}\|_{C(Q)}
		\right),\\[6pt]
		C_{q,bc} &= C_{q,bc}\!\left(
		\|\mathcal{R}_{bc,\pmb{\theta}_{u}^\ast}\|_{C(\partial Q)}
		\right), 
		&C_{q,ic} &= C_{q,ic}\!\left(
		\|\mathcal{R}_{ic,\pmb{\theta}_{u}^{\ast}}\|_{C(\Omega)}
		\right).
	\end{align*}
\end{theorem}
\begin{proof}
	See Appendix \ref{proof::theo2AppendixD}
\end{proof}

The theoretical framework considered here is developed for source inversion with known diffusion coefficients $D$ and velocity field $V$, whereas the numerical experiments investigate the more challenging problem of simultaneously estimating the source, the diffusion, the velocity and the solution.  An estimate of the error bound for this type joint parameter estimation and source inverse problem considered in this work remains challenging. Deriving such a result would require establishing a conditional stability estimate for the simultaneous recovery of multiple coefficients and source terms from the measurements of the solution. In the absence of additional information, sparse measurements of the solution alone may be insufficient to ensure identifiability, in addition to the severe ill-posedness of the problem. Nevertheless, the theoretical analysis provides important qualitative insight into the joint estimation setting, as the corresponding PINN loss still contains analogous residual, boundary, data loss terms. Therefore, the dependence of reconstruction accuracy on training loss, measurement density, and noise level identified in the theoretical analysis remains relevant for explaining and interpreting the numerical results.

\section{Conclusion}\label{sec:conclusion}

This work presents a weighted adaptive method based on the neural tangent kernel of PINNs to solve a source inversion problem with parameter estimation from scarce data. The presented framework demonstrated robust performance, achieving good performances in 2D and 3D advection diffusion equations with constant and spatially varying coefficients, despite the severe ill-posedness of the problem. Our numerical results demonstrate that the source function, the solution and the velocity can be accurately estimated even under limited number of observations. In addition, they also reveal the difficulty in estimating the diffusion coefficient compared to the velocity parameter, and this observation is addressed in \cite{stockie2011mathematics} as a common observation in the field of atmospheric dispersion. Although the presented results are promising, one should expect some improvements in the approximation of the diffusion by increasing the spatial coverage and density of measurements data. Our numerical experiments also demonstrate faster convergence of PINNs in such type of source inverse problems on ADE compared to the corresponding forward problem, even when the latter is informed by measurements of the solution. This observation suggests that, the optimization loss landscape for ADE might be more favorable for PINN's training in the inverse formulation than that of the corresponding forward formulation therefore, illustrating how PINNs in some setting can be particularly best suited for inverse than forward problems, as suggested as future investigation by \textit{Wang et al.} see \cite{wang2022and}.\\
Our theoretical result provides an estimate of the error bounds in approximating the solution and the source function when the coefficients of the ADE are known. This bound provides a theoretical understanding of how the noise level in the data, the optimization accuracy and the quadrature error affect the estimation of the source and the recovery of the solution. Although extending this analysis to the simultaneous recovery of velocity, diffusion coefficients and source terms remains an open problem which could be investigated further in the future, the numerical results demonstrate the potential of PINNs for joint parameter identification and source estimation from sparse measurements.

\section*{Funding}
This work was made possible by a grant
from DeepMind Technologies Limited (provided through the African Institute for
Mathematical Sciences). The statements made and views expressed are solely the
responsibility of the authors. The authors sincerely thank the African Institute for Mathematical Sciences Research and Innovation Centre for their support and resources provided during this research.
BH was supported by the National Science Foundation grants 
DMS DMS-2337678, "CAREER: Gaussian Processes for Scientific Machine
Learning: Theoretical Analysis and Computational Algorithms"
and DMS-2208535 "Machine Learning for Bayesian Inverse Problems".

\section*{Conflict of interest}{All authors declare no conflicts of interests.}

\bibliographystyle{unsrt}  
\bibliography{references}

@article{wang2022and,
  title={When and why PINNs fail to train: A neural tangent kernel perspective},
  author={Wang, Sifan and Yu, Xinling and Perdikaris, Paris},
  journal={Journal of Computational Physics},
  volume={449},
  pages={110768},
  year={2022},
  publisher={Elsevier}
}

@book{seinfeld2016atmospheric,
  title={Atmospheric chemistry and physics: from air pollution to climate change},
  author={Seinfeld, John H and Pandis, Spyros N},
  year={2016},
  publisher={John Wiley \& Sons}
}

@article{rasht2022physics,
  title={Physics-informed neural networks (PINNs) for wave propagation and full waveform inversions},
  author={Rasht-Behesht, Majid and Huber, Christian and Shukla, Khemraj and Karniadakis, George Em},
  journal={Journal of Geophysical Research: Solid Earth},
  volume={127},
  number={5},
  pages={e2021JB023120},
  year={2022},
  publisher={Wiley Online Library}
}

@article{depina2022application,
  title={Application of physics-informed neural networks to inverse problems in unsaturated groundwater flow},
  author={Depina, Ivan and Jain, Saket and Mar Valsson, Sigurdur and Gotovac, Hrvoje},
  journal={Georisk: Assessment and Management of Risk for Engineered Systems and Geohazards},
  volume={16},
  number={1},
  pages={21--36},
  year={2022},
  publisher={Taylor \& Francis}
}

@article{kim2024review,
  title={A review of physics informed neural networks for multiscale analysis and inverse problems},
  author={Kim, Dongjin and Lee, Jaewook},
  journal={Multiscale Science and Engineering},
  volume={6},
  number={1},
  pages={1--11},
  year={2024},
  publisher={Springer}
}

@article{sahin2024solving,
  title={Solving forward and inverse problems of contact mechanics using physics-informed neural networks},
  author={Sahin, Tarik and von Danwitz, Max and Popp, Alexander},
  journal={Advanced Modeling and Simulation in Engineering Sciences},
  volume={11},
  number={1},
  pages={11},
  year={2024},
  publisher={Springer}
}

@article{jalalian2025data,
  title={Data-Efficient Kernel Methods for Learning Differential Equations and Their Solution Operators: Algorithms and Error Analysis},
  author={Jalalian, Yasamin and Ramirez, Juan Felipe Osorio and Hsu, Alexander and Hosseini, Bamdad and Owhadi, Houman},
  journal={arXiv preprint arXiv:2503.01036},
  year={2025}
}

@article{jalalian2025hamiltonian,
  title={Data-efficient kernel methods for learning hamiltonian systems},
  author={Jalalian, Yasamin and Samir, Mostafa and Hamzi, Boumediene and Tavallali, Peyman and Owhadi, Houman},
  journal={arXiv preprint arXiv:2509.17154},
  year={2025}
}

@article{jagtap2022physics,
  title={Physics-informed neural networks for inverse problems in supersonic flows},
  author={Jagtap, Ameya D and Mao, Zhiping and Adams, Nikolaus and Karniadakis, George Em},
  journal={Journal of Computational Physics},
  volume={466},
  pages={111402},
  year={2022},
  publisher={Elsevier}
}

@article{chen2021physics,
  title={Physics-informed learning of governing equations from scarce data},
  author={Chen, Zhao and Liu, Yang and Sun, Hao},
  journal={Nature communications},
  volume={12},
  number={1},
  pages={6136},
  year={2021},
  publisher={Nature Publishing Group UK London}
}

@article{schuster2024review,
  title={Review of physics-informed machine-learning inversion of geophysical data},
  author={Schuster, Gerard T and Chen, Yuqing and Feng, Shihang},
  journal={Geophysics},
  volume={89},
  number={6},
  pages={T337--T356},
  year={2024},
  publisher={Society of Exploration Geophysicists}
}

@article{song2021wavefield,
  title={Wavefield reconstruction inversion via physics-informed neural networks},
  author={Song, Chao and Alkhalifah, Tariq A},
  journal={IEEE Transactions on Geoscience and Remote Sensing},
  volume={60},
  pages={1--12},
  year={2021},
  publisher={IEEE}
}

@article{huang2024microseismic,
  title={Microseismic source imaging using physics-informed neural networks with hard constraints},
  author={Huang, Xinquan and Alkhalifah, Tariq A},
  journal={IEEE Transactions on Geoscience and Remote Sensing},
  volume={62},
  pages={1--11},
  year={2024},
  publisher={IEEE}
}

@article{raissi2019deep,
  title={A deep learning framework for solving forward and inverse problems involving nonlinear partial differential equations},
  author={Raissi, M and Perdikaris, P and Karniadakis, GE},
  journal={J. Comput. Phys},
  volume={378},
  pages={686--707},
  year={2019}
}

@article{karniadakis2021physics,
  title={Physics-informed machine learning},
  author={Karniadakis, George Em and Kevrekidis, Ioannis G and Lu, Lu and Perdikaris, Paris and Wang, Sifan and Yang, Liu},
  journal={Nature Reviews Physics},
  volume={3},
  number={6},
  pages={422--440},
  year={2021},
  publisher={Nature Publishing Group UK London}
}

@article{yu2022gradient,
  title={Gradient-enhanced physics-informed neural networks for forward and inverse PDE problems},
  author={Yu, Jeremy and Lu, Lu and Meng, Xuhui and Karniadakis, George Em},
  journal={Computer Methods in Applied Mechanics and Engineering},
  volume={393},
  pages={114823},
  year={2022},
  publisher={Elsevier}
}

@article{zhang2025annealed,
  title={Annealed adaptive importance sampling method in PINNs for solving high dimensional partial differential equations},
  author={Zhang, Zhengqi and Li, Jing and Liu, Bin},
  journal={Journal of Computational Physics},
  volume={521},
  pages={113561},
  year={2025},
  publisher={Elsevier}
}

@article{hwang2019bayesian,
  title={Bayesian pollution source identification via an inverse physics model},
  author={Hwang, Youngdeok and Kim, Hang J and Chang, Won and Yeo, Kyongmin and Kim, Yongku},
  journal={Computational Statistics \& Data Analysis},
  volume={134},
  pages={76--92},
  year={2019},
  publisher={Elsevier}
}

@article{garcia2021simultaneous,
  title={Simultaneous model calibration and source inversion in atmospheric dispersion models},
  author={Garc{\'\i}a, Juan G and Hosseini, Bamdad and Stockie, John M},
  journal={Pure and Applied Geophysics},
  volume={178},
  pages={757--776},
  year={2021},
  publisher={Springer}
}

@article{hosseini2016bayesian,
  title={Bayesian estimation of airborne fugitive emissions using a Gaussian plume model},
  author={Hosseini, Bamdad and Stockie, John M},
  journal={Atmospheric Environment},
  volume={141},
  pages={122--138},
  year={2016},
  publisher={Elsevier}
}

@article{hosseini2017estimating,
  title={Estimating airborne particulate emissions using a finite-volume forward solver coupled with a Bayesian inversion approach},
  author={Hosseini, Bamdad and Stockie, John M},
  journal={Computers \& Fluids},
  volume={154},
  pages={27--43},
  year={2017},
  publisher={Elsevier}
}

@article{albani2021uncertainty,
  title={Uncertainty quantification and atmospheric source estimation with a discrepancy-based and a state-dependent adaptative MCMC},
  author={Albani, Roseane AS and Albani, Vinicius VL and Migon, H{\'e}lio S and Neto, Ant{\^o}nio J Silva},
  journal={Environmental Pollution},
  volume={290},
  pages={118039},
  year={2021},
  publisher={Elsevier}
}

@article{azevedo2014adaptive,
  title={An adaptive Monte Carlo Markov chain method applied to the flow involving self-similar processes in porous media},
  author={Azevedo, Juarez and Cardoso, Gildeberto S and Schnitman, Leizer},
  journal={Journal of Porous Media},
  volume={17},
  number={3},
  year={2014},
  publisher={Begel House Inc.}
}

@article{albani2019tikhonov,
  title={Tikhonov-type regularization and the finite element method applied to point source estimation in the atmosphere},
  author={Albani, Roseane AS and Albani, Vinicius VL},
  journal={Atmospheric Environment},
  volume={211},
  pages={69--78},
  year={2019},
  publisher={Elsevier}
}

@article{albani2020accurate,
  title={An accurate strategy to retrieve multiple source emissions in the atmosphere},
  author={Albani, Roseane AS and Albani, Vinicius VL},
  journal={Atmospheric Environment},
  volume={233},
  pages={117579},
  year={2020},
  publisher={Elsevier}
}

@article{addepalli2011source,
  title={Source characterization of atmospheric releases using stochastic search and regularized gradient optimization},
  author={Addepalli, B and Sikorski, K and Pardyjak, ER and Zhdanov, MS},
  journal={Inverse Problems in Science and Engineering},
  volume={19},
  number={8},
  pages={1097--1124},
  year={2011},
  publisher={Taylor \& Francis}
}

@article{raissi2017physics,
  title={Physics informed deep learning (part i): Data-driven solutions of nonlinear partial differential equations},
  author={Raissi, Maziar and Perdikaris, Paris and Karniadakis, George Em},
  journal={arXiv preprint arXiv:1711.10561},
  year={2017}
}

@article{rudy2017data,
  title={Data-driven discovery of partial differential equations},
  author={Rudy, Samuel H and Brunton, Steven L and Proctor, Joshua L and Kutz, J Nathan},
  journal={Science advances},
  volume={3},
  number={4},
  pages={e1602614},
  year={2017},
  publisher={American Association for the Advancement of Science}
}

@article{lu2021deepxde,
  title={DeepXDE: A deep learning library for solving differential equations},
  author={Lu, Lu and Meng, Xuhui and Mao, Zhiping and Karniadakis, George Em},
  journal={SIAM review},
  volume={63},
  number={1},
  pages={208--228},
  year={2021},
  publisher={SIAM}
}

@article{jacot2018neural,
  title={Neural tangent kernel: Convergence and generalization in neural networks},
  author={Jacot, Arthur and Gabriel, Franck and Hongler, Cl{\'e}ment},
  journal={Advances in neural information processing systems},
  volume={31},
  year={2018}
}

@article{stockie2011mathematics,
  title={The mathematics of atmospheric dispersion modeling},
  author={Stockie, John M},
  journal={Siam Review},
  volume={53},
  number={2},
  pages={349--372},
  year={2011},
  publisher={SIAM}
}

@article{lin1996analytical,
  title={Analytical solutions of the atmospheric diffusion equation with multiple sources and height-dependent wind speed and eddy diffusivities},
  author={Lin, Jin-Sheng and Hildemann, Lynn M},
  journal={Atmospheric Environment},
  volume={30},
  number={2},
  pages={239--254},
  year={1996},
  publisher={Elsevier}
}

@article{hosseini2016airborne,
  title={Airborne contaminant source estimation using a finite-volume forward solver coupled with a Bayesian inversion approach},
  author={Hosseini, Bamdad and Stockie, John M},
  journal={arXiv preprint arXiv:1607.03518},
  year={2016}
}

@book{arya1999air,
  title={Air pollution meteorology and dispersion},
  author={Arya, S Pal and others},
  volume={310},
  year={1999},
  publisher={Oxford University Press New York}
}

@article{Hu2024,
  title = {Tackling the curse of dimensionality with physics-informed neural networks},
  volume = {176},
  ISSN = {0893-6080},
  journal = {Neural Networks},
  publisher = {Elsevier BV},
  author = {Hu,  Zheyuan and Shukla,  Khemraj and Karniadakis,  George Em and Kawaguchi,  Kenji},
  year = {2024},
  month = aug,
  pages = {106369}
}

@article{taylor1915eddy,
  title={I. Eddy motion in the atmosphere},
  author={Taylor, Geoffrey Ingram},
  journal={Philosophical Transactions of the Royal Society of London. Series A, Containing Papers of a Mathematical or Physical Character},
  volume={215},
  number={523-537},
  pages={1--26},
  year={1915},
  publisher={The Royal Society London}
}

@article{ShankarRao2007,
  title = {Source estimation methods for atmospheric dispersion},
  volume = {41},
  ISSN = {1352-2310},
  number = {33},
  journal = {Atmospheric Environment},
  publisher = {Elsevier BV},
  author = {Shankar Rao,  K.},
  year = {2007},
  month = oct,
  pages = {6964–6973}
}

@book{enting2002inverse,
  title={Inverse problems in atmospheric constituent transport},
  author={Enting, Ian G},
  year={2002},
  publisher={Cambridge University Press}
}

@article{huang2015source,
  title={Source area identification with observation from limited monitor sites for air pollution episodes in industrial parks},
  author={Huang, Zihan and Wang, Yuan and Yu, Qi and Ma, Weichun and Zhang, Yan and Chen, Limin},
  journal={Atmospheric Environment},
  volume={122},
  pages={1--9},
  year={2015},
  publisher={Elsevier}
}

@article{Senocak2008,
  title = {Stochastic event reconstruction of atmospheric contaminant dispersion using Bayesian inference},
  volume = {42},
  ISSN = {1352-2310},
  number = {33},
  journal = {Atmospheric Environment},
  publisher = {Elsevier BV},
  author = {Senocak,  Inanc and Hengartner,  Nicolas W. and Short,  Margaret B. and Daniel,  W. Brent},
  year = {2008},
  month = oct,
  pages = {7718–7727}
}

@article{Wade2013,
  title = {Stochastic reconstruction of multiple source atmospheric contaminant dispersion events},
  volume = {74},
  ISSN = {1352-2310},
  journal = {Atmospheric Environment},
  publisher = {Elsevier BV},
  author = {Wade,  Derek and Senocak,  Inanc},
  year = {2013},
  month = aug,
  pages = {45–51}
}

@article{wang2021understanding,
  title={Understanding and mitigating gradient flow pathologies in physics-informed neural networks},
  author={Wang, Sifan and Teng, Yujun and Perdikaris, Paris},
  journal={SIAM Journal on Scientific Computing},
  volume={43},
  number={5},
  pages={A3055--A3081},
  year={2021},
  publisher={SIAM}
}

@article{liu2021dual,
  title={A dual-dimer method for training physics-constrained neural networks with minimax architecture},
  author={Liu, Dehao and Wang, Yan},
  journal={Neural Networks},
  volume={136},
  pages={112--125},
  year={2021},
  publisher={Elsevier}
}

@article{wight2020solving,
  title={Solving Allen-Cahn and Cahn-Hilliard equations using the adaptive physics informed neural networks},
  author={Wight, Colby L and Zhao, Jia},
  journal={arXiv preprint arXiv:2007.04542},
  year={2020}
}

@article{braga2021self,
  title={Self-Adaptive Physics-Informed Neural Networks using a Soft Attention Mechanism},
  author={Braga-Neto, Levi McClenny Ulisses},
  year={2021}
}

@article{wang2021eigenvector,
  title={On the eigenvector bias of Fourier feature networks: From regression to solving multi-scale PDEs with physics-informed neural networks},
  author={Wang, Sifan and Wang, Hanwen and Perdikaris, Paris},
  journal={Computer Methods in Applied Mechanics and Engineering},
  volume={384},
  pages={113938},
  year={2021},
  publisher={Elsevier}
}

@article{faroughi2025neural,
  title={Neural Tangent Kernel Analysis to Probe Convergence in Physics-informed Neural Solvers: PIKANs vs. PINNs},
  author={Faroughi, Salah A and Mostajeran, Farinaz},
  journal={arXiv preprint arXiv:2506.07958},
  year={2025}
}

@book{braess2001finite,
  title={Finite elements: Theory, fast solvers, and applications in solid mechanics},
  author={Braess, Dietrich},
  year={2001},
  publisher={Cambridge University Press}
}

@book{strikwerda2004finite,
  title={Finite difference schemes and partial differential equations},
  author={Strikwerda, John C},
  year={2004},
  publisher={SIAM}
}

@misc{TorchPhysics,
  author = {TorchPhysics},
  title = {TorchPhysics, a Python library of (mesh-free) deep learning methods to solve differential equations},
  year = {2023},
  publisher = {GitHub},
  howpublished = {https://github.com/boschresearch/torchphysics}}

@article{mishra2022estimates,
  title={Estimates on the generalization error of physics-informed neural networks for approximating a class of inverse problems for PDEs},
  author={Mishra, Siddhartha and Molinaro, Roberto},
  journal={IMA Journal of Numerical Analysis},
  volume={42},
  number={2},
  pages={981--1022},
  year={2022},
  publisher={Oxford University Press}
}

@article{yamamoto2009carleman,
  title={Carleman estimates for parabolic equations and applications},
  author={Yamamoto, Masahiro},
  journal={Inverse problems},
  volume={25},
  number={12},
  pages={123013},
  year={2009}
}

@article{zhang2025convergence,
  title={Convergence analysis of a PINNs-based approach to the inverse source problem of the heat equation with local measurements},
  author={Zhang, Xuezhao and Li, Zhiyuan},
  journal={Journal of Computational and Applied Mathematics},
  pages={117077},
  year={2025},
  publisher={Elsevier}
}

@article{hu2025conditional,
  title={Conditional stability and numerical reconstruction of a parabolic inverse source problem using Carleman estimates},
  author={Hu, Tianhao and Huang, Xinchi and Jin, Bangti and Quan, Qimeng and Zhou, Zhi},
  journal={arXiv preprint arXiv:2508.15406},
  year={2025}
}

@article{naderi2024towards,
  title={Towards chemical source tracking and characterization using physics-informed neural networks},
  author={Naderi, Forouzan and Perez-Raya, Issac and Yadav, Sangeeta and Kalajahi, Amin Pashaei and Mamun, Zayeed Bin and D’Souza, Roshan M},
  journal={Atmospheric Environment},
  volume={334},
  pages={120679},
  year={2024},
  publisher={Elsevier}
}

\appendix
\section{Proof of lemma~\ref{lem:ntk}} \label{sec:prooflemma}
\begin{proof}
We recall the loss function to minimize:
\begin{equation}
     \mathcal{L}(\pmb{\Theta}) = \lambda_{r}\mathcal{L}_{r}(\pmb{\Theta}) + \lambda_{b}\mathcal{L}_{b}(\pmb{\theta}_{u}) + \lambda_{z}\mathcal{L}_{z}(\pmb{\theta}_{u}), \quad \pmb{\Theta}=\{\pmb{\theta}_{u}, \pmb{\theta}_{f}, \gamma\}
    \label{loss2}
\end{equation}
   
and the gradient flow system $\frac{d\pmb{\Theta}(s)}{ds}= -\nabla_{\pmb{\Theta}}\mathcal{L}(\pmb{\Theta}(s)), \quad \text{with } \pmb{\Theta}(s)=\{\pmb{\theta}_{u}(s), \pmb{\theta}_{f}(s), \gamma(s)\}$ which is explicitly written as
\begin{eqnarray}
\begin{bmatrix}
     \frac{d\pmb{\theta}_{u}(s)}{ds}\\
     \frac{d\pmb{\theta}_{f}(s)}{ds}\\
     \frac{d\gamma(s)}{ds}
\end{bmatrix}=-\begin{bmatrix}
     \nabla_{\pmb{\theta}_{u}}\mathcal{L}(\pmb{\Theta}(s))\\
     \nabla_{\pmb{\theta}_{f}}\mathcal{L}(\pmb{\Theta}(s))\\
     \nabla_{\gamma}\mathcal{L}(\pmb{\Theta}(s))
\end{bmatrix}
    \label{gradflow2}
\end{eqnarray}
By replacing (\ref{loss2}) in (\ref{gradflow2}), we have
\begin{eqnarray*}
\frac{d\pmb{\theta}_{{u}}(s)}{ds}&=& - \nabla_{\pmb{\theta}_{u}}\mathcal{L}(\pmb{\Theta}(s))\\
&=& -\frac{2\lambda_{r}}{N_{r}} \sum_{i=1}^{N_{r}}\bigg( \mathcal{F}[u(\mathbf{x}^{i}_{r},t^{i}_{r};\pmb{\theta}_{u}(s));\gamma(s)]-f(\mathbf{x}^{i}_{r},t^{i}_{r},\pmb{\theta}_{f}(s))\bigg)\frac{\partial\mathcal{F}[u(\mathbf{x}^{i}_{r},t^{i}_{r};\pmb{\theta}_{u}(s));\gamma(s)]}{\partial\pmb{\theta}_{u}}\\
&-& \frac{2\lambda_{b}}{N_{b}} \sum_{i=1}^{N_{b}}\bigg(\mathcal{B}[u(\mathbf{x}^{i}_{b},t^{i}_{b};\pmb{\theta}_{u}(s))-h(\mathbf{x}^{i}_{b},t^{i}_{b})]\bigg)\frac{\partial \mathcal{B}[u(\mathbf{x}^{i}_{b},t^{i}_{b};\pmb{\theta}_{u}(s))]}{\partial \pmb{\theta}_{u}}\\
&-& \frac{2\lambda_{z}}{N_{z}} \sum_{i=1}^{N_{z}} \bigg(\tau_{i}[u(\mathbf{x}^{i}_{z},t^{i}_{z};\pmb{\theta}_{u}(s))-\mathbf{z}^\ast_{i}]\bigg)\frac{\partial \tau_{i}[u(\mathbf{x}^{i}_{z},t^{i}_{z};\pmb{\theta}_{u}(s))]}{\partial \pmb{\theta}_{u}}
\end{eqnarray*}
Similarly, we have
\begin{eqnarray*}
\frac{d\pmb{\theta}_{f}(s)}{ds}&=& \frac{2\lambda_{r}}{N_{r}} \sum_{i=1}^{N_{r}}\bigg( \mathcal{F}[u(\mathbf{\mathbf{x}}^{i}_{r},t^{i}_{r};\pmb{\theta}_{u}(s));\gamma(s)]-f(\mathbf{x}^{i}_{r},t^{i}_{r},\pmb{\theta}_{f}(s))\bigg)\frac{\partial f(\mathbf{x}^{i}_{r},t^{i}_{r},\pmb{\theta}_{f}(s))}{\partial\pmb{\theta}_{f}}\\
\frac{d\gamma(s)}{ds}&=& -\frac{2\lambda_{r}}{N_{r}} \sum_{i=1}^{N_{r}}\bigg( \mathcal{F}[u(\mathbf{x}^{i}_{r},t^{i}_{r};\pmb{\theta}_{u}(s));\gamma(s)]-f(\mathbf{x}^{i}_{r},t^{i}_{r},\pmb{\theta}_{f}(s))\bigg)\frac{\partial \mathcal{F}[u(\mathbf{x}^{i}_{r},t^{i}_{r};\pmb{\theta}_{u}(s));\gamma(s)]}{\partial\pmb{\theta}_{f}}
\end{eqnarray*}
For $0\leq j \leq N_{r}$, we have
\begin{align*}
    & \frac{d\left(\mathcal{F}[u(\mathbf{x}^{j}_{r},t^{j}_{r};\pmb{\theta}_{u}(s));\gamma(s)]-f(\mathbf{x}^{j}_{r},t^{j}_{r},\pmb{\theta}_{f}(s))\right)}{ds} \\
    &=  \frac{d\left(\mathcal{F}[u(\mathbf{x}^{j}_{r},t^{j}_{r};\pmb{\theta}_{u}(s));\gamma(s)]-f(\mathbf{x}^{j}_{r},t^{j}_{r},\pmb{\theta}_{f}(s))\right)}{d\pmb{\Theta}} \cdot \frac{d\pmb{\Theta}(s)}{ds} \\
    &= \begin{bmatrix}
      \frac{d\left(\mathcal{F}[u(\mathbf{x}^{j}_{r},t^{j}_{r};\pmb{\theta}_{u}(s));\gamma(s)]-f(\mathbf{x}^{j}_{r},t^{j}_{r},\pmb{\theta}_{f}(s))\right)}{d\pmb{\theta}_{u}} \\
       \frac{d\left(\mathcal{F}[u(\mathbf{x}^{j}_{r},t^{j}_{r};\pmb{\theta}_{u}(s));\gamma(s)]-f(\mathbf{x}^{j}_{r},t^{j}_{r},\pmb{\theta}_{f}(s))\right)}{d\pmb{\theta}_{f}} \\
       \frac{d\left(\mathcal{F}[u(\mathbf{x}^{j}_{r},t^{j}_{r};\pmb{\theta}_{f}(s));\gamma(s)]-f(\mathbf{x}^{j}_{r},t^{j}_{r},\pmb{\theta}_{f}(s))\right)}{d\gamma}
    \end{bmatrix} \cdot \begin{bmatrix}
     \frac{d\pmb{\theta}_{u}(s)}{ds}\\
     \frac{d\pmb{\theta}_{f}(s)}{ds}\\
     \frac{d\gamma(s)}{ds}
\end{bmatrix}\\
&= \underbrace{\frac{d(\mathcal{F}[u(\mathbf{x}^{j}_{r},t^{j}_{r};\pmb{\theta}_{u}(s));\gamma(s)])}{d\pmb{\theta}_{u}} \times  \frac{d\pmb{\theta}_{u}(s)}{ds}}_{A} -  \underbrace{\frac{d (f(\mathbf{x}^{j}_{r},t^{j}_{r};\pmb{\theta}_{f}(s))}{d\pmb{\theta}_{f}} \times \frac{d\pmb{\theta}_{f}(s)}{ds}}_{B}\\
&+ \underbrace{\frac{d(\mathcal{F}[u(\mathbf{x}^{j}_{r},t^{j}_{r};\pmb{\theta}_{u}(s));\gamma(s)])}{d\mathbf{\gamma}} \times  \frac{d\mathbf{\gamma(s)}}{ds}}_{C}\\
&= A-B+C\\
A&=\frac{d(\mathcal{F}[u(\mathbf{x}^{j}_{r},t^{j}_{r};\pmb{\theta}_{u}(s));\gamma(s)])}{d\pmb{\theta}_{u}} \bigg[-\frac{2\lambda_{r}}{N_{r}} \sum_{i=1}^{N_{r}}\bigg( \mathcal{F}[u(\mathbf{x}^{i}_{r},t^{i}_{r};\pmb{\theta}_{u}(s));\gamma(s)]-f(\mathbf{x}^{i}_{r},t^{i}_{r},\pmb{\theta}_{f}(s))\bigg)\\
&\times\frac{\partial\mathcal{F}[u(\mathbf{x}^{i}_{r},t^{i}_{r};\pmb{\theta}_{u}(s));\gamma(s)]}{\partial\pmb{\theta}_{u}}- \frac{2\lambda_{b}}{N_{b}} \sum_{i=1}^{N_{b}}\bigg(\mathcal{B}[u(\mathbf{x}^{i}_{b},t^{i}_{b};\pmb{\theta}_{u}(s))-h(\mathbf{x}^{i}_{b},t^{i}_{b})]\bigg)\frac{\partial \mathcal{B}[u(\mathbf{x}^{i}_{b},t^{i}_{b};\pmb{\theta}_{u}(s))]}{\partial \pmb{\theta}_{u}}\\
&-\frac{2\lambda_{z}}{N_{z}} \sum_{i=1}^{N_{z}} \bigg(\tau_{i}[u(\mathbf{x}^{i}_{z},t^{i}_{z};\pmb{\theta}_{u}(s))-\mathbf{z}^\ast_{i}]\bigg)\frac{\partial \tau_{i}[u(\mathbf{x}^{i}_{z},t^{i}_{z};\pmb{\theta}_{u}(s))]}{\partial \pmb{\theta}_{u}}\bigg] \\ 
&= -\frac{2\lambda_{r}}{N_{r}} \sum_{i=1}^{N_{r}}\bigg( \mathcal{F}[u(\mathbf{x}^{i}_{r},t^{i}_{r};\pmb{\theta}_{u}(s));\gamma(s)]-f(\mathbf{x}^{i}_{r},t^{i}_{r},\pmb{\theta}_{f}(s))\bigg)\\
&\times\bigg<\frac{\partial(\mathcal{F}[u(\mathbf{x}^{i}_{r},t^{i}_{r};\pmb{\theta}_{u}(s));\gamma(s)])}{\partial\pmb{\theta}_{u}},\frac{\partial(\mathcal{F}[u(\mathbf{x}^{j}_{r},t^{j}_{r};\pmb{\theta}_{u}(s));\gamma(s)])}{\partial\pmb{\theta}_{u}}\bigg>\\
&- \frac{2\lambda_{b}}{N_{b}} \sum_{i=1}^{N_{b}}\big(\mathcal{B}[u(\mathbf{x}^{i}_{b},t^{i}_{b};\pmb{\theta}_{u}(s))-h(\mathbf{x}^{i}_{b},t^{i}_{b})]\big)\bigg< \frac{\partial \mathcal{B}[u(\mathbf{x}^{i}_{b},t^{i}_{b};\pmb{\theta}_{u}(s))]}{\partial \pmb{\theta}_{u}}, \frac{\partial(\mathcal{F}[u(\mathbf{x}^{j}_{r},t^{j}_{r};\pmb{\theta_u}(s));\gamma(s)])}{\partial\pmb{\theta}_{u}}\bigg>\\
  & - \frac{2\lambda_{z}}{N_{z}} \sum_{i=1}^{N_{z}} \big(\tau_{i}[u(\mathbf{x}^{i}_{z},t^{i}_{z};\pmb{\theta}_{u}(s))-\mathbf{z}^\ast_{i}]\big) \bigg< \frac{\partial \tau_{i}[u(\mathbf{x}^{i}_{d},t^{i}_{d};\pmb{\theta}_{u}(s))]}{\partial \pmb{\theta}_{u}},\frac{\partial(\mathcal{F}[u(\mathbf{x}^{j}_{r},t^{j}_{r};\pmb{\theta}_{u}(s));\gamma(s)])}{\partial\pmb{\theta}_{u}}\bigg>\\
&= -\frac{2\lambda_{r}}{N_{r}} \sum_{i=1}^{N_{r}}\bigg( \mathcal{F}[u(\mathbf{x}^{i}_{r},t^{i}_{r};\pmb{\theta}_{u}(s));\gamma(s)]-f(\mathbf{x}^{i}_{r},t^{i}_{r},\pmb{\theta}_{f}(s))\bigg) (J_{rr}^{\pmb{\theta}_{u}}(s))_{ij}\\
&- \frac{2\lambda_{b}}{N_{b}} \sum_{i=1}^{N_{b}}\big(\mathcal{B}[u(\mathbf{x}^{i}_{b},t^{i}_{b};\pmb{\theta}_{u}(s))-h(\mathbf{x}^{i}_{b},t^{i}_{b})]\big)(\mathbf{K}_{br}^{\pmb{\theta}_{u}})_{ij}
- \frac{2\lambda_{z}}{N_{z}} \sum_{i=1}^{N_{z}} \big(\tau_{i}[u(\mathbf{x}^{i}_{z},t^{i}_{z};\pmb{\theta}_{u}(s))-\mathbf{z}^\ast_{i}]\big) (\mathbf{K}_{zr}^{\pmb{\theta}_{u}})_{ij}\\
B&= \frac{d (f(\mathbf{x}^{j}_{r},t^{j}_{r}),\pmb{\theta}_{f}(s))}{d\pmb{\theta}_{f}}\bigg[ \frac{2\lambda_{r}}{N_{r}} \sum_{i=1}^{N_{r}}\bigg( \mathcal{F}[u(\mathbf{x}^{i}_{r},t^{i}_{r};\pmb{\theta}_{u}(s));\gamma(s)]-f(\mathbf{x}^{i}_{r},t^{i}_{r},\pmb{\theta}_{f}(s))\bigg)\frac{\partial f(\mathbf{x}^{i}_{r},t^{i}_{r},\pmb{\theta}_{f}(s))}{\partial\pmb{\theta}_{f}}\bigg]\\
&= \frac{2\lambda_{r}}{N_{r}} \sum_{i=1}^{N_{r}}\bigg( \mathcal{F}[u(\mathbf{x}^{i}_{r},t^{i}_{r};\pmb{\theta}_{u}(s));\gamma(s)]-f(\mathbf{x}^{i}_{r},t^{i}_{r},\pmb{\theta}_{f}(s))\bigg)\bigg<\frac{\partial f(\mathbf{x}^{i}_{r},t^{i}_{r},\pmb{\theta}_{f}(s))}{\partial\pmb{\theta}_{f}},\frac{\partial f(\mathbf{x}^{j}_{r},t^{j}_{r},\pmb{\theta}_{f}(s))}{\partial\pmb{\theta}_{f}}\bigg>\\
&= \frac{2\lambda_{r}}{N_{r}} \sum_{i=1}^{N_{r}}\bigg( \mathcal{F}[u(\mathbf{x}^{i}_{r},t^{i}_{r};\pmb{\theta}_{u}(s));\gamma(s)]-f(\mathbf{x}^{i}_{r},t^{i}_{r},\pmb{\theta}_{f}(s))\bigg)(J_{rr}^{\pmb{\theta}_{f}}(s))_{ij}) 
\end{align*}
\begin{align*}
C&= - \frac{2\lambda_{r}}{N_{r}} \sum_{i=1}^{N_{r}}\big( \mathcal{F}[u(\mathbf{x}^{i}_{r},t^{i}_{r};\pmb{\theta}_{u}(s));\gamma(s)]-f(\mathbf{x}^{i}_{r},t^{i}_{r},\pmb{\theta}_{f}(s))\big)\\
&\times\bigg<\frac{\partial(\mathcal{F}[u(\mathbf{x}^{i}_{r},t^{i}_{r};\pmb{\theta}_{u}(s));\gamma(s)])}{\partial\gamma},\frac{\partial(\mathcal{F}[u(\mathbf{x}^{j}_{r},t^{j}_{r};\pmb{\theta}_{u}(s));\gamma(s)])}{\partial\gamma}\bigg>
\end{align*}
Therefore we have 
\begin{align*}
&\frac{d\left(\mathcal{F}[u(\mathbf{x}^{j}_{r},t^{j}_{r};\pmb{\theta}_{u}(s));\gamma(s)]-f(\mathbf{x}^{j}_{r},t^{j}_{r},\pmb{\theta}_{f}(s))\right)}{ds}\\
  &= -\frac{2\lambda_{r}}{N_{r}} \sum_{i=1}^{N_{r}}\left( \mathcal{F}[u(\mathbf{x}^{i}_{r},t^{i}_{r};\pmb{\theta}_{u}(s));\gamma(s)]-f(\mathbf{x}^{i}_{r},t^{i}_{r},\pmb{\theta}_{f}(s))\right) \left((J_{rr}^{\pmb{\theta}_{u}}(s))_{ij})+(J_{rr}^{\pmb{\theta}_{f}}(s))_{ij})+ (J_{rr}^{\gamma}(s))_{ij})\right) \\
 &- \frac{2\lambda_{b}}{N_{b}} \sum_{i=1}^{N_{b}}\big(\mathcal{B}[u(\mathbf{x}^{i}_{b},t^{i}_{b};\pmb{\theta}_{u}(s))-h(\mathbf{x}^{i}_{b},t^{i}_{b})]\big)(\mathbf{K}_{rb}^{\pmb{\theta}_{u}})_{ij}
- \frac{2\lambda_{z}}{N_{z}} \sum_{i=1}^{N_{z}} \left(\tau_{i}[u(\mathbf{x}^{i}_{z},t^{i}_{z};\pmb{\theta}_{u}(s))-\mathbf{z}^\ast_{i}]\right) (\mathbf{K}_{rz}^{\pmb{\theta}_{u}})_{ij}\\
    &=- \frac{2\lambda_{r}}{N_{r}} \sum_{i=1}^{N_{r}}\left( \mathcal{F}[u(\mathbf{x}^{i}_{r},t^{i}_{r};\pmb{\theta}_{u}(s));\gamma(s)]-f(\mathbf{x}^{i}_{r},t^{i}_{r},\pmb{\theta}_{f}(s))\right)\\
    &\times \bigg<\begin{pmatrix}
        \frac{\partial(\mathcal{F}[u(\mathbf{x}^{i}_{r},t^{i}_{r};\pmb{\theta}_{u}(s));\gamma(s)])}{\partial\pmb{\theta}_{u}}\\
        \frac{\partial f(\mathbf{x}^{i}_{r},t^{i}_{r},\pmb{\theta}_{f}(s))}{\partial\pmb{\theta}_{f}}\\
         \frac{\partial(\mathcal{F}[u(\mathbf{\mathbf{x}}^{i}_{r},t^{i}_{r};\pmb{\theta}_{u}(s));\gamma(s)])}{\partial\gamma}
    \end{pmatrix},\begin{pmatrix}
        \frac{\partial(\mathcal{F}[u(\mathbf{x}^{j}_{r},t^{j}_{r};\pmb{\theta}_{u});\gamma(s)])}{\partial\pmb{\theta}_{u}}\\
        \frac{\partial f(\mathbf{x}^{j}_{r},t^{j}_{r},\pmb{\theta}_{f}(s))}{\partial\pmb{\theta}_{f}}\\
         \frac{\partial(\mathcal{F}[u(\mathbf{x}^{i}_{r},t^{i}_{r};\pmb{\theta}_{u}(s));\gamma(s)])}{\partial\gamma}
    \end{pmatrix}\bigg> \\
    &- \frac{2\lambda_{b}}{N_{b}} \sum_{i=1}^{N_{b}}\big(\mathcal{B}[u(\mathbf{x}^{i}_{b},t^{i}_{b};\pmb{\theta}_{u}(s))-h(\mathbf{x}^{i}_{b},t^{i}_{b})]\big)(\mathbf{K}_{br}^{\pmb{\theta}_{u}})_{ij}
- \frac{2\lambda_{z}}{N_{z}} \sum_{i=1}^{N_{z}} \big(\tau_{i}[u(\mathbf{x}^{i}_{z},t^{i}_{z};\pmb{\theta}_{u}(s))-\mathbf{z}^\ast_{i}]\big) (\mathbf{K}_{zr}^{\pmb{\theta}_{u}})_{ij}\\
\end{align*}
and we denote $(\mathbf{K}_{rr}(s)_{ij}= \bigg<\begin{pmatrix}
        \frac{\partial(\mathcal{F}[u(\mathbf{x}^{i}_{r},t^{i}_{r};\pmb{\theta}_{u}(s));\gamma(s)])}{\partial\pmb{\theta}_{u}}\\
        \frac{\partial f(\mathbf{x}^{i}_{r},t^{i}_{r},\pmb{\theta}_{f}(s))}{\partial\pmb{\theta}_{f}}\\
         \frac{\partial(\mathcal{F}[u(\mathbf{x}^{i}_{r},t^{i}_{r};\pmb{\theta}_{u}(s));\gamma(s)])}{\partial\gamma}
    \end{pmatrix},\begin{pmatrix}
        \frac{\partial(\mathcal{F}[u(\mathbf{x}^{j}_{r},t^{j}_{r};\pmb{\theta}_{u}(s));\gamma(s)])}{\partial\pmb{\theta}_{u}}\\
        \frac{\partial f(\mathbf{x}^{j}_{r},t^{j}_{r},\pmb{\theta}_{f}(s))}{\partial\pmb{\theta}_{f}}\\
         \frac{\partial(\mathcal{F}[u(\mathbf{x}^{i}_{r},t^{i}_{r};\pmb{\theta}_{u}(s));\gamma(s)])}{\partial\gamma}
    \end{pmatrix}\bigg>$\\
For $0\leq j \leq N_b$, we have 
\begin{align*}
    \frac{d\mathcal{B}[u(\mathbf{x}^{j}_{b},t^{j}_{b};\pmb{\theta}_{u}(s))]}{ds} &=  \frac{d\mathcal{B}[u(\mathbf{x}^{j}_{b},t^{j}_{b};\pmb{\theta}_{u}(s))]}{d\pmb{\Theta}} \times \frac{d\pmb{\Theta}(s)}{ds} = \frac{d\mathcal{B}[u(\mathbf{x}^{j}_{b},t^{j}_{b};\pmb{\theta}_{u}(s))]}{d\pmb{\theta}_{u}} \times \frac{d\pmb{\theta}_{u}(s)}{ds}
\end{align*}
\begin{align*}
&=\frac{d\mathcal{B}[u(\mathbf{x}^{j}_{b},t^{j}_{b};\pmb{\theta}_{u}(s))]}{d\pmb{\theta}_{u}} \bigg[-\frac{2\lambda_{r}}{N_{r}} \sum_{i=1}^{N_{r}}\bigg( \mathcal{F}[u(\mathbf{x}^{i}_{r},t^{i}_{r};\pmb{\theta}_{u}(s));\gamma(s)]-f(\mathbf{x}^{i}_{r},t^{i}_{r},\pmb{\theta}_{f}(s))\bigg)\\
&\times \frac{\partial\big(\mathcal{F}[u(\mathbf{x}^{i}_{r},t^{i}_{r};\pmb{\theta}_{u}(s));\gamma(s)]\big)}{\partial\pmb{\theta}_{u}}\\
&- \frac{2\lambda_{b}}{N_{b}} \sum_{i=1}^{N_{b}}\big(\mathcal{B}[u(\mathbf{x}^{i}_{b},t^{i}_{b};\pmb{\theta}_{u}(s))-h(\mathbf{x}^{i}_{b},t^{i}_{b})]\big)\frac{\partial \mathcal{B}[u(\mathbf{x}^{i}_{b},t^{i}_{b};\pmb{\theta}_{u}(s))]}{\partial \pmb{\theta}_{u}}\\
&- \frac{2\lambda_{z}}{N_{z}} \sum_{i=1}^{N_{z}} \big(\tau_{i}[u(\mathbf{x}^{i}_{z},t^{i}_{z};\pmb{\theta}_{u}(s))-\mathbf{z}^\ast_{i}]\big)\frac{\partial \tau_{i}[u(\mathbf{x}^{i}_{z},t^{i}_{z};\pmb{\theta}_{u}(s))]}{\partial \pmb{\theta}_{u}} \bigg]\\
&= -\frac{2\lambda_{r}}{N_{r}}\sum_{i=1}^{N_{r}}\big( \mathcal{F}[u(\mathbf{x}^{i}_{r},t^{i}_{r};\pmb{\theta}_{u}(s));\gamma(s)]-f(\mathbf{x}^{i}_{r},t^{i}_{r},\pmb{\theta}_{f}(s))\big)\\
&\times \bigg<\frac{d\big(\mathcal{F}[u(\mathbf{x}^{i}_{r},t^{i}_{r};\pmb{\theta}_{u}(s));\gamma(s)]\big)}{d\pmb{\theta}_{u}},\frac{d\mathcal{B}[u(\mathbf{x}^{j}_{b},t^{j}_{b};\pmb{\theta}_{u}(s))]}{d\pmb{\theta}_{u}}\bigg>\\
&- \frac{2\lambda_{b}}{N_{b}} \sum_{i=1}^{N_{b}}\big(\mathcal{B}[u(\mathbf{x}^{i}_{b},t^{i}_{b};\pmb{\theta}_{u}(s))-h(\mathbf{x}^{i}_{b},t^{i}_{b})]\big) \bigg<\frac{d\mathcal{B}[u(\mathbf{x}^{i}_{b},t^{i}_{b};\pmb{\theta}_{u}(s))]}{d\pmb{\theta}_{u}}, \frac{d\mathcal{B}[u(\mathbf{x}^{j}_{b},t^{j}_{b};\pmb{\theta}_{u}(s))]}{d\pmb{\theta}_{u}} \bigg>\\
&-\frac{2\lambda_{z}}{N_{z}} \sum_{i=1}^{N_{z}} \big(\tau_{i}[u(\mathbf{x}^{i}_{z},t^{i}_{z};\pmb{\theta}_{u}(s))-\mathbf{z}^\ast_{i}]\big) \bigg<\frac{d\tau_{i}[u(\mathbf{x}^{i}_{z},t^{i}_{z};\pmb{\theta}_{u}(s))]}{d\pmb{\theta}_{u}},\frac{d\mathcal{B}[u(\mathbf{x}^{j}_{b},t^{j}_{b};\pmb{\theta}_{u}(s))]}{d\pmb{\theta}_{u}}\bigg>\\
&=  -\frac{2\lambda_{r}}{N_{r}}\sum_{i=1}^{N_{r}}\big(\mathcal{F}[u(\mathbf{x}^{i}_{r},t^{i}_{r};\pmb{\theta}_{u}(s));\gamma(s)]-f(\mathbf{x}^{i}_{r},t^{i}_{r},\pmb{\theta}_{f}(s))\big) (\mathbf{K}_{rb}(s))_{ij}\\
&- \frac{2\lambda_{b}}{N_{b}} \sum_{i=1}^{N_{b}}\big(\mathcal{B}[u(\mathbf{x}^{i}_{b},t^{i}_{b};\pmb{\theta}_{u}(s))-h(\mathbf{x}^{i}_{b},t^{i}_{b})]\big)(\mathbf{K}_{bb}^{\pmb{\theta}_{u}}(s))_{ij}\\
&-\frac{2\lambda_{z}}{N_{z}} \sum_{i=1}^{N_{z}} \big(\tau_{i}[u(\mathbf{x}^{i}_{z},t^{i}_{z};\pmb{\theta}_{u}(s))-\mathbf{z}^\ast_{i}]\big)(\mathbf{K}_{zb}^{\pmb{\theta}_{u}}(s))_{ij}
\end{align*}
For $0\leq j \leq N_z$, we have 
\begin{align*}
& \frac{d\tau_{j}[u(\mathbf{x}^{j}_{b},t^{j}_{b};\pmb{\theta}_{u}(s))]}{ds}\\
&=  \frac{d\tau_{j}[u(\mathbf{x}^{j}_{b},t^{j}_{b};\pmb{\theta}_{u}(s))]}{d\pmb{\theta_{u}}} \times \frac{d\pmb{\theta}_{u}(s)}{ds}\\
&= \frac{d\tau_{j}[u(\mathbf{x}^{j}_{b},t^{j}_{b};\pmb{\theta}_{u}(s))]}{d\mathbf{\theta}_{u}} \bigg[-\frac{2\lambda_{r}}{N_{r}} \sum_{i=1}^{N_{r}}\bigg( \mathcal{F}[u(\mathbf{x}^{i}_{r},t^{i}_{r};\pmb{\theta}_{u}(s));\gamma(s)]-f(\mathbf{x}^{i}_{r},t^{i}_{r};\pmb{\theta}_{f}(s))\bigg)\\
&\times \frac{\partial\big(\mathcal{F}[u(\mathbf{x}^{i}_{r},t^{i}_{r};\pmb{\theta}_{u}(s));\gamma(s)]-f(\mathbf{x}^{i}_{r},t^{i}_{r};\pmb{\theta}_{f}(s))\big)}{\partial\pmb{\theta}_{u}}\\
&- \frac{2\lambda_{b}}{N_{b}} \sum_{i=1}^{N_{b}}\big(\mathcal{B}[u(\mathbf{x}^{i}_{b},t^{i}_{b};\pmb{\theta}_{u}(s))-h(\mathbf{x}^{i}_{b},t^{i}_{b})]\big)\frac{\partial \mathcal{B}[u(\mathbf{x}^{i}_{b},t^{i}_{b};\pmb{\theta}_{u}(s))]}{\partial \pmb{\theta}_{u}}\\
&- \frac{2\lambda_{z}}{N_{z}} \sum_{i=1}^{N_{z}} \big(\tau_{i}[u(\mathbf{x}^{i}_{z},t^{i}_{z};\pmb{\theta}_{u}(s))-\mathbf{z}^\ast_{i}]\big)\frac{\partial \tau_{i}[u(\mathbf{x}^{i}_{z},t^{i}_{z};\pmb{\theta}_{u}(s))]}{\partial \pmb{\theta}_{u}} \bigg]\\
&= -\frac{2\lambda_{r}}{N_{r}}\sum_{i=1}^{N_{r}}\big( \mathcal{F}[u(\mathbf{x}^{i}_{r},t^{i}_{r};\pmb{\theta}_{u}(s));\gamma(s)]-f(\mathbf{x}^{i}_{r},t^{i}_{r};\pmb{\theta}_{f}(s))\big)\\
\times& \bigg<\frac{d\big(\mathcal{F}[u(\mathbf{x}^{j}_{r},t^{j}_{r};\pmb{\theta}_{u}(s));\gamma(s)]- f(\mathbf{x}^{i}_{r},t^{i}_{r}; \pmb{\theta}_{f}(s))\big)}{d\pmb{\theta}_{u}} ,\frac{d\tau_{j}[u(\mathbf{x}^{j}_{b},t^{j}_{b};\pmb{\theta}_{u}(s))]}{d\pmb{\theta}_{u}} \bigg>\\
&- \frac{2\lambda_{b}}{N_{b}} \sum_{i=1}^{N_{b}}\big(\mathcal{B}[u(\mathbf{x}^{i}_{b},t^{i}_{b};\pmb{\theta}_{u}(s))-h(\mathbf{x}^{i}_{b},t^{i}_{b})]\big) \bigg<\frac{d\mathcal{B}[u(\mathbf{x}^{i}_{b},t^{i}_{b};\pmb{\theta}_{u}(s))]}{d\pmb{\theta}_{u}}, \frac{d\tau_{j}[u(\mathbf{x}^{j}_{b},t^{j}_{b};\pmb{\theta}_{u}(s))]}{d\pmb{\theta}_{u}} \bigg>\\
&-\frac{2\lambda_{z}}{N_{z}} \sum_{i=1}^{N_{z}} \big(\tau_{i}[u(\mathbf{x}^{i}_{z},t^{i}_{z};\pmb{\theta}_{u}(s))-\mathbf{z}^\ast_{i}]\big) \bigg<\frac{d\tau_{i}[u(\mathbf{x}^{i}_{z},t^{i}_{z};\pmb{\theta}_{u}(s))]}{d\pmb{\theta}_{u}},\frac{d\tau_{j}[u(\mathbf{x}^{j}_{b},t^{j}_{b};\pmb{\theta}_{u}(s))]}{d\pmb{\theta}_{u}} \bigg>\\
&=  -\frac{2\lambda_{r}}{N_{r}}\sum_{i=1}^{N_{r}}\big( \mathcal{F}[u(\mathbf{x}^{i}_{r},t^{i}_{r};\pmb{\theta}_{u}(s));\gamma(s)]-f(\mathbf{x}^{i}_{r},t^{i}_{r},\pmb{\theta}_{f}(s))\big)(\mathbf{K}_{rz}^{\pmb{\theta}_{u}}(s))_{ij}\\
&- \frac{2\lambda_{b}}{N_{b}} \sum_{i=1}^{N_{b}}\big(\mathcal{B}[u(\mathbf{x}^{i}_{b},t^{i}_{b};\pmb{\theta}_{u}(s))-h(\mathbf{x}^{i}_{b},t^{i}_{b})]\big)\mathbf{K}_{bz}^{\pmb{\theta}_{u}}(s)-\frac{2\lambda_{z}}{N_{z}} \sum_{i=1}^{N_{z}} \big(\tau_{i}[u(\mathbf{x}^{i}_{z},t^{i}_{z};\pmb{\theta}_{u}(s))-\mathbf{z}^\ast_{i}]\big)\mathbf{K}_{zz}^{\pmb{\theta}_{u}}(s).
\end{align*}
\end{proof}

\section{ Numerical results on the variation of Noise for the 2D ADE with constant coefficients using pointwise and accumulative observations}

\begin{figure}[H] 
    \centering
    \setlength{\tabcolsep}{5pt}
    \renewcommand{\arraystretch}{1.2}
    
    \begin{tabular}{c c c c}
       \raisebox{40pt}{\rotatebox[origin=c]{90}{\textbf{$1\%$ noise }}} & 
        \begin{overpic}[width=0.28\textwidth]
           {figures/saved_plots_4Obs/1percent4Obsf_predicted.png}
             \put(1,73){\textbf{Predicted f with 4 obs}}
        \end{overpic} &
         \begin{overpic}[width=0.28\textwidth]
            {figures/saved_plots15Obs/1percent15Obsf_predicted.png}
           \put(1,73){\textbf{Predicted f with 15 obs}}
        \end{overpic} \\
       \raisebox{40pt}{\rotatebox[origin=c]{90}{\textbf{$5\%$ noise}}} &
        \begin{overpic}[width=0.28\textwidth]
           {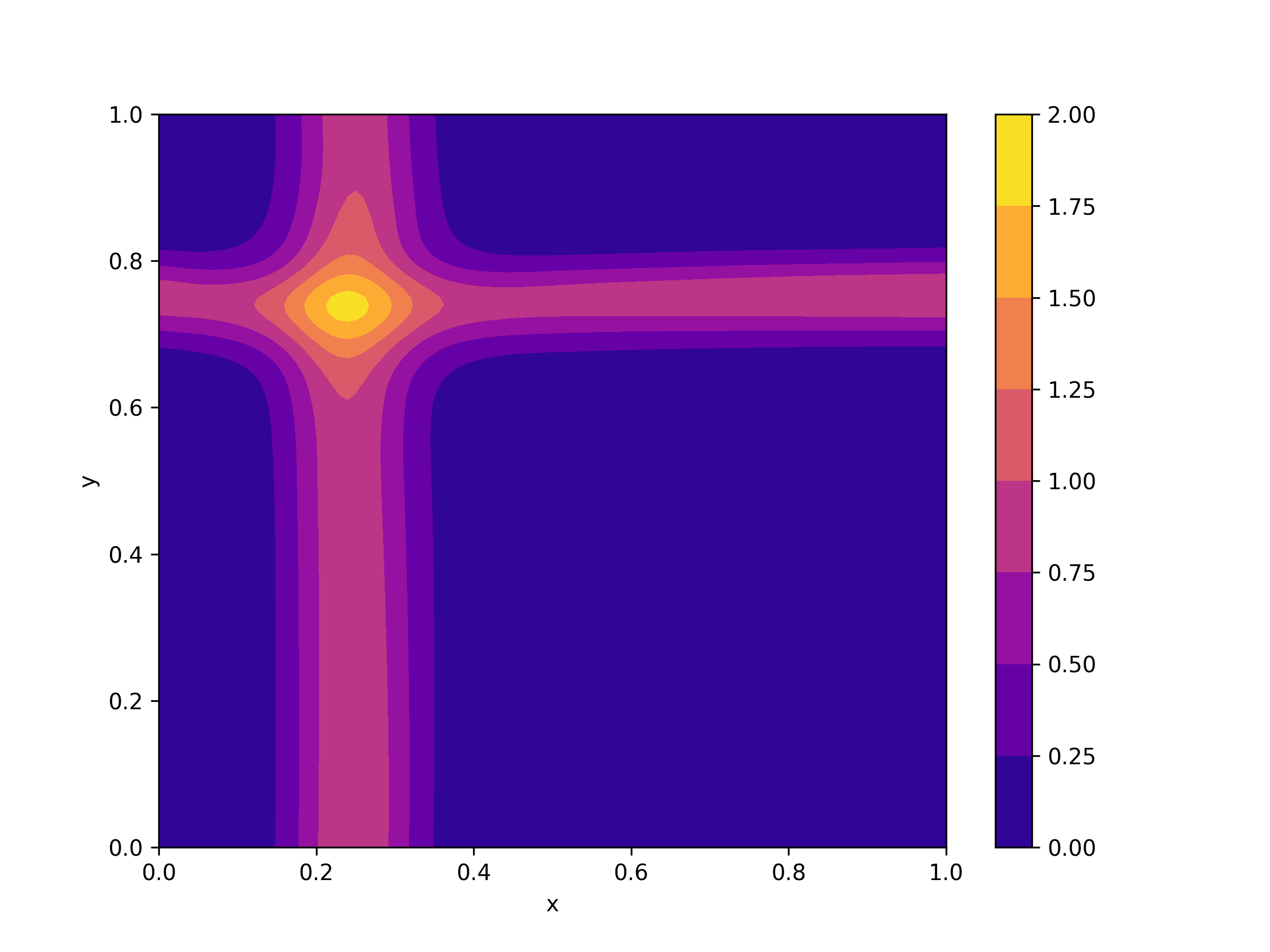}
        \end{overpic} &
        \begin{overpic}[width=0.28\textwidth]
           {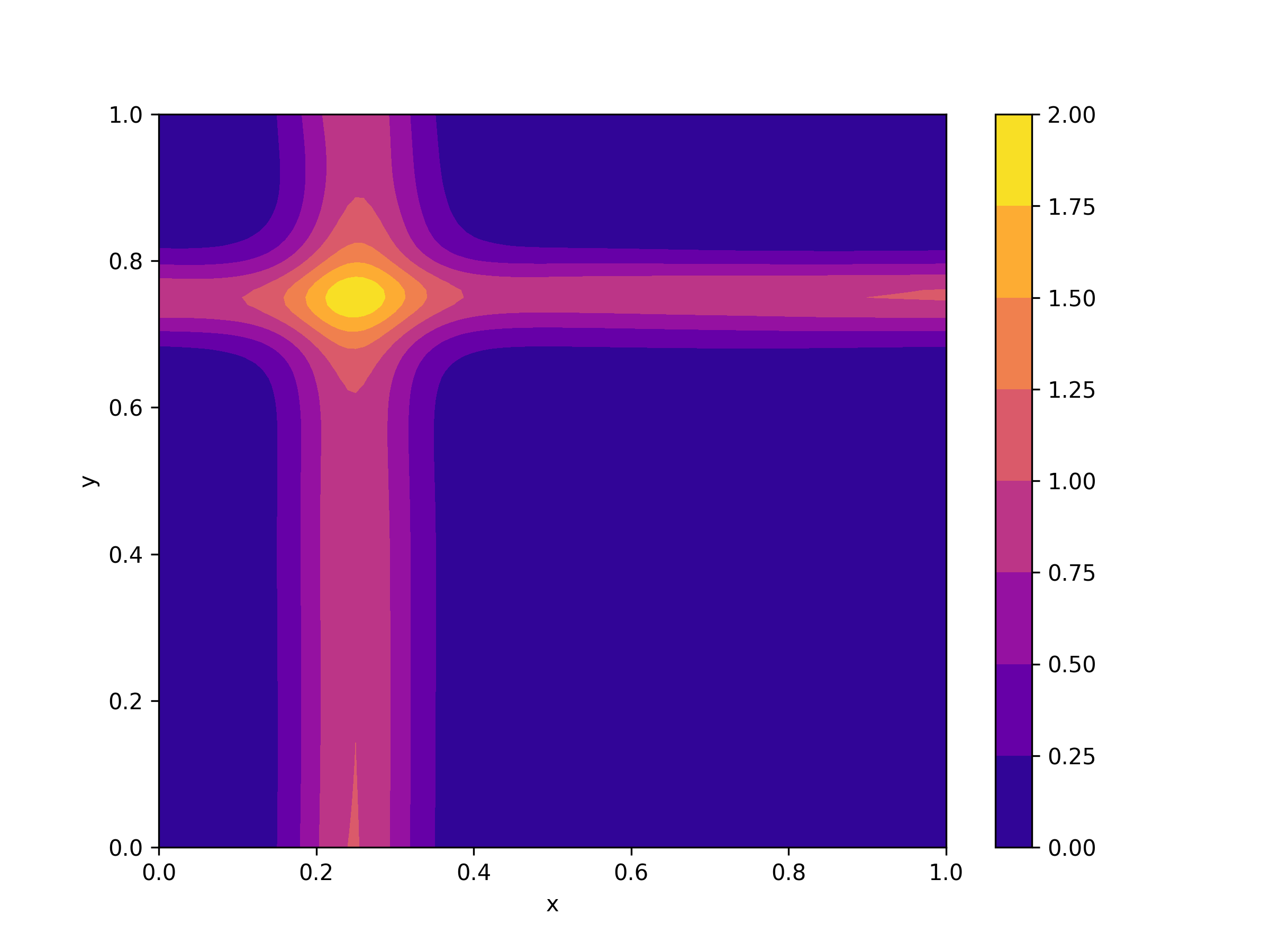}
        \end{overpic} &
        \begin{overpic}[width=0.28\textwidth]
            {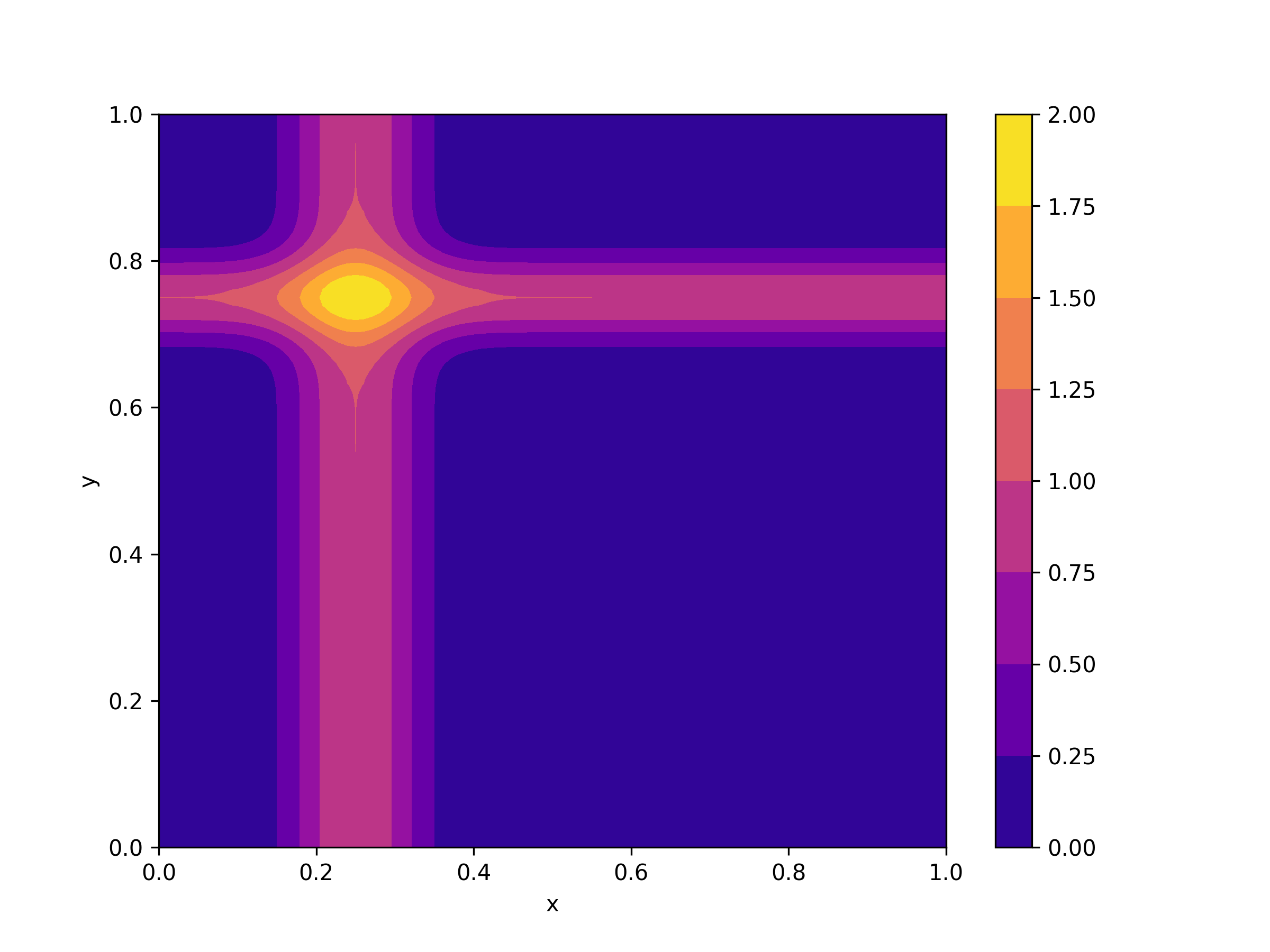}
            \put(20,73){\textbf{Exact f}}
        \end{overpic} \\ 
        \raisebox{40pt}{\rotatebox[origin=c]{90}{\textbf{$10\%$ noise}}} &
        \begin{overpic}[width=0.28\textwidth]
            {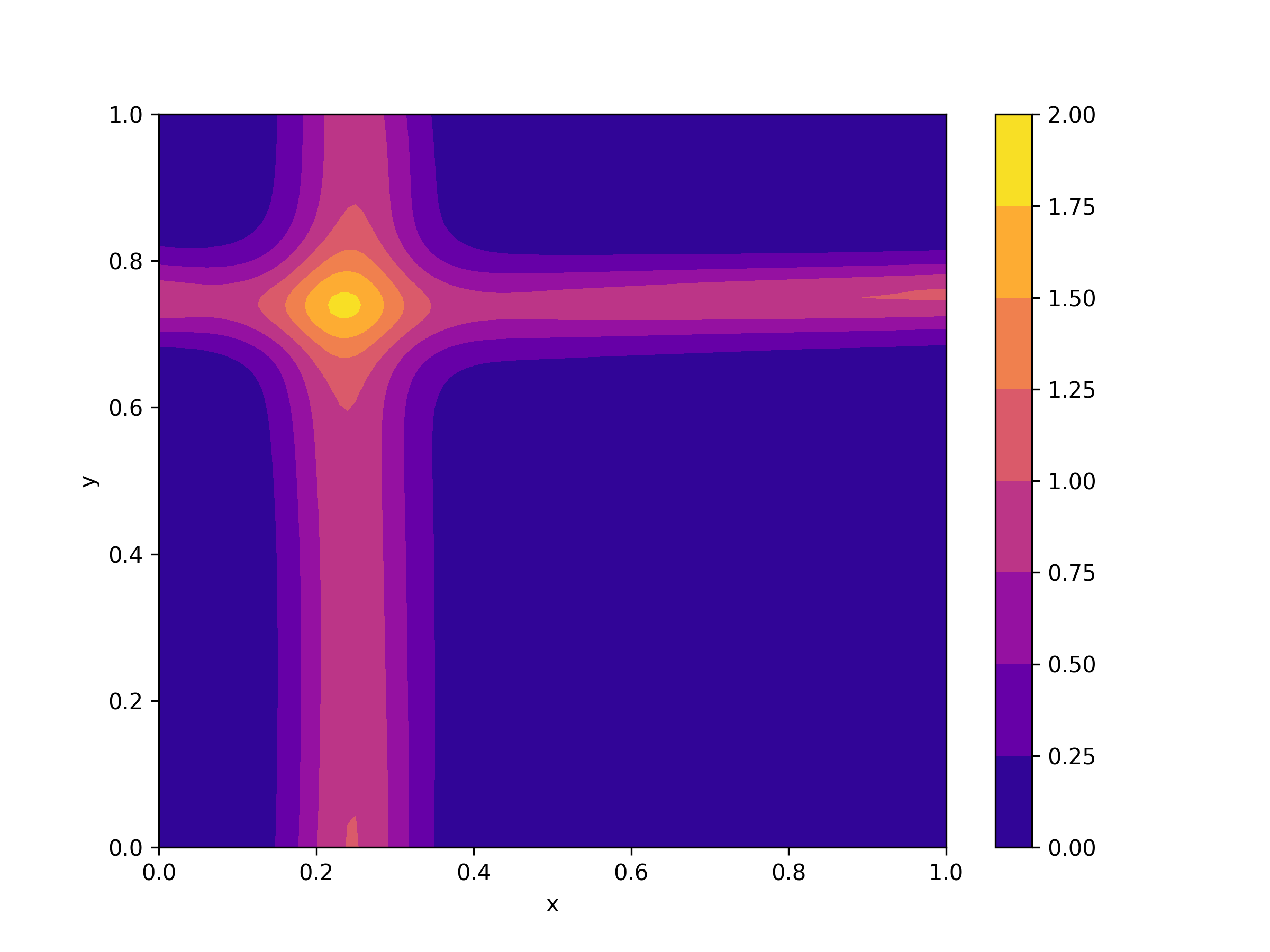}
        \end{overpic} &
        \begin{overpic}[width=0.28\textwidth]{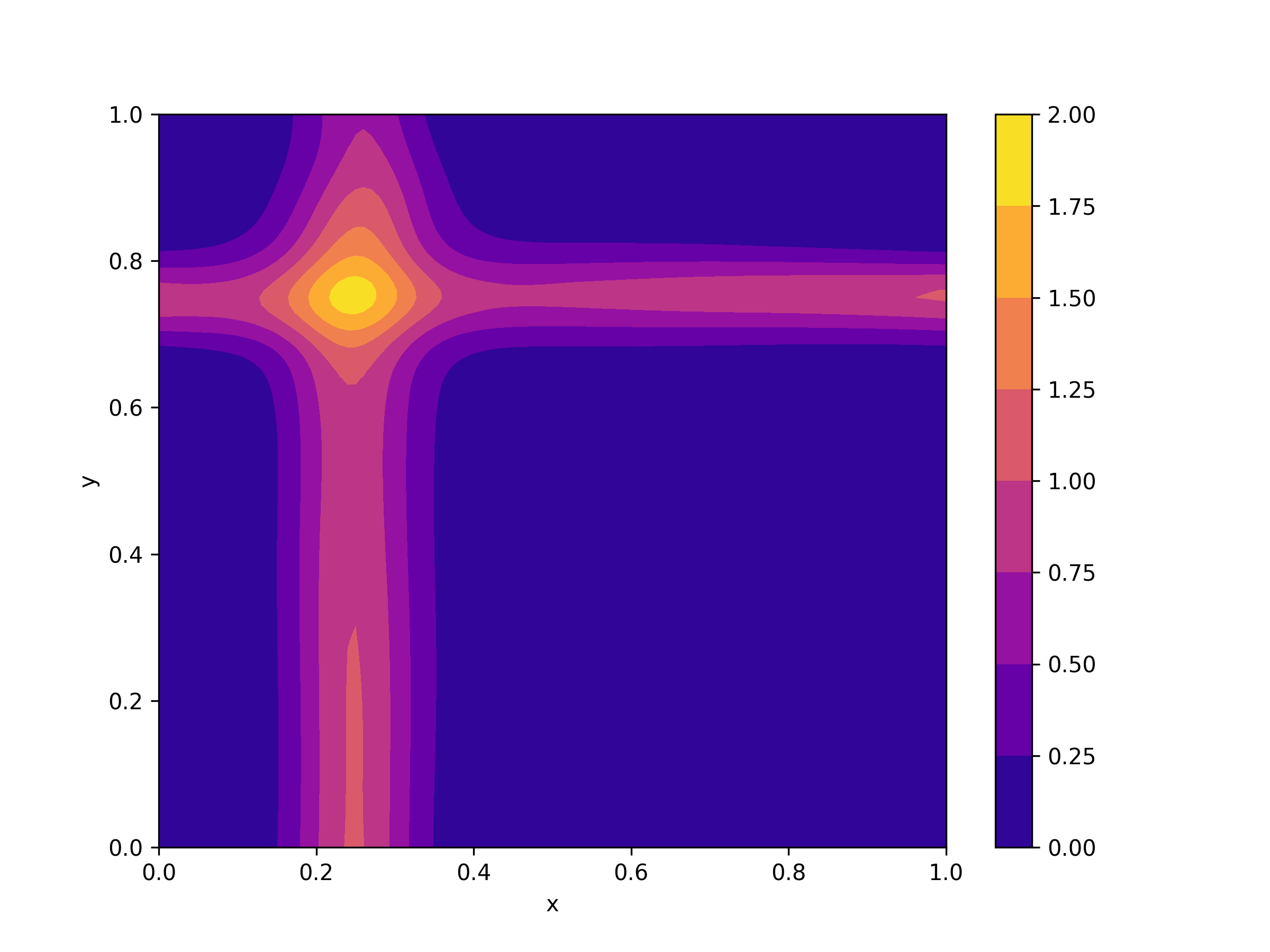}
        \end{overpic} &
    \end{tabular}
    \caption{ \textbf{Predicted source function $f$ vs exact $f$ (\textbf{\textit{Third column}}) using 4 (\textbf{\textit{First column}}) and 15 pointwise observations (\textbf{\textit{Second column}})  with $1\%$, $5\%$ and $10\%$ of noise respectively with pointwise observations.}}
    \label{sourceId}
\end{figure}

\begin{figure}[H] 
    \centering
    \setlength{\tabcolsep}{5pt}
    \renewcommand{\arraystretch}{1.2}
    \begin{tabular}{c c c c c}
     \raisebox{35pt}{\rotatebox[origin=c]{90}{\textbf{$1\%$ noise }}} &
       \begin{overpic}[height=0.09\textheight]
        {figures/saved_plots_4Obs/1percent4Obsu_error1.00.png}
            \put(20,74){\textbf{Error on $u$}} 
            \end{overpic} &
            \begin{overpic}[height=0.09\textheight]
        {figures/saved_plots_4Obs/1percent4Obserror_f.png}
            \put(20,74){\textbf{Error on $f$}} 
            \end{overpic} &
        \begin{overpic}[height=0.09\textheight]
        {figures/saved_plots15Obs/1percent15Obsu_error1.00.png}
            \put(20,74){\textbf{Error on u}} 
            \end{overpic} &
             \begin{overpic}[height=0.09\textheight]
        {figures/saved_plots15Obs/1percent15Obserror_f.png}
            \put(20,74){\textbf{Error on f}} 
            \end{overpic} \\
     \raisebox{35pt}{\rotatebox[origin=c]{90}{\textbf{$5\%$ noise }}} &
       \begin{overpic}[height=0.09\textheight]
        {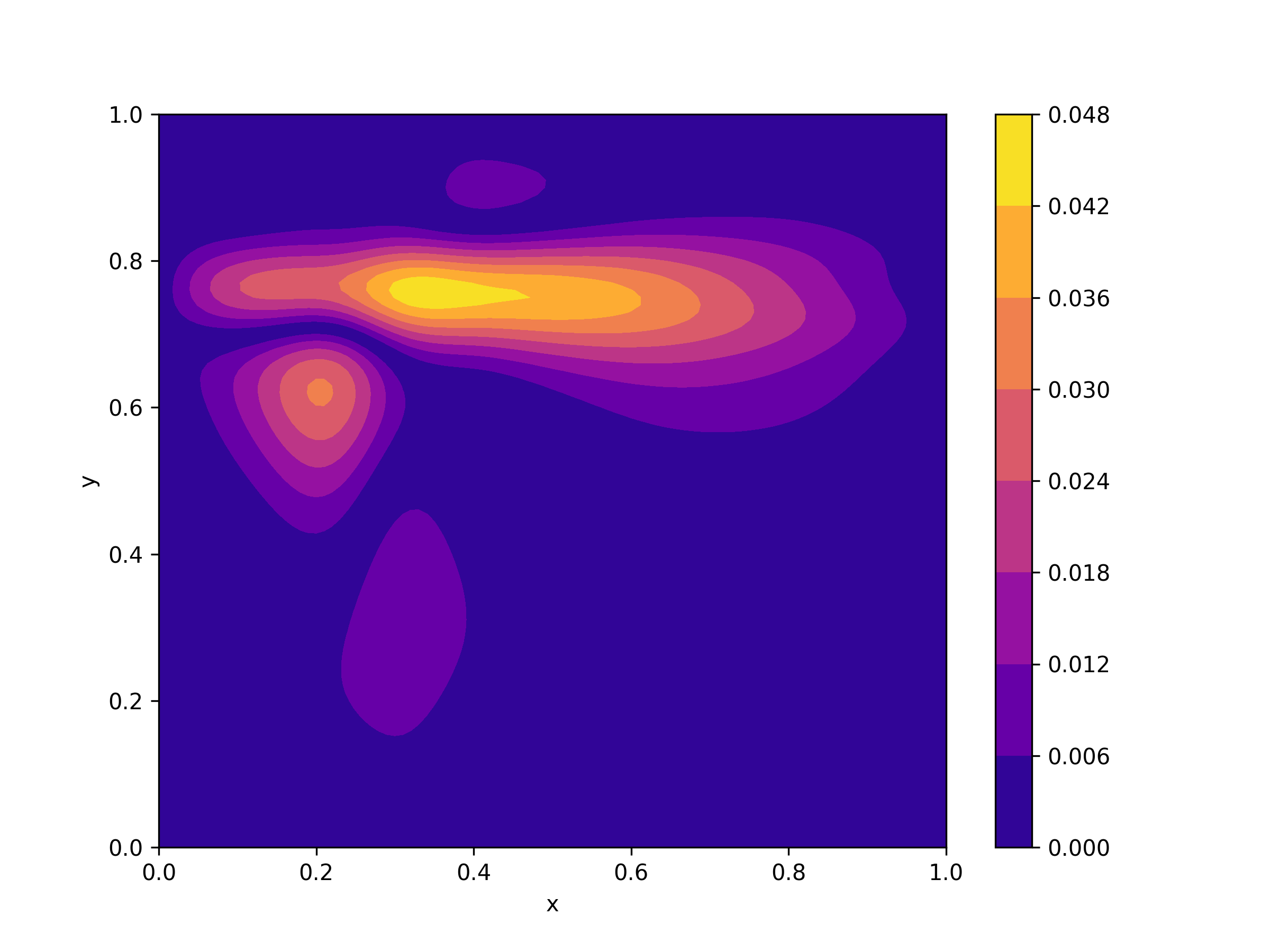}
            \end{overpic} &
            \begin{overpic}[height=0.09\textheight]
        {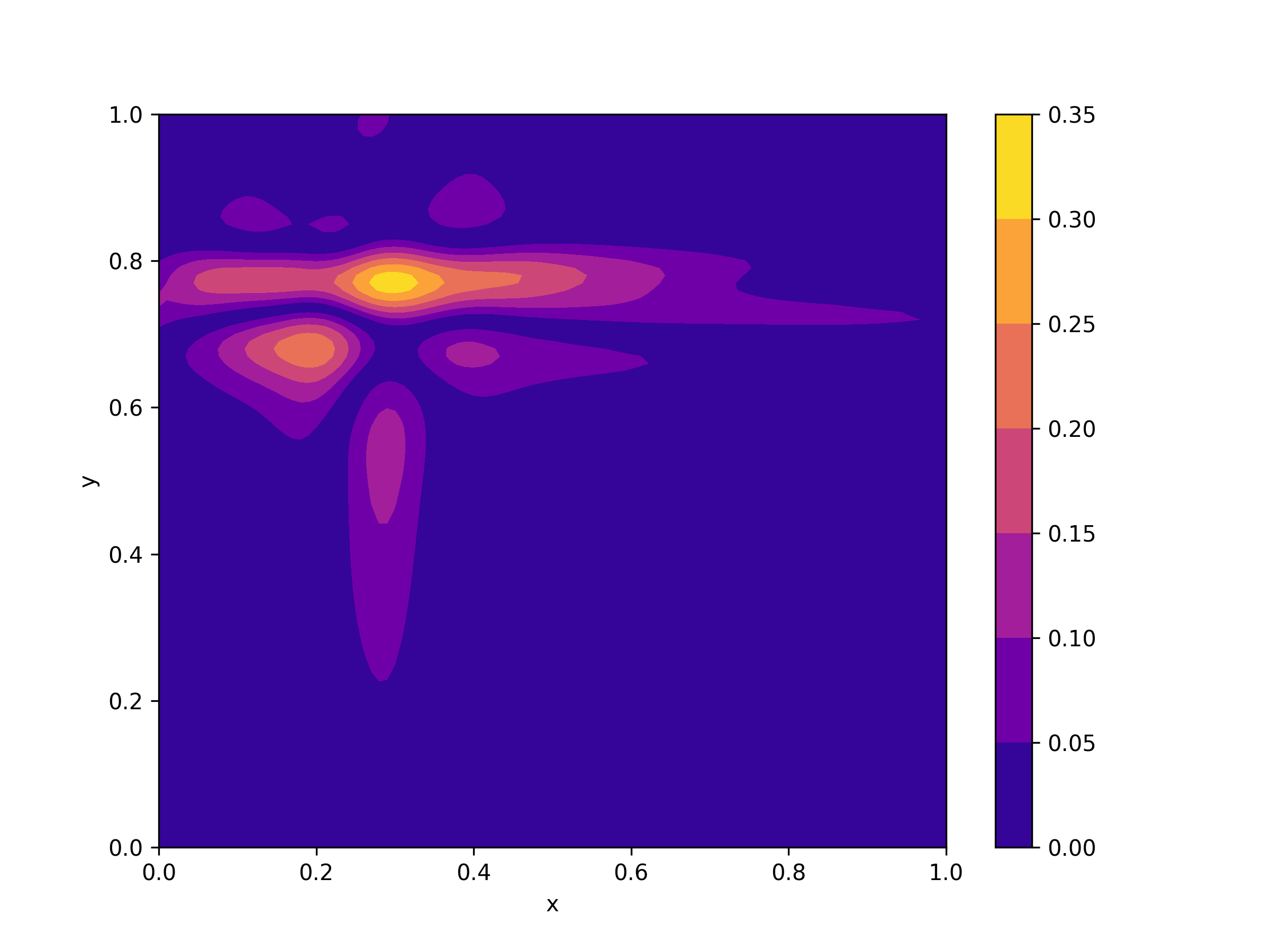}
            \end{overpic} &
             \begin{overpic}[height=0.09\textheight]
        {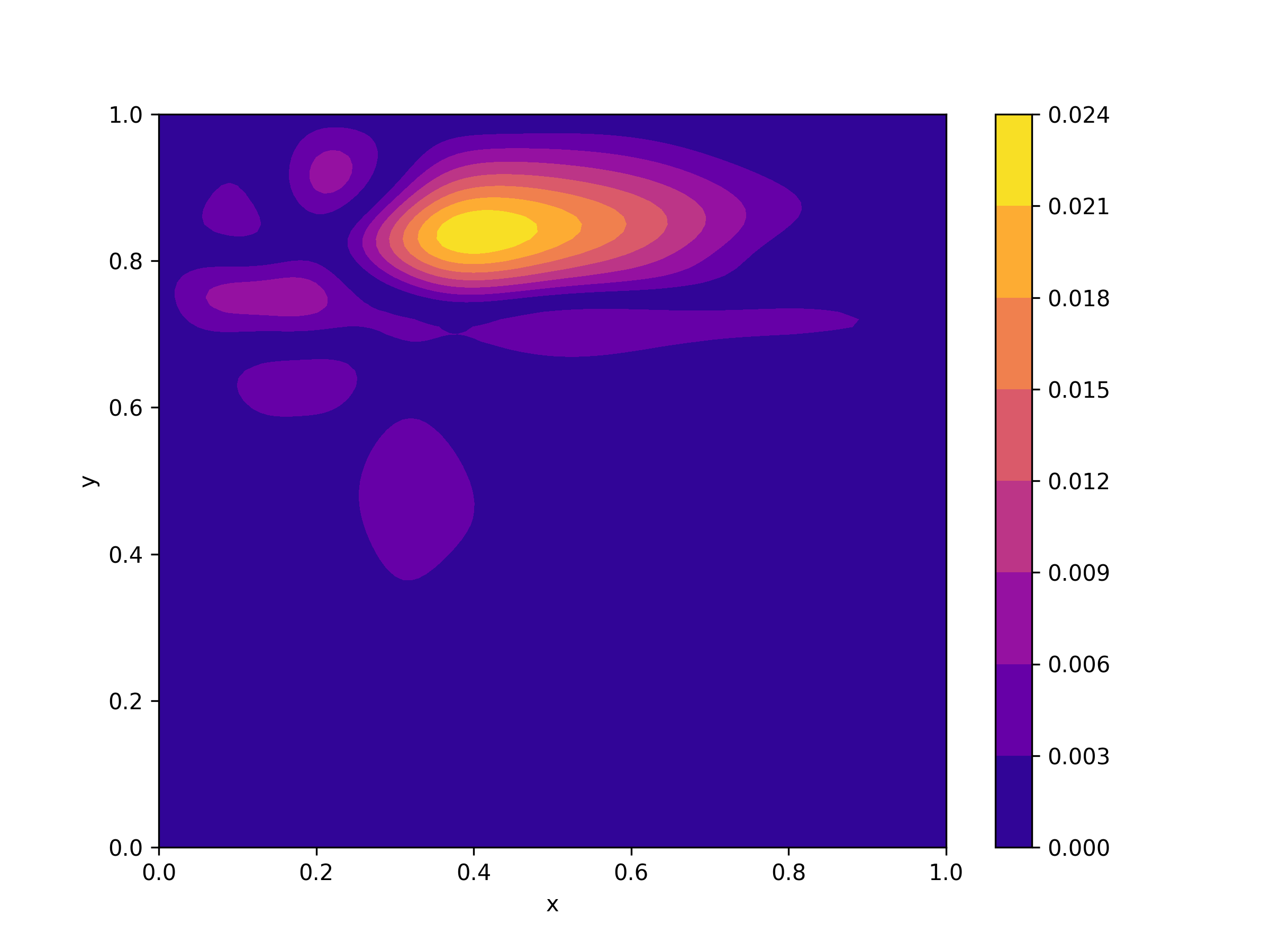}
            \end{overpic} &
             \begin{overpic}[height=0.09\textheight]
        {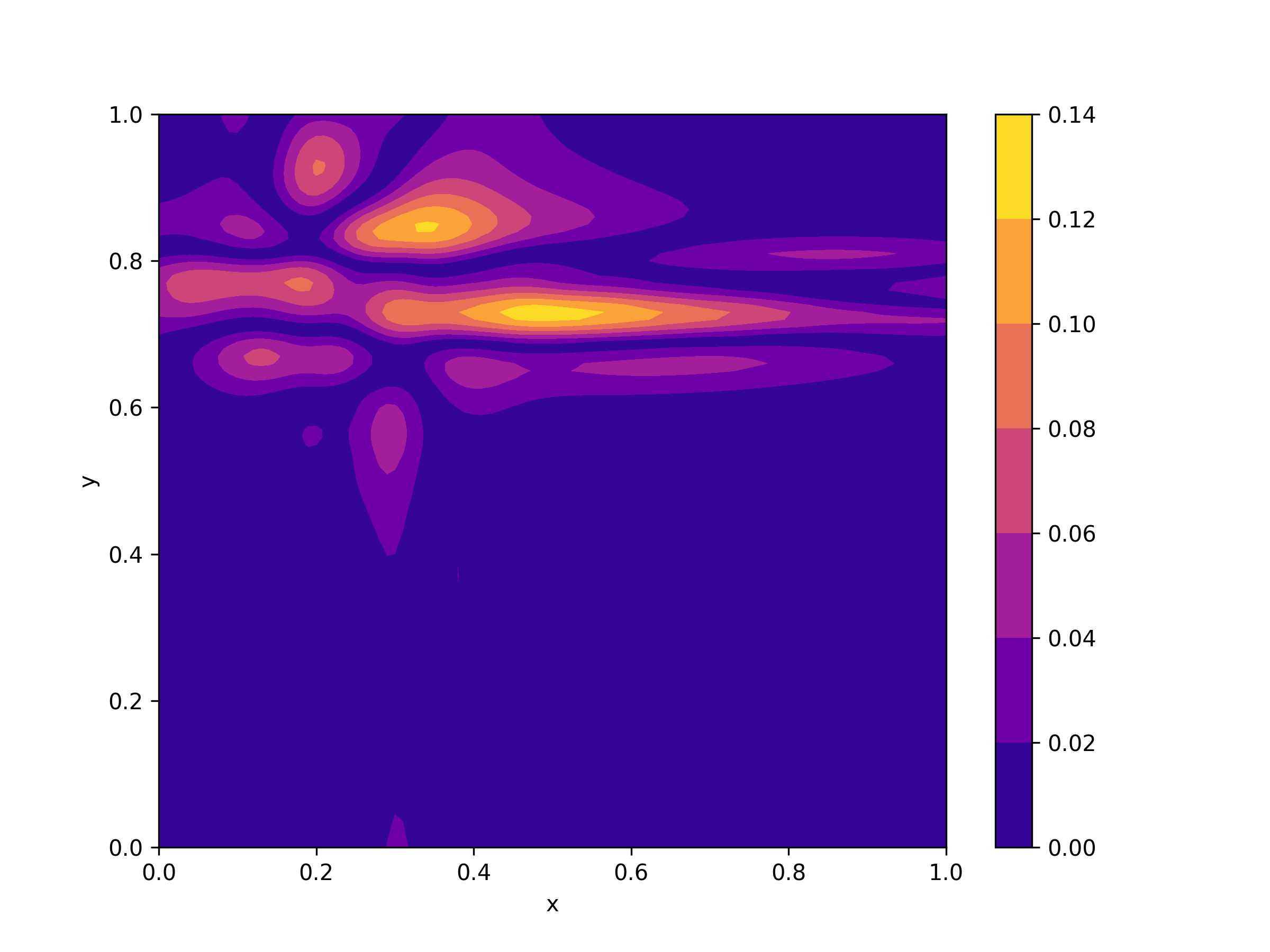}
            \end{overpic} \\
    \raisebox{35pt}{\rotatebox[origin=c]{90}{\textbf{$10\%$ noise }}} &
       \begin{overpic}[height=0.09\textheight]
        {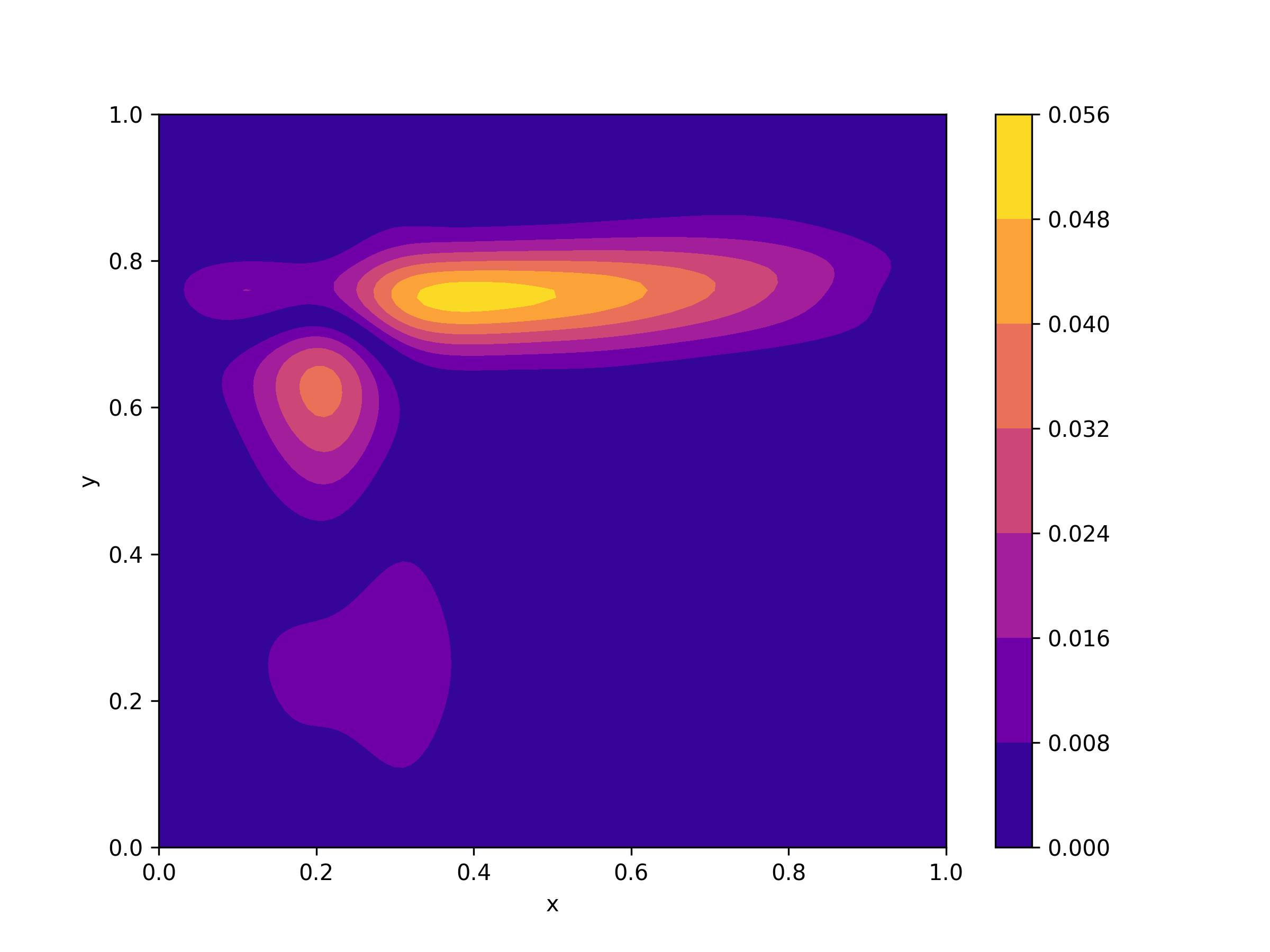}
             \put(50,-5){\makebox(0,0){\textbf{4 observations}}}
            \end{overpic} &
            \begin{overpic}[height=0.09\textheight]
        {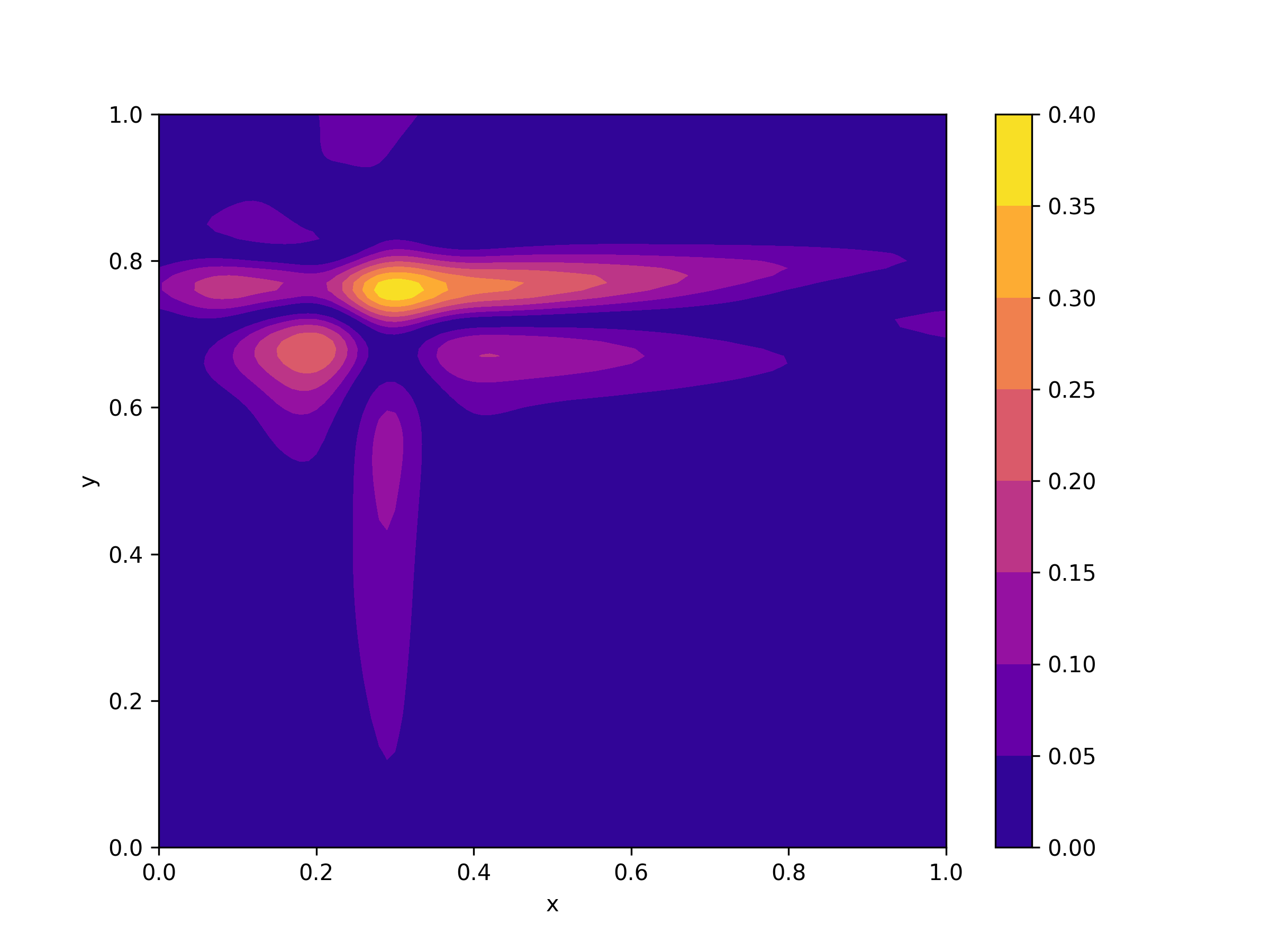}
             \put(50,-5){\makebox(0,0){\textbf{4 observations}}}
            \end{overpic} &
             \begin{overpic}[height=0.09\textheight]
        {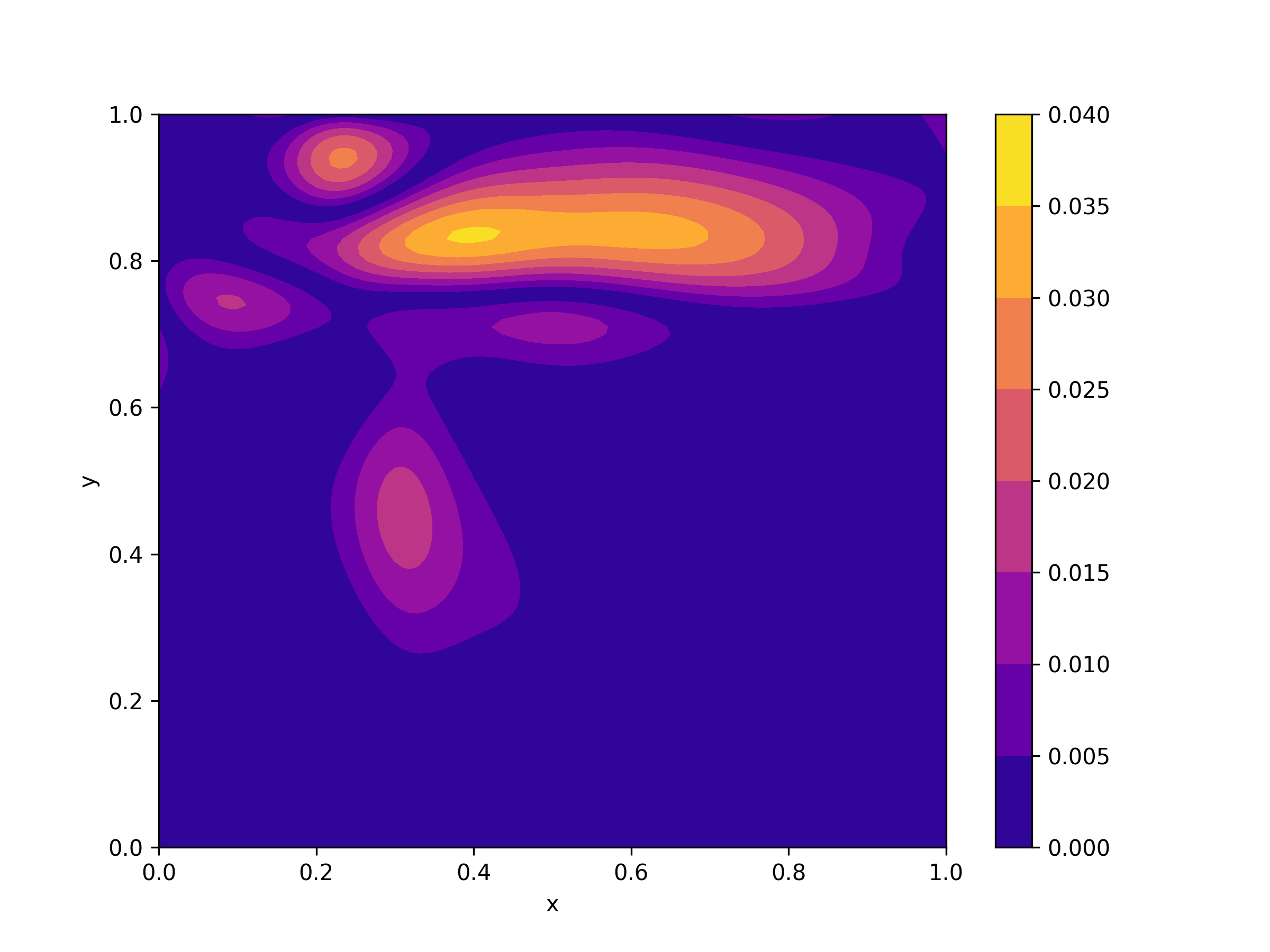}
             \put(50,-5){\makebox(0,0){\textbf{15 observations}}}
            \end{overpic} &
             \begin{overpic}[height=0.09\textheight]
        {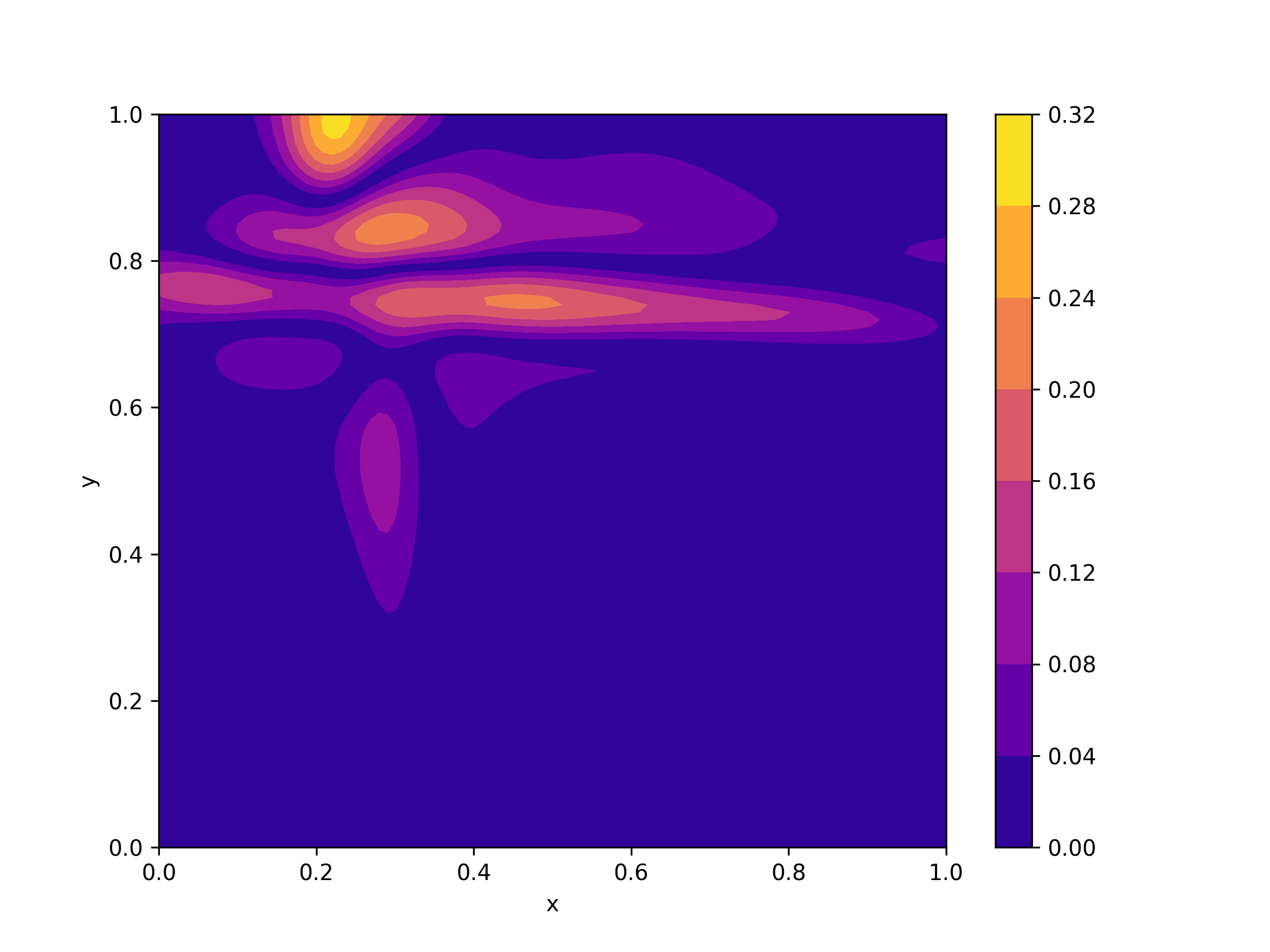}
             \put(50,-5){\makebox(0,0){\textbf{15 observations}}}
            \end{overpic} 
    \end{tabular}
    \caption{\textbf{Contour plots of the absolute error on $u$ at $t=1$ and $f$ with $1\%, 5\%$ and $10\%$ of noise on pointwise observations.}}
     \label{ErrorId}
    \end{figure}

\begin{figure}[H] 
    \centering
    \setlength{\tabcolsep}{5pt}
    \renewcommand{\arraystretch}{1.2}
    \begin{tabular}{c c c c c}
     \raisebox{35pt}{\rotatebox[origin=c]{90}{\textbf{$5\%$ noise }}} &
       \begin{overpic}[height=0.09\textheight]
        {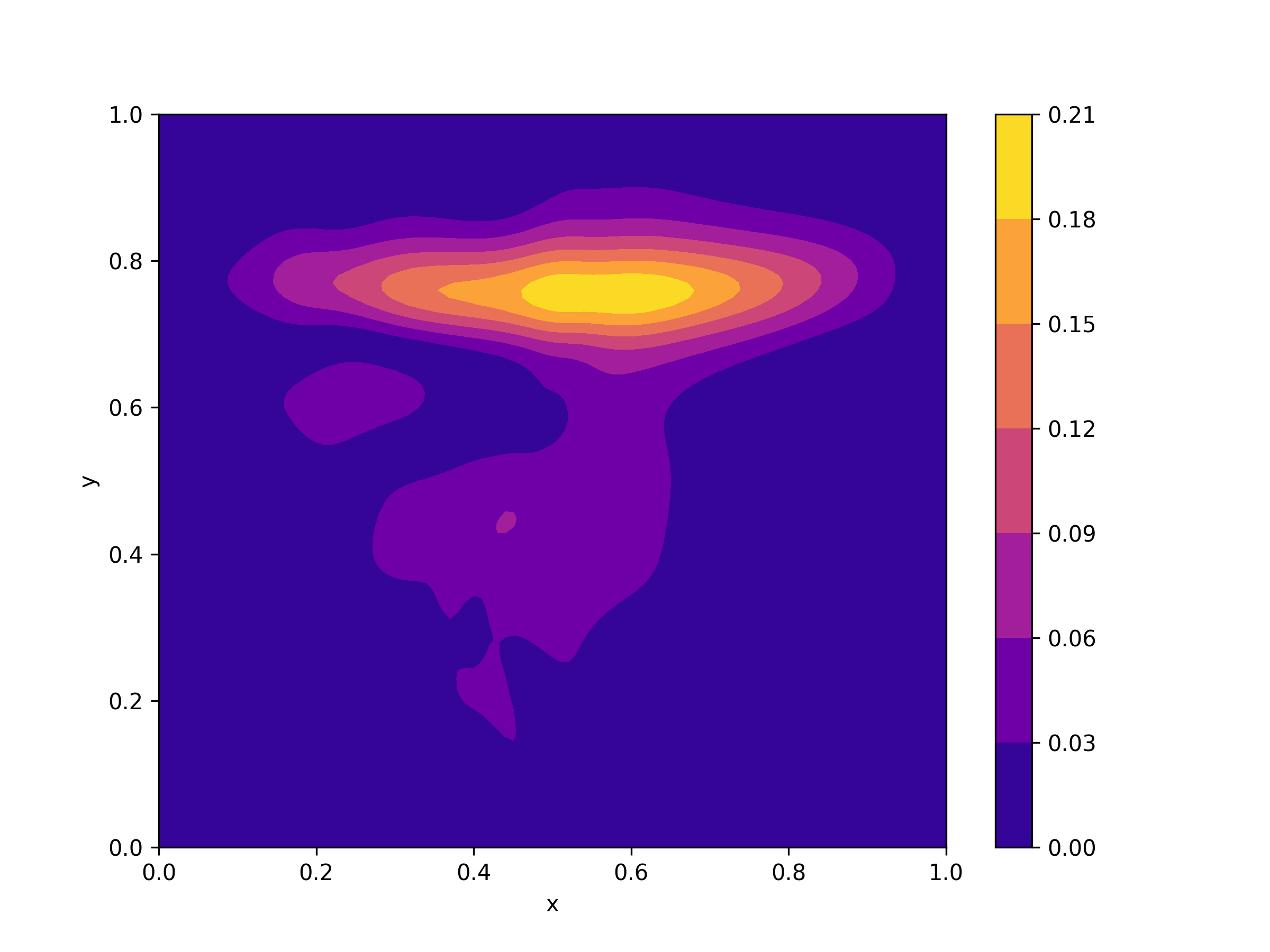}
            \put(20,74){\textbf{Error on $u$ at $t=1.0$}} 
            \end{overpic} &
            \begin{overpic}[height=0.09\textheight]
        {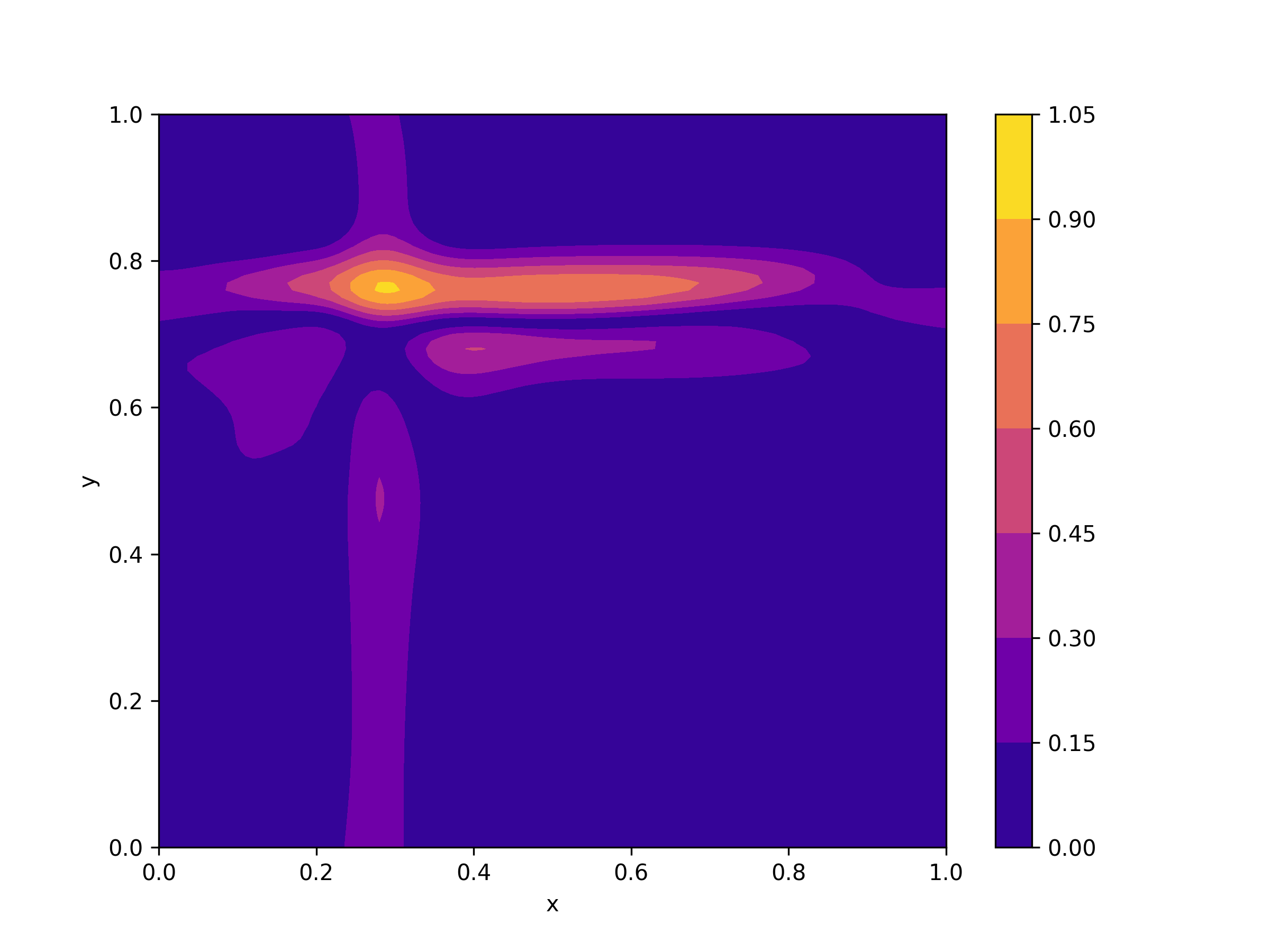}
            \put(20,74){\textbf{Error on $f$}} 
            \end{overpic} &
        \begin{overpic}[height=0.09\textheight]
        {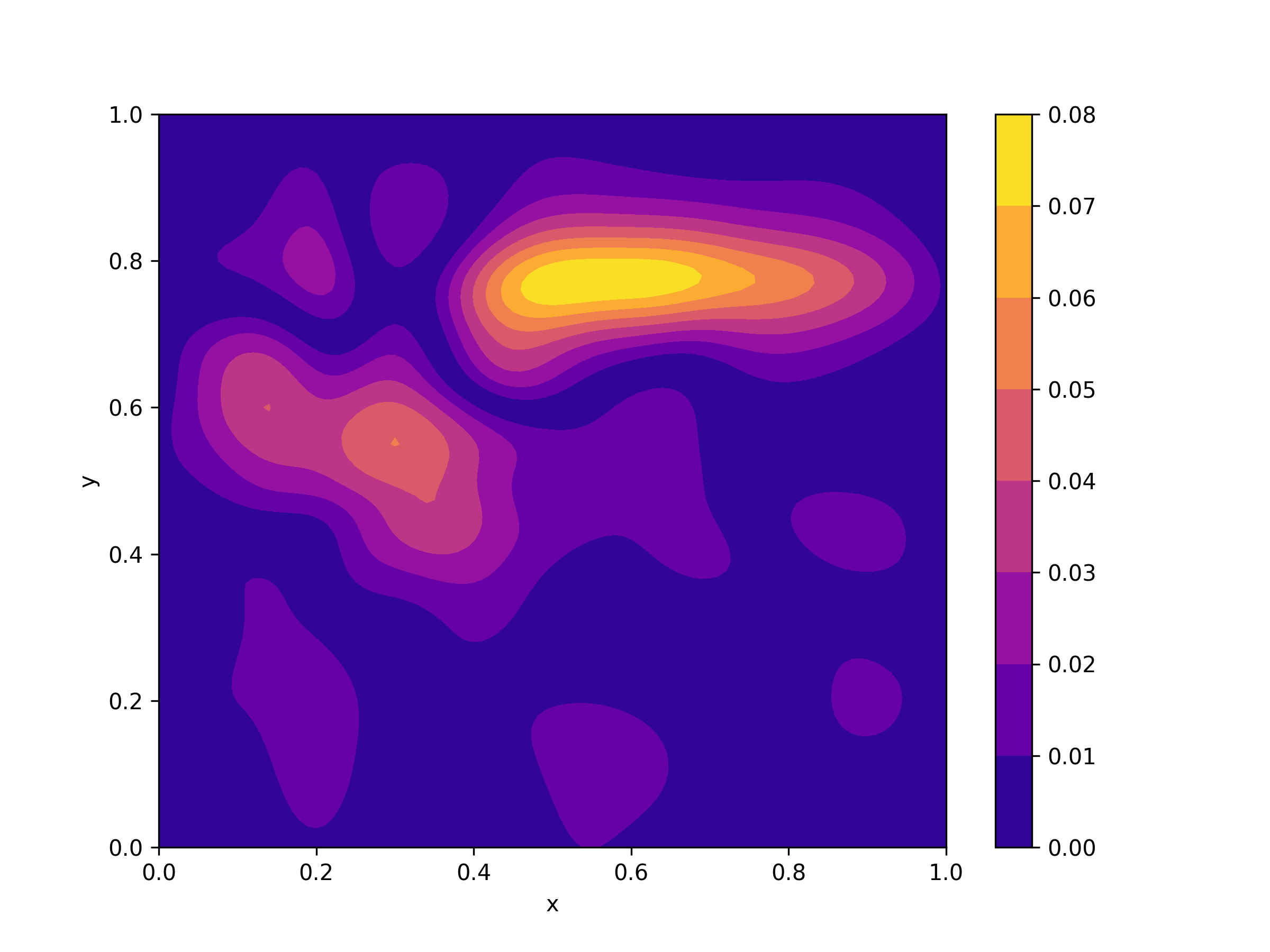}
            \put(20,74){\textbf{Error on u at $t=1.0$}} 
            \end{overpic} &
             \begin{overpic}[height=0.09\textheight]
        {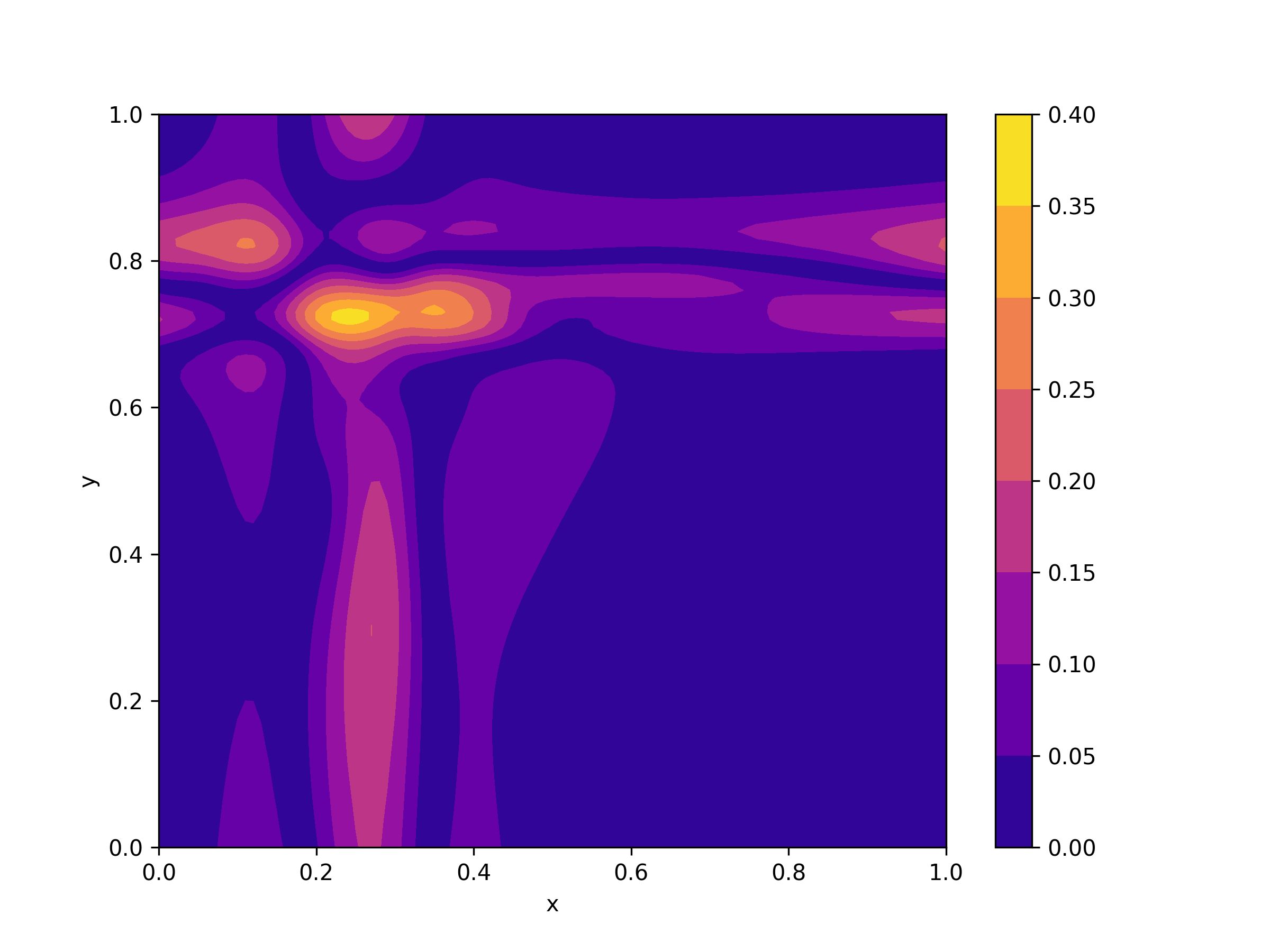}
            \put(20,74){\textbf{Error on f}} 
            \end{overpic} \\
     \raisebox{35pt}{\rotatebox[origin=c]{90}{\textbf{$10\%$ noise }}} &
       \begin{overpic}[height=0.09\textheight]
        {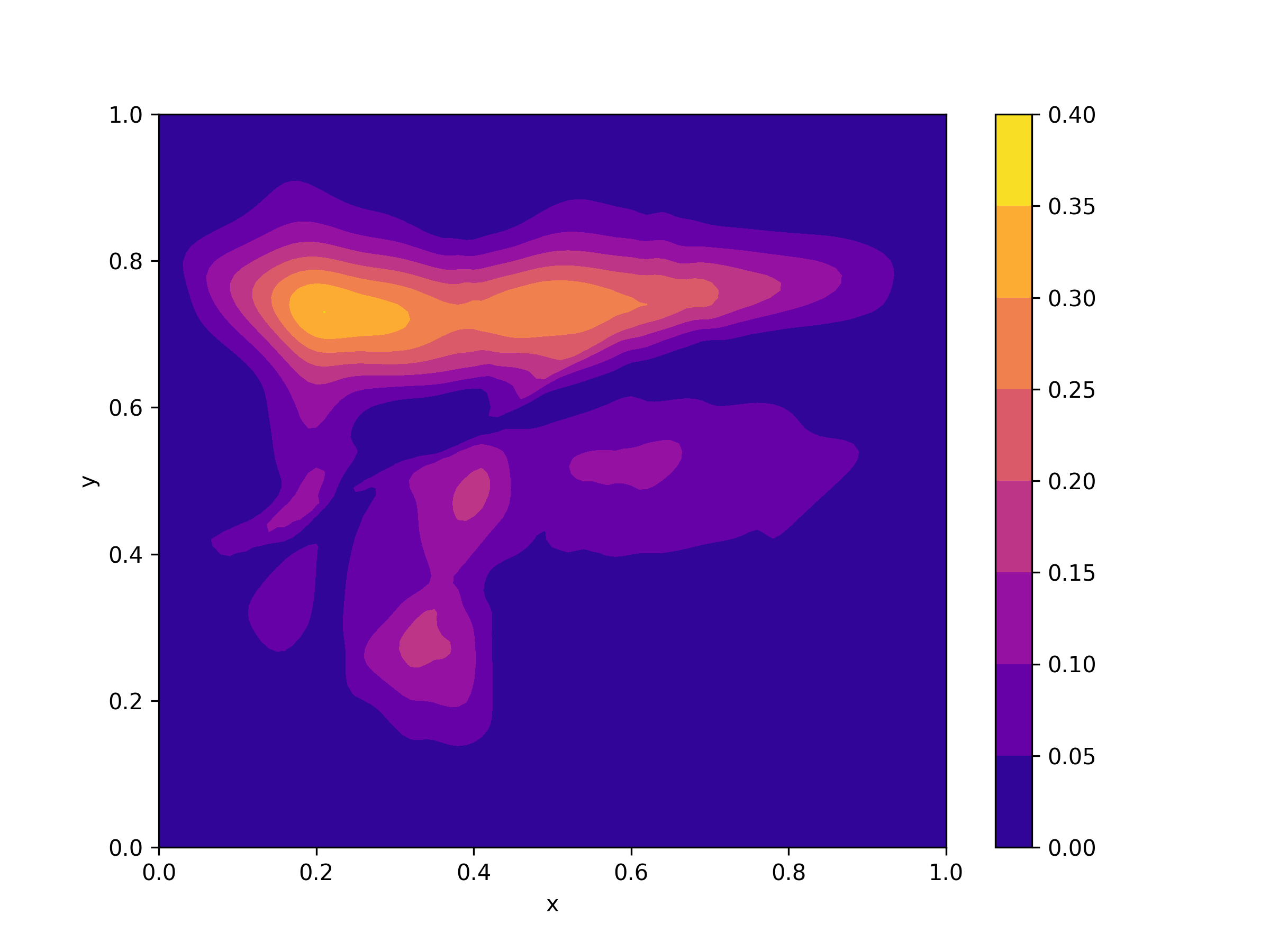}
            \put(50,-5){\makebox(0,0){\textbf{4 observations}}}
            \end{overpic} &
            \begin{overpic}[height=0.09\textheight]
        {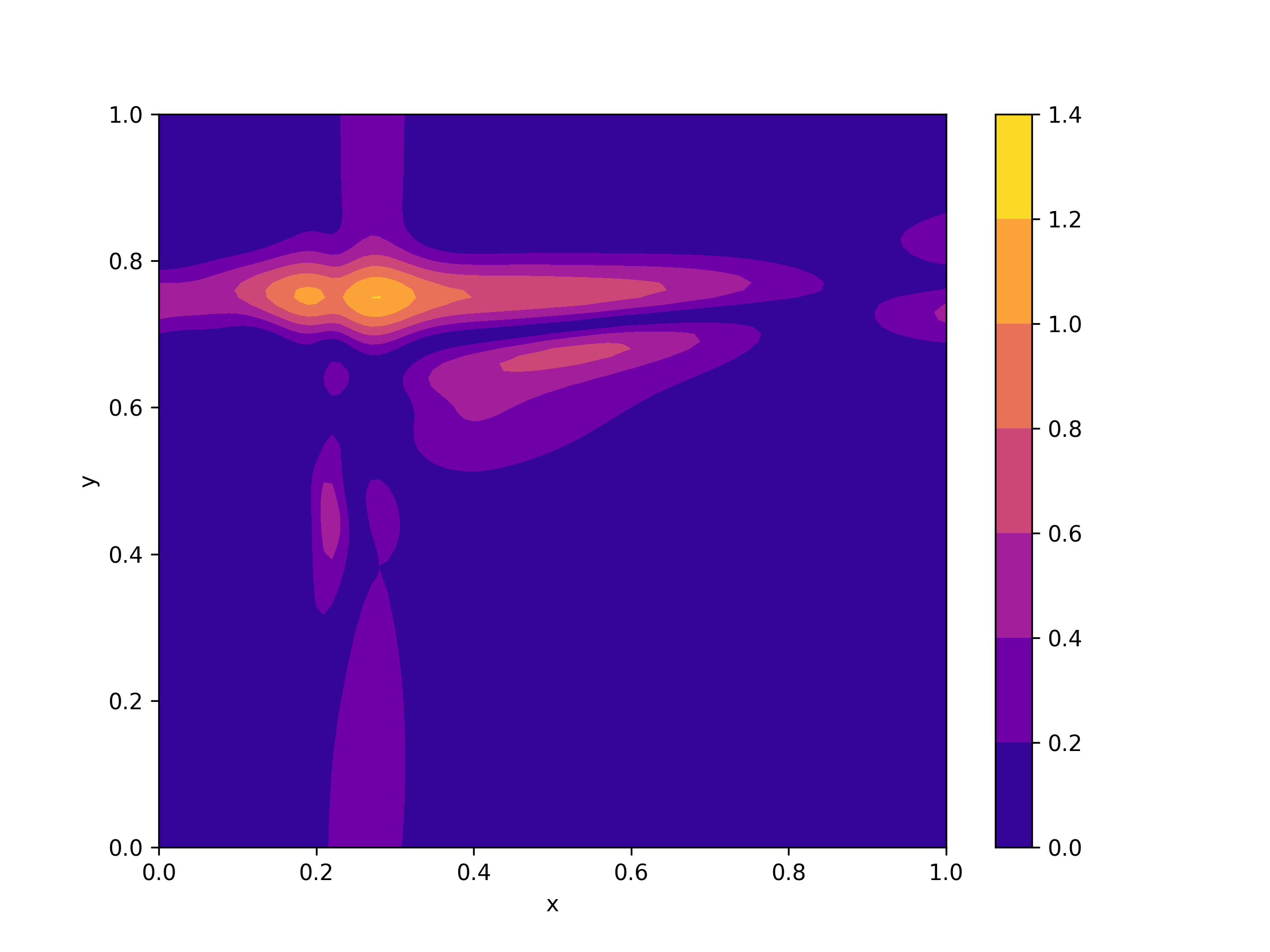}
            \put(50,-5){\makebox(0,0){\textbf{4 observations}}}
            \end{overpic} &
             \begin{overpic}[height=0.09\textheight]
        {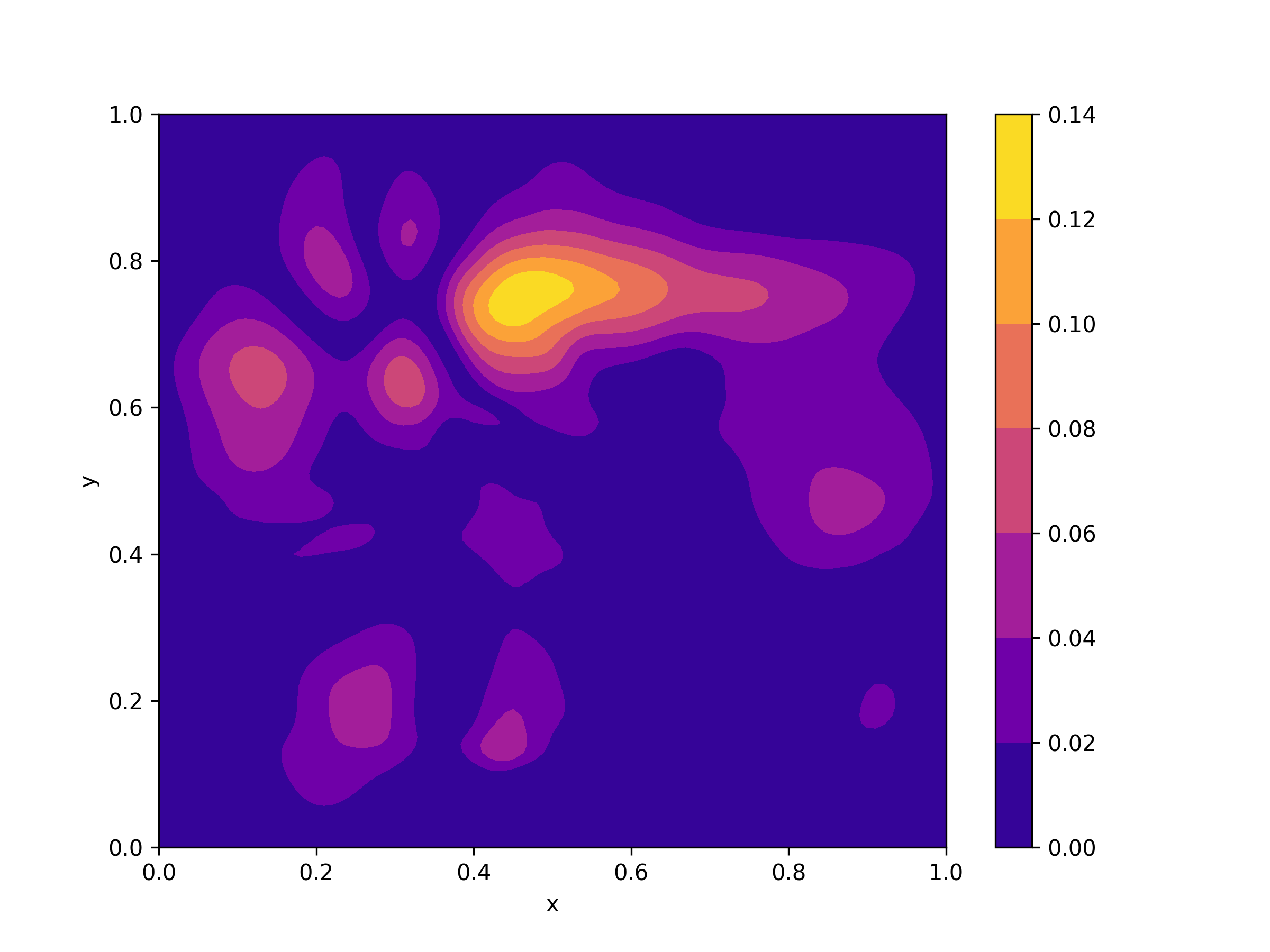}
            \put(50,-5){\makebox(0,0){\textbf{15 observations}}} 
            \end{overpic} &
             \begin{overpic}[height=0.09\textheight]
        {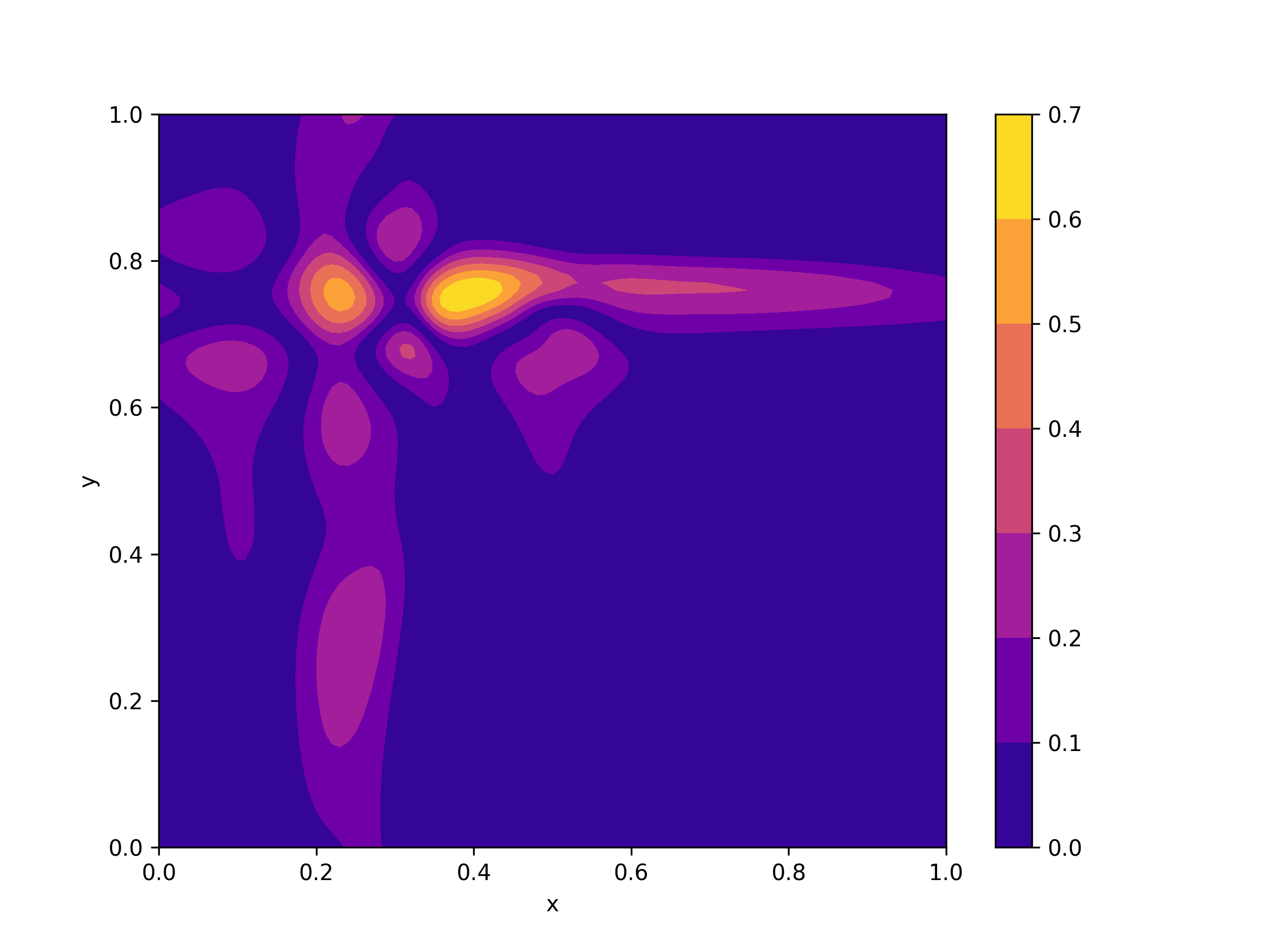}
            \put(50,-5){\makebox(0,0){\textbf{15 observations}}} 
            \end{overpic}
    \end{tabular}
    \caption{\textbf{Contour plots of the absolute error on $u$ at $t=1$ and $f$ with  $5\%$ and $10\%$ of noise on 4 and 15 accumulative observations.}}
    \label{ErrorAVG}
    \end{figure}

Despite the variation of noise, the high-ill-posedness of our problem and the measurement locations chosen not too close to emission sources, as shown on the figure \ref{fig:ExactsolSource}, we can deduce from the results on the figures \ref{sourceId}, \ref{ErrorAVG}, \ref{ErrorId} that in such scenario the proposed PINNs approach can still exhibit good performances in identifying the location of emission sources and recovering the solution of the ADE. 
\begin{figure}[H] 
    \centering
    \setlength{\tabcolsep}{5pt}
    \renewcommand{\arraystretch}{1.2}
    
    \begin{tabular}{c c c c}
      \raisebox{40pt}{\rotatebox[origin=c]{90}{\textbf{$1\%$ noise}}} & 
       \begin{overpic}[width=0.28\textwidth]
        {figures/saved_plots_4Obs/1percent4Obsu_predicted_t1.00.png}
            \put(1,73){\textbf{Predicted u with 4 obs}}
            \end{overpic} &
       \begin{overpic}[width=0.28\textwidth]
            {figures/saved_plots15Obs/1percent15Obsu_predicted_t1.00.png}
            \put(1,73){\textbf{Predicted u with 15 obs}} 
        \end{overpic} \\
        \raisebox{40pt}{\rotatebox[origin=c]{90}{\textbf{$5\%$ noise}}} &
        \begin{overpic}[width=0.28\textwidth]
            {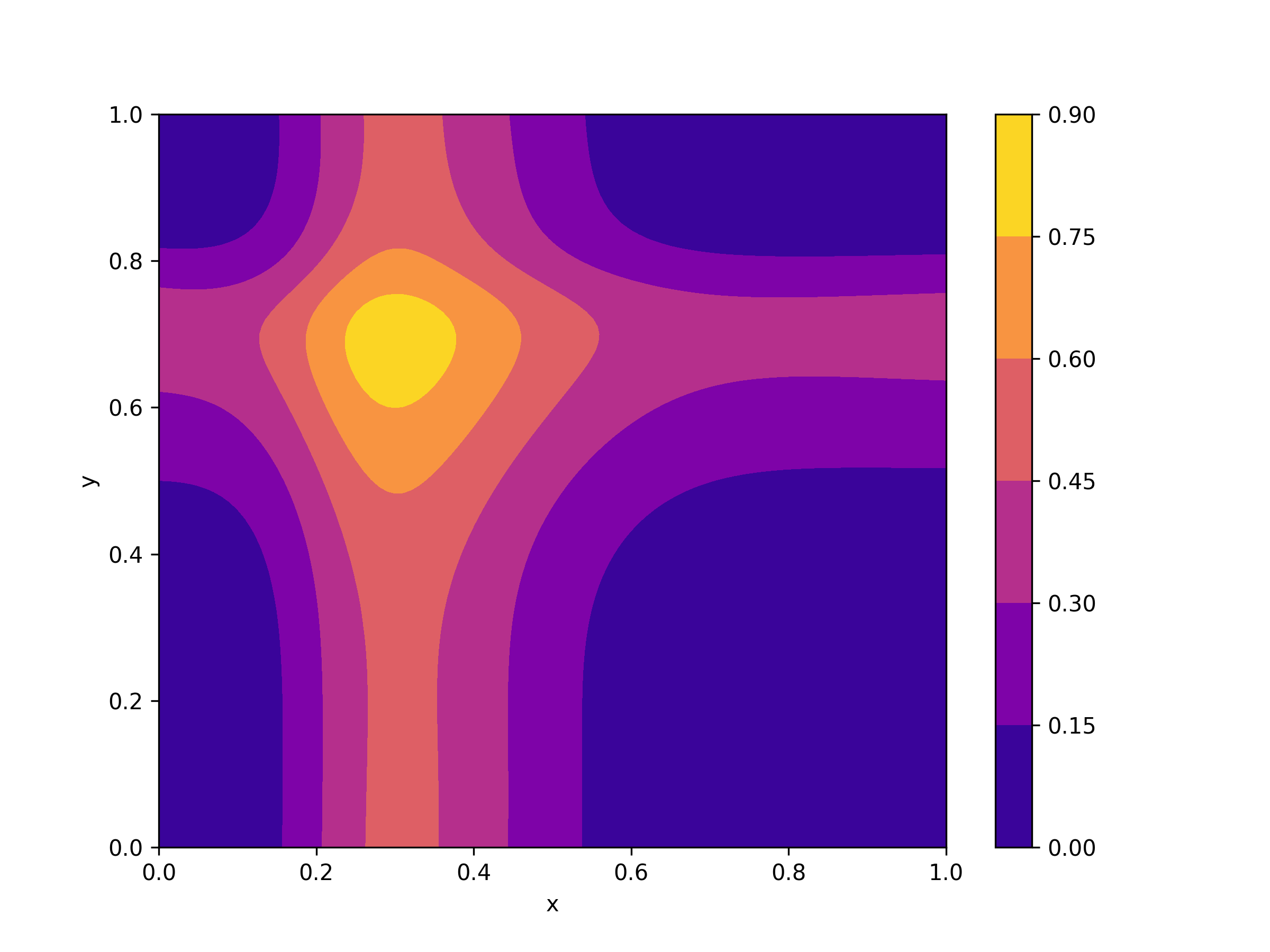}
        \end{overpic} &
        \begin{overpic}[width=0.28\textwidth]
            {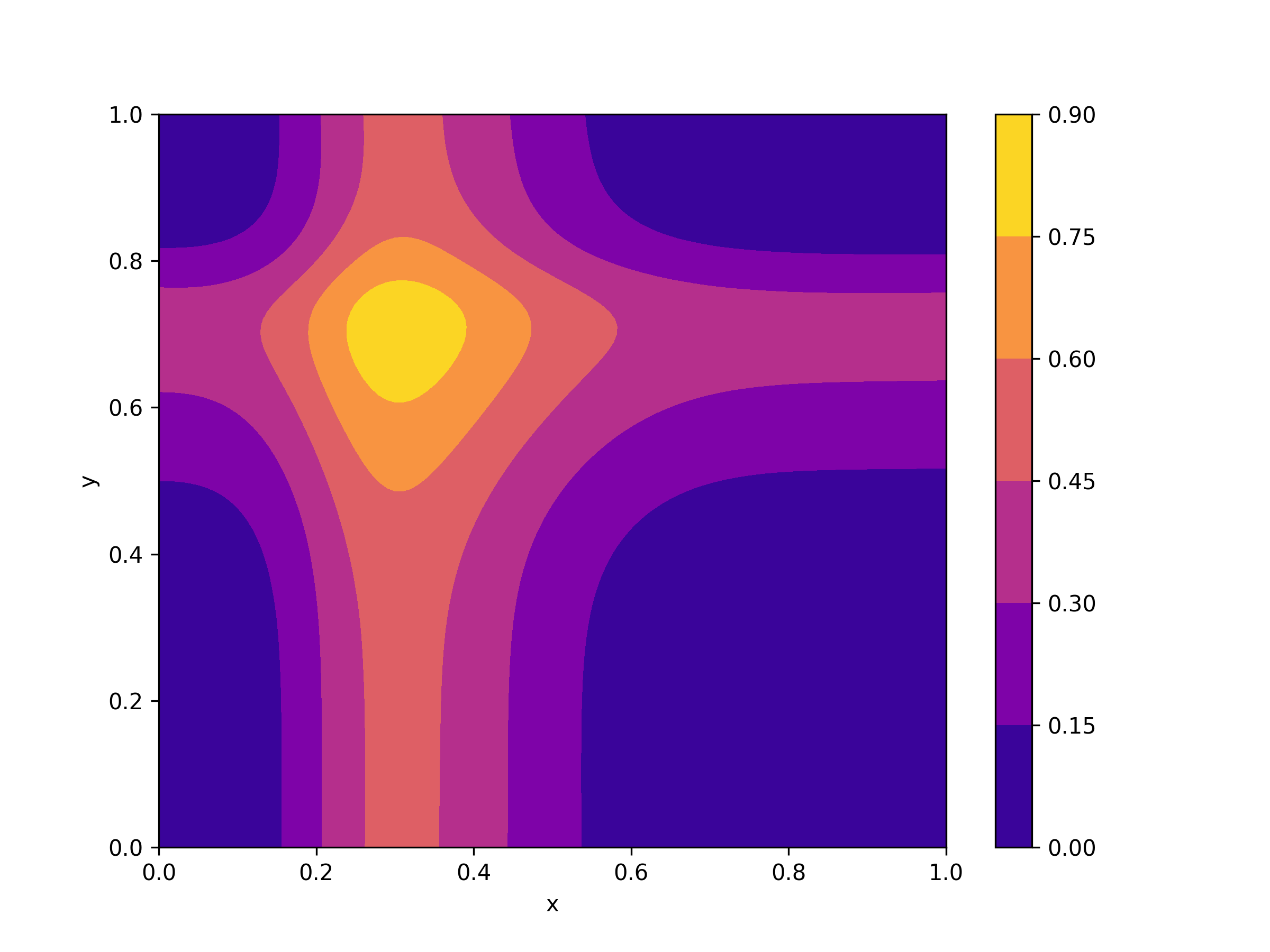}
        \end{overpic} &
        \begin{overpic}[width=0.28\textwidth]
            {figures/saved_plots_4Obs/1percent4Obsu_exact_t1.00.png}
             \put(20,73){\textbf{FEM solution u}}
        \end{overpic} \\  
       \raisebox{40pt}{\rotatebox[origin=c]{90}{\textbf{$10\%$ noise}}} &
        \begin{overpic}[width=0.28\textwidth]
            {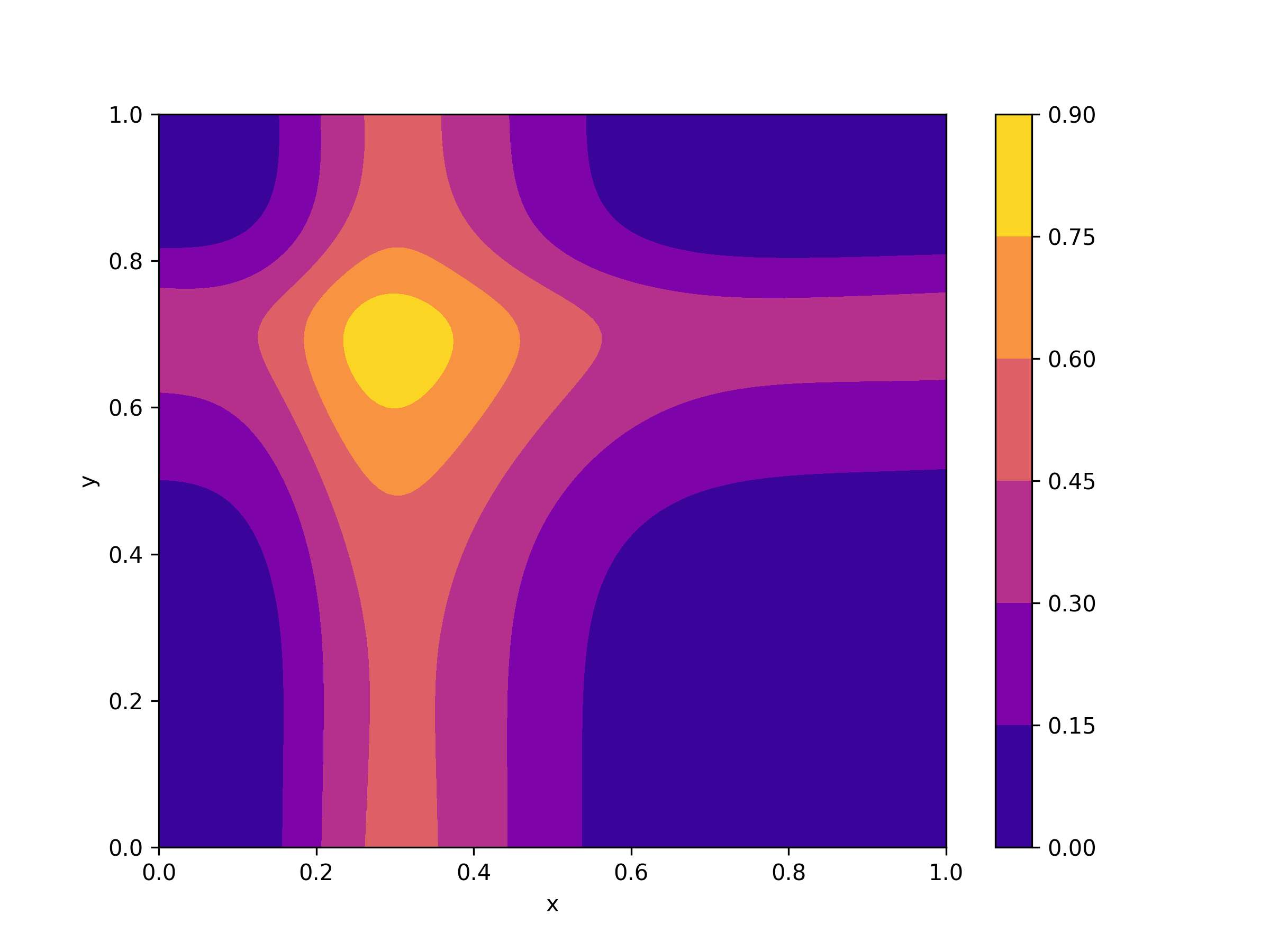}
        \end{overpic} &
        \begin{overpic}[width=0.28\textwidth]
            {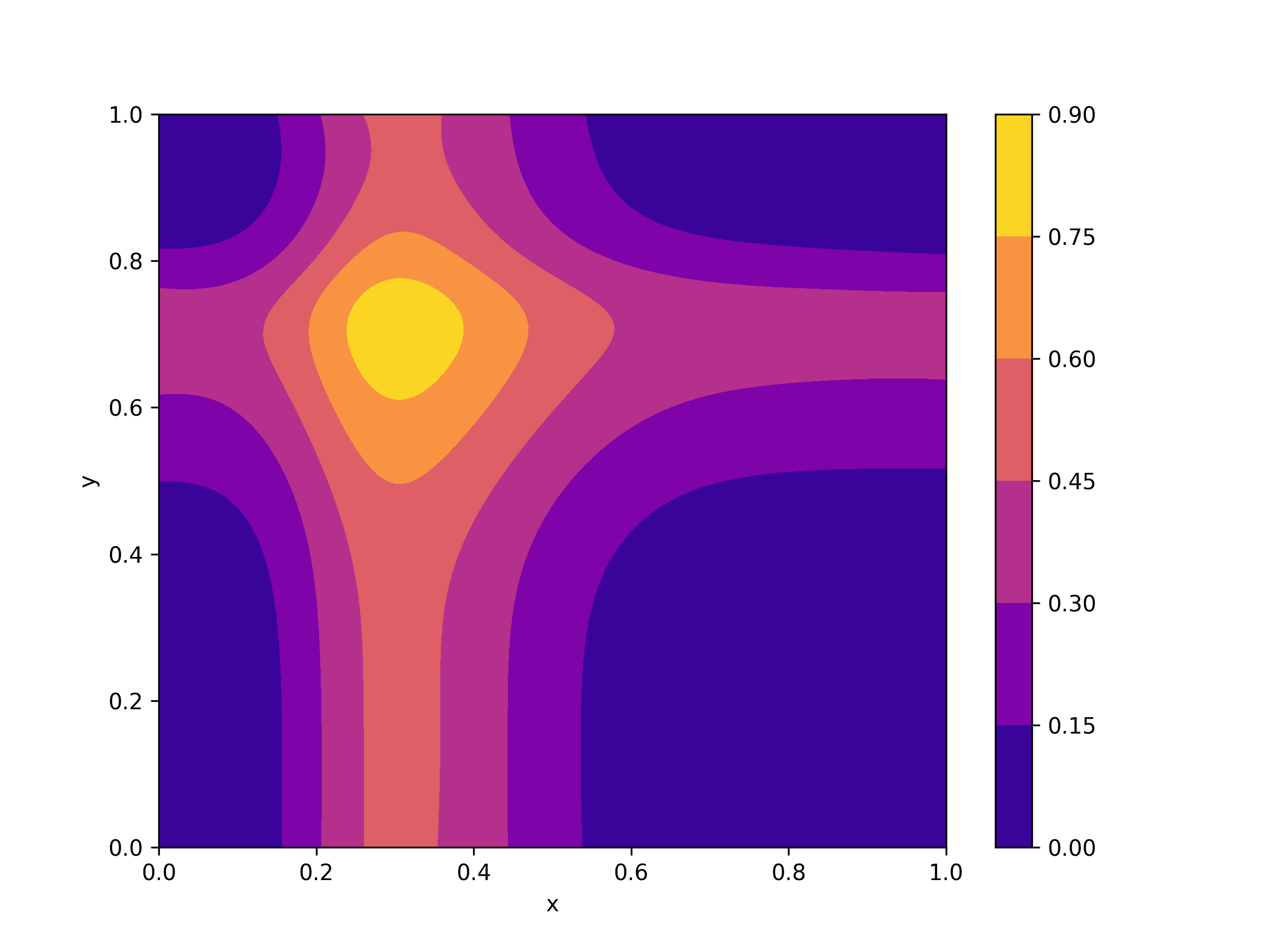}
        \end{overpic} &
    \end{tabular}
    \caption{\textbf{ Predicted solution $u$ at $t=1$ vs FEM solution (\textbf{\textit{Third column}}) using with 4 (\textit{\textbf{First column}}) and 15 pointwise observations (\textbf{\textit{Second column}}) with $1\%$, $5\%$ and $10\%$ of noise respectively.}}
    \label{solID}
\end{figure}

\section{Proof of Theorem \ref{cond_stab}} \label{proof::theo1AppendixC}

The proof of theorem \ref{cond_stab} relies on the following Carleman estimate:
\begin{lemma}[{\cite[Thm.~4.1]{yamamoto2009carleman}}]
	Let $\omega_0$ be an arbitrarily fixed sub-domain of $\Omega$ such that 
	$\overline{\omega_0} \subset \omega$. Then there exists a function 
	$\psi\in C^2(\overline{\Omega})$ such that
	\begin{equation}
		\psi(\mathbf{x}) > 0, \quad \mathbf{x} \in \Omega, \qquad 
		\psi\big|_{\partial\Omega} = 0, \qquad 
		|\nabla \psi(\mathbf{x})| > 0, \quad \mathbf{x} \in \overline{\Omega} \setminus \omega_0.
		\label{psieq}
	\end{equation}
\end{lemma}
Let us now consider $\nu>0$, and define
\begin{equation}\
	\varphi(\mathbf{x},t) := \frac{e^{\nu \psi(\mathbf{x})}}{(t-(t_0-\eta))(t_0+\eta-t)}, 
	\qquad 
	\alpha(\mathbf{x},t) := \frac{e^{\nu\psi(\mathbf{x})} - e^{2\nu\|\psi\|_{C(\overline{\Omega})}}}{(t-(t_0-\eta))(t_0+\eta-t)},
	\label{phialphaeq}
\end{equation}
We then also have the following lemma: 
\begin{lemma}[{\cite[Lemma.~2.1]{hu2025conditional}}]
	Let $\psi, \varphi$, and $\alpha$ be as above. Then we have the following estimate:
	\begin{equation}
		\alpha>0, \quad e^{2p\alpha(\mathbf{x},t)}\leq e^{2p\alpha(\mathbf{x},t_0)}\leq 1, \quad \partial_{t}\alpha\leq c\eta\varphi^{2},  \phi^{-1}\leq \eta^2 \quad \text{in } \bar{\Omega}\times ]t_0-\eta,t_0+\eta[
		\label{ineqalpha}
	\end{equation}
\end{lemma}
Now recall our shorthand notation $Au=V\cdot\nabla u -\nabla(D\nabla u)$. 
Then the following  Carleman estimate for a parabolic PDEs holds.
\begin{theorem}[{\cite[Thm.~4.1]{yamamoto2009carleman}}]\label{thm:4.1}
	There exists a number $\nu_0 > 0$ such that for an arbitrary $\nu \geq \nu_0$, we can chose a constant $p_0(\nu) \geq 0$ 
	satisfying: there exists a constant $C = C(p_0, \nu_0) > 0$ such that
	\begin{equation}\label{eq:carleman}
		\begin{aligned}
			\int_Q \left\{
			\frac{1}{p\varphi} 
			\left( |\partial_t u|^2 +  |A u|^2 \right)
			+ p\nu^2 \varphi |\nabla u|^2 
			+ p^3\nu^4 \varphi^3 |u|^2 
			\right\} e^{2p\alpha} \, dx\, dt \\
			\leq c \int_Q |Lu|^2 e^{2p\alpha} \, dt\, dx 
			+ c \int_{ \omega \times (0,T)} p^3\nu^4 \varphi^3 u^2 
			e^{2s\alpha} \, dt\, dx
		\end{aligned}
	\end{equation}
	for all $p > p_0$ and all $u \in  H^{1}(0,T;L^{2}(\Omega))\cap L^{2}(0,T;H^{2}(\Omega))$ satisfying \eqref{ade}, with $L$ is the differential operator associated to the corresponding PDE satisfied by $u$. 
\end{theorem}

We can now present the 
proof of Theorem~\ref{cond_stab}:
By \eqref{phialphaeq}, we note that
\begin{equation}
	\lim_{t \downarrow 0} \varphi(\mathbf{x},t)^k e^{2p\alpha(\mathbf{x},t)} 
	= \lim_{t \uparrow T} \varphi(\mathbf{x},t)^k e^{2p\alpha(\mathbf{x},t)} = 0, \quad \text{for any } k > 0. 
\end{equation}

Let $v = \partial_t u$. We can assume $t_0=T/2$ and $\eta=T/2$ , and therefore $v$ satisfies
\begin{eqnarray*}
	\left\{
	\begin{array}{cc}
		\partial_t v + Av = 0 &\quad \text{in } Q \\
		v(\mathbf{x},\frac{T}{2}) = \partial_t u(\mathbf{x},\frac{T}{2}) = -Au(\mathbf{x},\frac{T}{2}) + f(\mathbf{x}), &\quad \mathbf{x} \in \Omega  \\
		\nabla_{n}v(\mathbf{x},t)=0  &\quad \text{in } \partial\Omega\times(0,T)
	\end{array}
	\right.
\end{eqnarray*}

\medskip

\noindent Using triangular inequality, we have: 
$$
\int_\Omega |f|^2 e^{2p\alpha(\mathbf{x},\frac{T}{2})}\,d\mathbf{x} 
\leq c\bigg[\int_\Omega |v(\mathbf{x},\frac{T}{2})|^2 e^{2p\alpha(\mathbf{x},\frac{T}{2})}\,d\mathbf{x} + \int_\Omega |Au(\mathbf{x},\frac{T}{2})|^2 e^{2p\alpha(\mathbf{x},\frac{T}{2})}\,d\mathbf{x}\bigg]
$$
we set $J=\int_\Omega |v(\mathbf{x},\frac{T}{2})|^2 e^{2p\alpha(\mathbf{x},\frac{T}{2})}\,d\mathbf{x}$
\begin{align*}
	J &= \int_\Omega \int_0^{\frac{T}{2}} \partial_t\left(|v|^2 e^{2p\alpha(\mathbf{x},t)}\right)dt\,d\mathbf{x} \quad \text{(since } \lim_{t\rightarrow0}e^{2p\alpha(\mathbf{x},0)}=0)\\
	&= 2\int_\Omega \int_0^T \left(v\,\partial_t v 
	+ p(\partial_t \alpha)|v|^2\right) e^{2p\alpha(\mathbf{x},t)}\,dt\,d\mathbf{x}\\
	&=   2\int_\Omega \int_0^T \left[\left(\frac{1}{p\varphi}\,\partial_t v\right)(p\varphi v) 
	+ p\varphi^{2}|v|^2\right] e^{2p\alpha(\mathbf{x},t)}\,dt\,d\mathbf{x} \quad \text{(since } \partial_{t}\alpha\leq c\varphi^{2} \text{ from \eqref{ineqalpha}})
\end{align*}

\noindent By Young's inequality $ab \leq \frac{\varepsilon^{-1}}{2}a^2 + \frac{\varepsilon}{2}b^2$ (for every $a,b,\epsilon \in \mathbb{R}_{+}$), we have
$$
v\partial_t v\leq \frac{1}{2p^{2}\varphi^{2}}|\partial_t v|^2 
+ \frac{p^{2}\varphi^{2}}{2}|v|^2
$$

\noindent Therefore we have:
\begin{align*}
	J &\leq c \int_\Omega \int_0^{\frac{T}{2}} \left[
	\frac{1}{p^{2}\varphi^{2}}|\partial_t v|^2 
	+ p^{2}\varphi^{2}|v|^2 
	+ 2p\varphi^2|v|^2
	\right] e^{2p\alpha(\mathbf{x},t)}\,dt\,dx\\
	&\leq c \int_\Omega \int_0^{T} \left[
	\frac{1}{p^{2}\varphi^{2}}|\partial_t v|^2 
	+ p^{2}\varphi^{2}|v|^2 
	+ 2p\varphi^2|v|^2
	\right] e^{2p\alpha(\mathbf{x},t)}\,dt\,dx\\
	&\leq c \int_\Omega \int_0^{T} \left[
	\frac{1}{p^{2}\varphi^{2}}|\partial_t v|^2 
	+ p^{2}\varphi^{2}|v|^2 
	+ p\varphi^2|v|^2
	\right] e^{2p\alpha(\mathbf{x},t)}\,dt\,d\mathbf{x} \\
	&= Cp^{-1} \int_\Omega \int_0^T \left[
	\frac{1}{p\varphi}|\partial_t v|^2 
	+ p^3\varphi^{2}|v|^2 
	+ p^2\varphi^{2}|v|^2
	\right] e^{2p\alpha(\mathbf{x},t)}\,dt\,d\mathbf{x}\\
	&\leq Cp^{-1} \int_\Omega \int_0^T \left[
	\frac{1}{p\varphi}|\partial_t v|^2 
	+ p^{3}\varphi^{2}|v|^2 
	+ p^{2}\varphi^{2}|v|^2
	\right] e^{2p\alpha(\mathbf{x},t)}\,dt\,d\mathbf{x}\\
	& \leq Cp^{-1} \int_\Omega \int_0^T \left[
	\frac{1}{p\varphi}|\partial_t v|^2 
	+ p^{3}\nu^{4}\varphi^{2}|v|^2 
	\right] e^{2p\alpha(\mathbf{x},t)}\,dt\,d\mathbf{x} \quad \text{for large } \nu>0\\
	&\leq Cp^{-1} \int_Q \left[
	\frac{1}{s\varphi}|\partial_t v|^2 
	+ p^{3}\nu^{4}\varphi^{3}|v|^2 
	\right] e^{2p\alpha(\mathbf{x},t)}\,dt\,d\mathbf{x}
\end{align*}

By Theorem \ref{thm:4.1}, we have

\begin{align*}
	J
	&\leq c p^{-1} \int_Q |Lv|^2 e^{2p\alpha(\mathbf{x},t)} \, dt\, d\mathbf{x} 
	+ c \int_{Q_{\omega} } p^3\nu^4 \varphi^3 v^2 
	e^{2p\alpha(\mathbf{x},t)} \, dt\, d\mathbf{x}\\
	&= c \int_{Q_{\omega} } p^2\nu^4 \varphi^3 v^2 
	e^{2p\alpha(\mathbf{x},t)} \, dt\, d\mathbf{x}, \quad \text{since } Lv=\partial_{t}v+Av=0\\
	&\leq Ce^{c_{\omega}p}  \int_{Q_{\omega} } v^{2} \,d\mathbf{x} \quad \text{(since } \alpha \text{ is bounded on } Q_{\omega} )\\
\end{align*}
Consequently we have
\begin{align*}
	\int_\Omega |f|^2 e^{2p\alpha(\mathbf{x},\frac{T}{2})}\,d\mathbf{x} &\leq   Ce^{c_{\omega}p}  \int_{Q_{\omega} } v^{2} \,d\mathbf{x} + \int_\Omega |Au(\mathbf{x},\frac{T}{2})|^2 e^{2p\alpha(\mathbf{x},\frac{T}{2})}\,d\mathbf{x}\\
\end{align*}
we denote 
$$
\alpha_{\min}= \min_{\mathbf{x}\in\bar{\Omega}}\alpha(\mathbf{x},\frac{T}{2}), \quad \alpha_{\max}= \max_{\mathbf{x}\in\bar{\Omega}}\alpha(\mathbf{x},\frac{T}{2})
$$
we get 
\begin{align*}
	\int_\Omega |f|^2 \,d\mathbf{x} &\leq   C\frac{\alpha_{\max}}{\alpha_{\min}} \int_{Q_{\omega} } v^{2} \,d\mathbf{x} + \int_\Omega |Au(\mathbf{x},\frac{T}{2})|^2 \,d\mathbf{x}\\
\end{align*}

and therefore we have
\begin{align*}
	||f||^2_{L^{2}(\Omega)}\leq C \bigg[||\partial_{t}u||^{2}_{L^{2}(Q_{\omega})} + ||Au(\cdot,t_{0})||^{2}_{L^{2}(\Omega)} \bigg]
\end{align*}
where $C=C(\frac{\alpha_{\max}}{\alpha_{\min}},t_0, \omega, \Omega)$ and thus 
\begin{align*}
	||f||_{L^{2}(\Omega)}\leq C \bigg[||\partial_{t}u||_{L^{2}(Q_{\omega})} + ||Au(\cdot,t_0)||_{L^{2}(\Omega)} \bigg]
\end{align*}
which conclude the proof.

\section{Proof of Theorem \ref{thm::generrorbounds}} \label{proof::theo2AppendixD}
Our proof method follows 
that of \cite{mishra2022estimates} with 
additional arguments to account for optimization gap and discretization errors of our PINN.

We set $\hat{u}=u^\ast - u$, and thus $\hat{u}$ satisfies the system
\begin{equation}\label{eq:hat:u}
	\begin{cases}
		\partial_t \hat{u}(\mathbf{x},t) + A\hat{u}(\mathbf{x},t) 
		= (f^* - f)(\mathbf{x}) + \mathcal{R}^*_{int}(\mathbf{x},t), & \quad (\mathbf{x},t)\in Q
		\\
		\nabla_n \hat{u}(\mathbf{x},t) = \mathcal{R}^*_{bc}(t,\mathbf{x}), &\quad (\mathbf{x},t)\in \partial Q  \\
		\hat{u}(\mathbf{x},0) = \mathcal{R}^*_{ic,\pmb{\theta}^{\ast}_{u}}(\mathbf{x}),  &\quad \mathbf{x}\in\Omega
	\end{cases}
\end{equation}
with $\hat{u}(\mathbf{x},t_0) = \mathcal{R}^{\ast}_{z,\pmb{\theta}^\ast_{u}}(\mathbf{x}) - (z(\mathbf{x}) - z^\delta(\mathbf{x}))$\quad  $\mathbf{x}\in \omega$.

\noindent We set $\hat{u} = \hat{u}_1 + \hat{u}_2$, then we have:
\begin{equation}\label{eq:u1}
	\begin{cases}
		\partial_t \hat{u}_1(\mathbf{x},t) + A\hat{u}_{1}(\mathbf{x},t) 
		= (f^* - f)(\mathbf{x}), &\quad (\mathbf{x},t)\in Q
		\\
		\nabla_n \hat{u}_1(\mathbf{x},t) = 0, &\quad (\mathbf{x},t) \in \partial Q\\
		\hat{u}_1(\mathbf{x},0) = 0,  &\quad \mathbf{x}\in\Omega
	\end{cases}
\end{equation}
and
\begin{equation*}
	\hat{u}_1(\mathbf{x},t_0) = \mathcal{R}^*_{t_{0},\pmb{\theta}_{u}}(\mathbf{x}) - (z(x) - z^\delta(x)) - \hat{u}_2(\mathbf{x},t_0) 
	\quad \text{in } \omega
\end{equation*}
\begin{equation}\label{eq:u2}
	\begin{cases}
		\partial_t \hat{u}_2(\mathbf{x},t) +  A\hat{u}_{2}(\mathbf{x},t)  
		= \mathcal{R}^*_{int,\pmb{\theta}^{\ast}_{u},\pmb{\theta}^{\ast}_{f}}(\mathbf{x},t),&\quad (\mathbf{x},t) \in  Q   \\
		\nabla_n \hat{u}_2(\mathbf{x},t)= \mathcal{R}^*_{bc,\pmb{\theta}^{\ast}_u}(\mathbf{x},t), &\quad (\mathbf{x},t) \in \partial Q 
		\\
		\hat{u}_2(0,\mathbf{x}) = \mathcal{R}^*_{ic,\pmb{\theta}^{\ast}_{u}}(\mathbf{x}), &\quad \mathbf{x}\in \Omega 
	\end{cases}
\end{equation}

\noindent By the conditional stability \eqref{condstabF} on the system \eqref{eq:u1}, we have
\begin{align*}
	\|f^* - f\|_{L^2(\Omega)} &\leq C\bigg[\|\partial_{t}\hat{u}_{1}(\mathbf{x},t)\|_{L^2(Q_{\omega})} + \|A\hat{u}_1(\cdot,t_{0})\|_{L^2(\Omega)}\bigg] ,
\end{align*}
From the regularity theory on the forward problem \eqref{eq:u1}, we have
\begin{equation}\label{regtheoryu1}
	\|\partial_{t}\hat{u}_{1}(\mathbf{x},t)\|_{L^2(Q_{\omega})}\leq \|\hat{u}_{1}(\mathbf{x},t)\|_{H^{1}(0,T;L^{2}(\Omega)} \leq C\|f^{\ast}-f\|_{L^{2}(\Omega)}
\end{equation}

since $V,D$ are smooth functions and $z(\mathbf{x})-z^{\delta}(\mathbf{x})=\delta(\mathbf{x})$, we have: 
\begin{align*}
	\|A\hat{u}_{1}(\cdot,t_0)\|_{L^{2}(\Omega)}&\leq  \| A\mathcal{R}^{\ast}_{z,\pmb{\theta}^\ast_{u}}(\mathbf{x})\|_{L^{2}(\Omega)} + \|A(z(\mathbf{x})-z^{\delta}(\mathbf{x}))\|_{L^{2}(\Omega)} +\|A\hat{u}_{2}(\cdot,t_0)\|_{L^{2}(\Omega)}\\
	&\leq  \| A\mathcal{R}^{\ast}_{z,\pmb{\theta}^\ast_{u}}(\mathbf{x})\|_{L^{2}(\Omega)} + \|\delta\|_{H^{2}(\Omega)} +\|A\hat{u}_{2}(\cdot,t_0)\|_{L^{2}(\Omega)}\\
	&\leq \| A\mathcal{R}^{\ast}_{z,\pmb{\theta}^\ast_{u}}(\mathbf{x})\|_{L^{2}(\Omega)} + \|\delta\|_{H^{2}(\Omega)} + \|\partial_{t}\hat{u}_{2}(\cdot,t_0)\|_{L^{2}(\Omega)} + \|\mathcal{R}^{\ast}_{int,\pmb{\theta}^{\ast}_{u},\pmb{\theta}^{\ast}_{f}}(\cdot,t_0)\|_{L^{2}(\Omega)}\\
	&\leq \| A\mathcal{R}^{\ast}_{z,\pmb{\theta}^\ast_{u}}(\mathbf{x})\|_{L^{2}(\Omega)} + \|\delta\|_{H^{2}(\Omega)} + \|\partial_{t}\hat{u}_{2}(\mathbf{x},t)\|_{H^{1}(0,T;L^{2}(\Omega))}\\
	&+ \|\mathcal{R}^{\ast}_{int,\pmb{\theta}^{\ast}_{u},\pmb{\theta}^{\ast}_{f}}(\cdot,t_0)\|_{L^{2}(\Omega)}
\end{align*}
we set $h(t,\mathbf{x})=\partial_{t}\hat{u}_{2}(t,\mathbf{x})$, and thus $h$ satisfies
\begin{equation}\label{eq:h}
	\begin{cases}
		\partial_{t}h(\mathbf{x},t)+Ah(\mathbf{x},t)=\partial_{t}\mathcal{R}^{\ast}_{int,\pmb{\theta}^{\ast}_{u},\pmb{\theta}^{\ast}_{f}}(\mathbf{x},t), &\quad (\mathbf{x},t)\in Q\\
		\nabla_{n}h(\mathbf{x},t)= \partial_{t}\mathcal{R}^{\ast}_{bc}(\mathbf{x},t), &\quad (\mathbf{x},t)\in \partial Q\\
		h(\mathbf{x},0)= \mathcal{R}^{\ast}_{int,\pmb{\theta}^{\ast}_{u},\pmb{\theta}^{\ast}_{f}}(\mathbf{x},0) -A \mathcal{R}^*_{ic,\pmb{\theta}^{\ast}_{u}}(\mathbf{x}) &\quad \mathbf{x}\in\Omega
	\end{cases}
\end{equation}
by applying the regularity theory on the problem \eqref{eq:h}, we have
\begin{eqnarray} \label{regtheory::h}
	\|h(\mathbf{x},t)\|_{H^{1}(0,T;L^{2}(\Omega))}&\leq C\bigg[ \|\partial_{t}\mathcal{R}^{\ast}_{int,\pmb{\theta}^{\ast}_{u},\pmb{\theta}^{\ast}_{f}}(\mathbf{x},t)\|_{L^{2}(Q)}+ \| \partial_{t}\mathcal{R}^{\ast}_{bc}(\mathbf{x},t)\|_{H^{1}(0,T;L^{2}(\partial\Omega))}  \nonumber\\
	&+ \|\mathcal{R}^{\ast}_{int,\pmb{\theta}^{\ast}_{u},\pmb{\theta}^{\ast}_{f}}(\mathbf{x},0)\|_{L^{2}(\Omega)} + \|A \mathcal{R}^*_{ic,\pmb{\theta}^{\ast}_{u}}(\mathbf{x})\|_{L^{2}(\Omega)} \bigg]
\end{eqnarray}
and thus we have
\begin{align}\label{eq:Au0}
	\|A\hat{u}_{1}(\cdot,t_0)\|_{L^{2}(\Omega)}&\leq C \bigg[\| \mathcal{R}^{\ast}_{z,\pmb{\theta}^\ast_{u}}(\mathbf{x})\|_{H^{2}(\Omega)} + \|\delta\|_{H^{2}(\Omega)} + \|\partial_{t}\mathcal{R}^{\ast}_{int,\pmb{\theta}^{\ast}_{u},\pmb{\theta}^{\ast}_{f}}(\mathbf{x},t)\|_{L^{2}(Q)}\nonumber\\
	&+ \| \partial_{t}\mathcal{R}^{\ast}_{bc}(\mathbf{x},t)\|_{H^{1}(0,T;L^{2}(\partial\Omega))} +
	\|\mathcal{R}^{\ast}_{int,\pmb{\theta}^{\ast}_{u},\pmb{\theta}^{\ast}_{f}}(\mathbf{x},0)\|_{L^{2}(\Omega)} + \| \mathcal{R}^*_{ic,\pmb{\theta}^{\ast}_{u}}(\mathbf{x})\|_{H^{2}(\Omega)} \nonumber\\
	&+ \|\mathcal{R}^{\ast}_{int,\pmb{\theta}^{\ast}_{u},\pmb{\theta}^{\ast}_{f}}(\cdot,t_0)\|_{L^{2}(\Omega)} + \|\delta\|_{H^{2}(\Omega)}\bigg]
\end{align}
By combining \eqref{regtheoryu1} and \eqref{eq:Au0}, we have
\begin{align}
	\mathcal{E}_{f} = \|f^{\ast}-f\|_{L^{2}(\Omega)} &\leq C\bigg[  \|\mathcal{R}^{\ast}_{int,\pmb{\theta}^{\ast}_{u},\pmb{\theta}^{\ast}_{f}}(\mathbf{x},t)\|_{H^{1}(0,T;L^{2}(\Omega))} + \|\mathcal{R}^{\ast}_{z,\pmb{\theta}^\ast_{u}}(\mathbf{x})\|_{H^{2}(\Omega)}\nonumber\\
	&+ \| \partial_{t}\mathcal{R}^{\ast}_{bc}(\mathbf{x},t)\|_{H^{1}(0,T;L^{2}(\partial\Omega))} + \| \mathcal{R}^*_{ic,\pmb{\theta}^{\ast}_{u}}(\mathbf{x})\|_{H^{2}(\Omega)}+  \|\delta\|_{H^{2}(\Omega)}\bigg].
\end{align}
In addition, by using the regularity theory on the PDE \eqref{eq:hat:u}, we have
\begin{align*}
	\mathcal{E}_{u}&= \|\hat{u}||_{C([0,T];L^{2}(\Omega))}\\
	&\leq C\bigg[ \|(f^* - f)(\mathbf{x}) + \mathcal{R}^*_{int}(\mathbf{x},t)\|_{L^{2}(Q)} + \|\mathcal{R}^{\ast}_{bc}(\mathbf{x},t)\|_{H^{1}(0,T;L^{2}(\partial\Omega))}+ \|\mathcal{R}^{\ast}_{ic}(\mathbf{x})\|_{L^{2}(\Omega)}\bigg] \nonumber\\
	&\leq C \bigg[ \|\mathcal{R}^*_{int}(\mathbf{x},t)\|_{L^{2}(Q)} + \|f^{\ast}-f\|_{L^{2}(\Omega)}+ \|\mathcal{R}^{\ast}_{bc}(\mathbf{x},t)\|_{H^{1}(0,T;L^{2}(\partial\Omega))}+ \|\mathcal{R}^{\ast}_{ic}(\mathbf{x})\|_{L^{2}(\Omega)}\bigg] \\
	&\leq C \bigg[ \|\mathcal{R}^*_{int}(\mathbf{x},t)\|_{L^{2}(Q)} + \|\mathcal{R}^{\ast}_{bc}(\mathbf{x},t)\|_{H^{1}(0,T;L^{2}(\partial\Omega))}+ \|\mathcal{R}^{\ast}_{ic}(\mathbf{x})\|_{L^{2}(\Omega)}\bigg] \\
	&+ C\bigg[\|\mathcal{R}^{\ast}_{z,\pmb{\theta}^\ast_{u}}(\mathbf{x})\|_{H^{2}(\Omega)}  
	+\|\mathcal{R}^{\ast}_{int,\pmb{\theta}^{\ast}_{u},\pmb{\theta}^{\ast}_{f}}(\mathbf{x},t)\|_{H^{1}(0,T;L^{2}(\Omega))} + \| \partial_{t}\mathcal{R}^{\ast}_{bc}(\mathbf{x},t)\|_{H^{1}(0,T;L^{2}(\partial\Omega))}\\
	 &+ \| \mathcal{R}^*_{ic,\pmb{\theta}^{\ast}_{u}}(\mathbf{x})\|_{H^{2}(\Omega)}
	+  \|\delta\|_{H^{2}(\Omega)}\bigg]\\
	&\leq C\bigg[ \|\mathcal{R}^{\ast}_{int,\pmb{\theta}^{\ast}_{u},\pmb{\theta}^{\ast}_{f}}(\mathbf{x},t)\|_{H^{1}(0,T;L^{2}(\Omega))} +
	\| \mathcal{R}^{\ast}_{bc}(\mathbf{x},t)\|_{H^{2}(0,T;L^{2}(\partial\Omega))}+  \|\mathcal{R}^{\ast}_{z,\pmb{\theta}^\ast_{u}}(\mathbf{x})\|_{H^{2}(\Omega)}  \\
	&+  \| \mathcal{R}^*_{ic,\pmb{\theta}^{\ast}_{u}}(\mathbf{x})\|_{H^{2}(\Omega)} + \|\delta\|_{H^{2}(\Omega)}\bigg]
\end{align*}
From \eqref{discloss} and \eqref{eq:quadrature_error} we can conclude \eqref{genError:f} and \eqref{genError:u}.

\end{document}